\documentclass[graybox]{svmult}

\usepackage{makeidx}         
\usepackage{graphicx}        

 \usepackage{capt-of}
 \usepackage{lipsum}
    
\makeindex             

\newcommand{\sllm}{\mathbf{SLLM}}
\newcommand{\bs}{\backslash}
\newcommand{\semantics}[1]{[\![ #1 ]\!]}
\newcommand{\fdvect}{\mathbf{FdVect}}
\newcommand{\ev}{\mathrm{ev}}

\usepackage{proof}
\usepackage{caption}
\usepackage{subcaption}
\usepackage{booktabs}

\usepackage{float}

\usepackage{url}

\usepackage{siunitx}
\usepackage{tikz}
\usepackage{pgfplots}
\usepackage{amsfonts}
\pgfplotsset{width=6cm,compat=1.12}

\usepackage{tikz}
\usepackage{pgfplots}
\usetikzlibrary{trees}
\usetikzlibrary{topaths}
\usetikzlibrary{decorations.pathmorphing}
\usetikzlibrary{decorations.markings}
\usetikzlibrary{matrix,backgrounds,folding}
\usetikzlibrary{chains,scopes,positioning,fit}
\usetikzlibrary{shapes,shapes.geometric,shapes.misc}

\tikzset{->-/.style={decoration={
  markings,
  mark=at position .5 with {\arrow{>}}},postaction={decorate}}}

\pgfdeclarelayer{edgelayer}
\pgfdeclarelayer{nodelayer}
\pgfsetlayers{background,edgelayer,nodelayer,main}

\tikzstyle{none}=[fill=none]
\tikzstyle{title}=[fill=none]
\tikzstyle{lab}=[fill=white, font=\tiny]
\tikzstyle{new style 0}=[fill=white, draw=black, shape=trapezium]
\tikzstyle{new style 1}=[fill=white, draw=black, shape=circle]
\tikzstyle{new style 2}=[fill=white, draw=green, shape=circle]
\tikzstyle{new style 3}=[fill=white, draw=red, shape=circle]
\tikzstyle{new style 4}=[fill=white, draw=black, thick, shape=regular polygon, regular polygon sides=3]
\tikzstyle{element}=[fill=white, draw=black, shape=trapezium]
\tikzstyle{functional}=[fill=white, draw=black, shape=trapezium,shape border rotate=180]
\tikzstyle{counit}=[fill=black, shape=circle]
\tikzstyle{bigClasp}=[fill=white, draw=black, shape=circle, minimum width=1.7cm]

\tikzstyle{downArrow}=[none]
\tikzstyle{thickArr}=[line width = 2pt]
\tikzstyle{thick}=[line width = 2pt]
\tikzstyle{thickerArr}=[->-, line width = 4pt]
\tikzstyle{thicker}=[line width = 4pt]
\tikzstyle{dashed}=[-, dashed]
\tikzstyle{every loop}=[]

\begin{document}

\title*{A Quantum Natural Language Processing Approach to Pronoun Resolution}

\author{Wazni, Lo, McPheat, Sadrzadeh}
\institute{Hadi Wazni \at University College London, \email{hadi.wazni.20@ucl.ac.uk}
\and Kin Ian Lo \at University College London \email{kin.lo.20@ucl.ac.uk}
\and Lachlan McPheat \at University College London \email{l.mcpheat@ucl.ac.uk}
\and Mehrnoosh Sadrzadeh \at University College London \email{m.sadrzadeh@ucl.ac.uk}}
%
%
\maketitle

\abstract{We use the Lambek Calculus with soft sub-exponential modalities to model and reason about discourse relations such as anaphora and ellipsis. A semantics for this logic is obtained by using truncated Fock spaces, developed in our previous work. We depict these semantic computations via a  new string diagram.  The Fock Space semantics has the advantage that its terms are learnable  from large corpora of data  using machine learning and they can be experimented with on mainstream natural language tasks. Further, and thanks to an existing translation from vector spaces to quantum circuits,  we can also  learn these terms  on  quantum computers and their simulators,  such as the IBMQ range.   We  extend the existing translation to Fock spaces and develop quantum circuit semantics for discourse relations.  We then experiment with the IBMQ AerSimulations of these circuits in a definite pronoun resolution task, where the highest accuracies were recorded for models when the  anaphora was resolved.}

\section{Introduction}
\label{sec:intro}

Seminal work of Lambek in 1958 \cite{Lambek58} showed  that the simple  logic of  concatenation   and its  residuals form a \emph{Syntactic Calculus}. This calculus,  referred to as the \emph{Lambek Calculus}, could model and reason about  grammatical structures of sentences of natural language. The atomic formulae of the logic model \emph{basic} grammatical types, e.g. noun phrases $n$ and declarative sentences $s$. Concatenations of formulae  model compositions of types. 


Subsequent work of  Moortgat  in 1996 \cite{moortgat1996multimodal},  and Jaeger in  1998 \cite{jager1998multi},  Morrill in 2015, 2016 \cite{morrill2015computational,morrill2016logic}, and  Kanovich et. al in 2016, 2020 \cite{Kanovich2016,Kanovich2020}  showed that adding modalities to the Lambek calculus increases its expressive power. Whereas Moortgat used these modalities to restrict  associativity, Morrill was in favour of using them for island types and iterative conjunctives. Kanovitch et. al focused on parasitic gaps.  All these models worked at the  sentence level. The work of Jaeger, on the other hand, extended the application of Lambek Calculus to the discourse level and to model the relationship  between two or more sentences.   An example of such a relationship is coreference, that is, when words from different sentences refer to each other.  This phenomena is observed in anaphora, when a pronoun such as `he' refers to a noun phrase, such as `John', in the discourse `John slept. He snored.'.   Another example of coreference is ellipsis, when an ellipsis marker such as `too' refers to a verb phrase such as `slept' in the discourse  `John slept. Bill did too'. 

In previous work \cite{mcpheat2021LACL}, we combined Jaeger's idea  with   \emph{Lambek Calculus with Soft Sub-Exponentials} introduced by Kanovitch et. al in \cite{Kanovich2020}, since it had better meta logical properties, i.e. finite rules, cut elimination,  strong normalisation, and decidability.  We showed how it can  model and reason about  anaphora and ellipsis and developed a  finite dimensional vector spaces for it. This development was in the style of the  vector space semantics of Lambek calculus \cite{Coeckeetal2013}, but with a novel feature. The soft sub-exponentiated formulae  were interpreted as \emph{truncated Fock spaces}.  A Fock space is the direct sum of all tensor powers of a vector space. Fixing the field to be $\mathbb{F}$,  the Fock space  over a vector space $V$ is  
\[
    \mathbb{F} \oplus V \oplus (V \otimes V) \oplus (V \otimes V \otimes V \otimes V) \oplus \cdots
\]
 A truncated version of this space is obtained restricting the  Fock space to its $k_0$'th tensor power, for $k_0$ a fixed bound, defined as follows:
 \[
     \mathbb{F} \oplus V \oplus (V \otimes V) \oplus (V \otimes V \otimes V \otimes V) \oplus \underbrace{(V \otimes V \otimes \cdots \otimes V)}_{k_0}\,.
 \]

The bound $k_0$ is  fixed by the logic.  The soft sub-exponentiated formulae of the logic $!A$ can be thought of as \emph{storages of formulae}, which only contain  $k_0$ copies of a formula $A$. Access to these copies is obtained by a $!$ elimination rule in the logic and by projection to the right level  of a truncated Fock space in the semantics. When modelling coreference relations such as  anaphora,   only a fixed number of  pronouns  refer to a noun phrase in any given discourse.  An upper bound can easily be determined from these fixed numbers, e.g. by averaging. 

Historically, Lambek Calculus has  enjoyed a relational semantics and  a question arises that  `what is the benefit of working with a vector space semantics?'.  The answer comes from recent advances in Natural Language Processing, where vector  semantics are learnable via machine learning algorithms such as neural networks. Having a vector space semantics for Lambek Calculus and its modal extensions enables us to work with machine learnt vector  representations of words, sentences, and discourse units in a structured way. In practice, however, these semantics rely on higher order tensors building which is computationally costly. A recent line of research \cite{meichanetzidis2020quantum,QnlpInPractice}  known as \emph{Quantum Natural Language Processing} (QNLP) argues that this problem is solvable by using Quantum computers.  Tensors are  native to Quantum computers and can be learnt by them using  less resources.   

We use the vector space-to-Quantum circuit translation  of \cite{meichanetzidis2020quantum,QnlpInPractice} and turn our truncated Fock space semantics into  Quantum circuits using a novel string diagrammatic calculus that we develop for truncated Fock spaces. We then use  the \emph{DisCoPy} and  \emph{Lambeq} tools \cite{de_Felice_2021,lambeq_paper} to learn the parameters of these circuits on a definite pronoun anaphora resolution task inspired by the Winograd Schema Challenge \cite{levesque2012}.  This challenge consists of sentences with ambiguous anaphoric reference relations, where a definite pronoun in the second sentence can either refer to the subject or the object of the first sentence. The goal of this task is to disambiguate the reference relation. We implement the task as a classification task and train a model using the IBMQ \emph{AerSimulator} \cite{IBMQ-sim}. We implemented eight different models in which we experimented with presence or lack of grammatical structure and discourse structure. All the models converged  but the highest accuracies were recorded for models where anaphora is resolved, irregardless of  the model encoding the grammatical structure or not. The lowest accuracies were observed for models when anaphora was not resolved and there was no notion of  grammar. 

The current results are obtained over a small dataset of 144 entries where each grammatical type is only modelled by a one qubit quantum state. Experimenting with a large scale dataset and solidifying these results is  work in progress.

\section{Lambek calculus with soft Subexponentials ($\sllm$) and its applications to modelling discourse structure}

The formulae of  Lambek calculus with soft Subexponentials, in short $\sllm$ are generated using the following BNF.  

\[A,B::= A\in At \mid A\cdot B \mid A\bs B \mid A/B \mid !A \mid \nabla A\]

The set of atoms can be any set of indices. For the linguistic application purposes of this paper, we set the atoms to be  $\{s,n\}$ with $s$ representing the grammatical  `declarative sentence' and $n$ the  `noun phrase'. With the above BNF over this set of atoms we can generate the usual set of complex  types, including  the formula for an adjective $n/ n$, the formula for an intransitive verb  $n\bs s$ and the formula for a transitive verb  $n\bs s /n$. 

Given the types of the calculus, we  define its \emph{sequents} to be pairs, written $\Gamma \longrightarrow A$, where $\Gamma$ is a finite list of formulas, $\Gamma = A_1, A_2,\ldots, A_n$ and $A$ is a formula. Given this notion of  sequents, we define the logical rules  of $\sllm$ in a Gentzen calculus style below.

\begin{table}
\[\begin{array}{ll}
\,\\\\
\infer[I]
{A\longrightarrow A}
{}
\\\\
\infer[\bs_L]
{\Sigma_1,\Gamma, A\bs B, \Sigma_2 \longrightarrow C}
{\Gamma \longrightarrow A & \Sigma_1, B, \Sigma_2 \longrightarrow C}
& \qquad
\infer[\bs_R]
{\Gamma \longrightarrow A\bs B}
{A,\Gamma \longrightarrow B}
\\\\
\infer[/_L]
{\Sigma_1, B/ A,\Gamma, \Sigma_2 \longrightarrow C}
{\Gamma \longrightarrow A & \Sigma_1, B, \Sigma_2 \longrightarrow C}
& \qquad
\infer[/_R]
{\Gamma \longrightarrow B/A}
{\Gamma,A \longrightarrow B}
\\\\
\infer[\cdot_L]
{\Gamma_1, A\cdot B, \Gamma_2 \longrightarrow C}
{\Gamma_1, A, B, \Gamma_2 \longrightarrow C}
& \qquad
\infer[\cdot_R]
{\Gamma_1,\Gamma_2 \longrightarrow A\cdot B}
{\Gamma_1 \longrightarrow A 
& 
\Gamma_2 \longrightarrow B}
\\\\
\infer[!_L]
{\Gamma_1,!A, \Gamma_2 \longrightarrow B}
{\Gamma_1,\overbrace{A,A,\ldots, A}^{n\mbox{ times}}, \Gamma_2 \longrightarrow B} 
& \qquad
\infer[!_R]
{!A \longrightarrow !B}
{A \longrightarrow B}
\\\\
\infer[\nabla_L]
{\Gamma_1,\nabla A, \Gamma_2 \longrightarrow B}
{\Gamma_1,A, \Gamma_2 \longrightarrow B}
& \qquad
\infer[\nabla_R]
{\nabla A\longrightarrow \nabla B}
{A\longrightarrow B}
\\\\
\infer[perm]
{\Gamma_1,\nabla A,\Gamma_2, \Gamma_3 \longrightarrow B}
{\Gamma_1,\Gamma_2,\nabla A, \Gamma_3 \longrightarrow B}
& \qquad
\infer[perm']
{\Gamma_1,\Gamma_2,\nabla A, \Gamma_3 \longrightarrow B}
{\Gamma_1,\nabla A,\Gamma_2, \Gamma_3 \longrightarrow B}
\end{array}\]
\caption{Sequent presentation of $\sllm$}
\label{tab:SLLMpresentation}
\end{table}

In linguistic applications,  $\Gamma$ often represents the grammatical types of a  string of words, and $A$  is the result of the composition of these types. An example of a grammatical rule of English is that  a noun  phrase such as (\textit{John}, composes with an intransitive verb  such as \textit{sleeps}  to form a sentence, that is \textit{John sleeps}. This  is modelled  in $\sllm$ via  sequent $n, n\bs s \longrightarrow s$, and its proof tree, which is as follows:

\begin{equation}
\infer[\bs_L]
{n,n\bs s \longrightarrow s}
{\infer[]{n\longrightarrow n}{} 
&
\infer[]{s\longrightarrow s}{}}
\label{proof:JohnSleeps}
\end{equation}


The modality $!$ is known as a soft sub-exponential. The  banged formulae are thought of as \emph{storages} and the $!$-modality itself as a \emph{projection} operation. Reading $!_L$-structural rule from bottom to top, we are projecting from the storage $!A$ which contains  $k_0$ formulae  $A$, to $n$ copies of $A$, for $1\leq n\leq k_0$. This allows for the existence of a  controlled   implicit notion of copying in the syntax, which is not the same as the usual on-the-nose notion of copying: we are  not replacing $!A$ with copies of $!A$ nor are we replacing $A$ with copies of $A$. We are just projecting from a storage containing many copies of $A$ to a smaller number of those copies. This implicit notion of copying comes from Soft Linear Logic \cite{Lafont2004} has a neat interpretation in the vector space semantics as we show in \ref{subsec:VSS}. The second modality, $\nabla$, is the one that allows its formulas to be permuted. This is seen in the rules $perm$ and $perm'$ in table \ref{tab:SLLMpresentation}.

The bounding of $n$ is strictly necessary, as having an unbounded number of copies makes the calculus' derivability problem undecidable. This bound is not a problem in terms of modelling language, as the bound corresponds to the number of times one may refer to something in a discourse. A bad but symbolic bound then would be the number of words in the discourse you want to analyse. 

The way we model discourse phenomena such as  anaphora and ellipsis is similar to that of   J\"{a}ger's \cite{jager1998multi,jager2006anaphora}. In J\"{a}ger's work  the word that is being referred to is explicitly copied and moved  to the site of the referring word, where it gets applied. In our framework, however, we do not have explicit copying and use  a storage type for the word that is being referred to. For example, if we wish to model the discourse \textit{John sleeps. He snores.}, we need to first assign a storage type to \textit{John}, then project two copies from it,  and move one  to  the site of the type of \textit{He}.  Then apply the type of  \textit{He} to the moved copy of \textit{John}.  So we  assign the type    $!\nabla n$ to \textit{John}, and  assign the type $\nabla n\bs n$ to the pronoun \textit{He}.  This latter represents a function which takes a projected-from-a-storage  noun as its (left)input and returns a noun  as its outputs. This assignment is summarised below:   

\[\{(\text{\emph{John}}:\ !\nabla n),(\text{\emph{sleeps}}: n\bs s), (\text{\emph{He}}:\ \nabla n\bs n), (\text{\emph{snores}}:n\bs s)\}.\]

\noindent  The discourse  structure of \textit{John sleeps. He snores.} is modelled  by the following $\sllm$ proof tree:

\[\scalebox{0.85}{
\infer[!L]
{!\nabla n, n\bs s,\ \nabla n\bs n, n\bs s \longrightarrow s,s}
{\infer[perm]{\nabla n,\,\nabla n, n\bs s,\ \nabla n\bs n, n\bs s \longrightarrow s,s}
{\infer[\nabla_L]{\nabla n, n\bs s,\,\nabla n, \ \nabla n\bs n, n\bs s \longrightarrow s,s}
{\infer[\bs_L]{n, n\bs s,\,\nabla n, \ \nabla n\bs n, n\bs s \longrightarrow s,s}
{\infer[]{n \longrightarrow n}{}
& 
\infer[\bs_L]{s,\,\nabla n, \ \nabla n\bs n, n\bs s \longrightarrow s,s}
{\infer[]{\nabla n \longrightarrow \ \nabla n}{}
&
\infer[\bs_L]{s,n,n\bs s \longrightarrow s,s}
{\infer[]{n \longrightarrow n }{}
&
\infer[]{s,s \longrightarrow s,s}{}}}}}}}
}\]

\noindent
Reading the proof from bottom to top, we see that the application of $!_L$ replaces the type of \textit{John} ($!\nabla n$) with two copies of \textit{John} ($\nabla n$). The application of $perm$ moves one of the copies next to \textit{He}. The following $\bs_L$ application identifies \textit{He} with \textit{John}, and the application of $\nabla_L$ `forgets' that the other copy of \textit{John} is a copy. The sequent above the application of $\nabla_L$ corresponds to the typing of \textit{John sleeps. John snores.} and the proof of it shows that this is indeed two sentences.


In order to show that the system can model  more complex example, we  consider the anaphoric reference: \textit{The ball hit the window and Bill caught it}, directly from the pronoun resolution dataset \cite{rahman-ng-2012-resolving}. Here,  \textit{it} refers to \textit{The ball}. For brevity we may type the whole noun phrase \textit{The ball} as $!\nabla n$, and the noun phrase \textit{the window} as $n$\footnote{The modelling can me made be more precise  by typing all instances  of \textit{the} to  $n/n$ and typing  \textit{ball} to  $!\nabla n$ and \textit{window}  to $n$.}
, and as usual we type \textit{and} as $s\bs s /s$ and both verbs \textit{hit} and \textit{caught} as $n\bs s /n$. Finally, we type the pronoun \textit{it} as $\nabla n \bs n$. Note that this example consists of only one sentence. One can easily break replace the \textit{and} with a full stop and work with two sentences. Since we already have a two-sentence example, we will model the one-sentence case, which is modelled in the following proof tree.

\begin{equation}
\infer[!_L]
{!\nabla n, n\bs s /n, n, s\bs s /s, n, n\bs s /n, \nabla n\bs n \longrightarrow s}
{\infer[perm]{\nabla n,\nabla n, n\bs s /n, n, s\bs s /s, n, n\bs s /n, \nabla n\bs n \longrightarrow s}
{\infer[perm]{\nabla n, n\bs s /n,\nabla n, n, s\bs s /s, n, n\bs s /n, \nabla n\bs n \longrightarrow s}
{\infer[perm]{\nabla n, n\bs s /n, n,\nabla n, s\bs s /s, n, n\bs s /n, \nabla n\bs n \longrightarrow s}
{\infer[perm]{\nabla n, n\bs s /n, n, s\bs s /s,\nabla n, n, n\bs s /n, \nabla n\bs n \longrightarrow s}
{\infer[perm]{\nabla n, n\bs s /n, n, s\bs s /s, n,\nabla n, n\bs s /n, \nabla n\bs n \longrightarrow s}
{\infer[\bs_L]{\nabla n, n\bs s /n, n, s\bs s /s, n, n\bs s /n,\nabla n, \nabla n\bs n \longrightarrow s}
{\infer[]{\nabla n \longrightarrow \nabla n}{}
&
\infer[\nabla_L]{\nabla n, n\bs s /n, n, s\bs s /s, n, n\bs s /n, n \longrightarrow s}
{\infer[\bs_L]{n, n\bs s /n, n, s\bs s /s, n, n\bs s /n, n \longrightarrow s}
{\infer[]{ n \longrightarrow n}{}
&
\infer[/_L]{s /n, n, s\bs s /s, n, n\bs s /n, n \longrightarrow s}
{\infer[]{ n \longrightarrow n}{}
&
\infer[\bs_L]{s,s\bs s /s, n, n\bs s /n, n \longrightarrow s}
{\infer[]{s \longrightarrow s}{}
&
\infer[\bs_L]{s /s, n, n\bs s /n, n \longrightarrow s}
{\infer[]{ n \longrightarrow n}{}
&
\infer[/_L]{s /s, s /n, n \longrightarrow s}
{\infer[]{ n \longrightarrow n}{}
&
\infer[/_L]{s /s, s \longrightarrow s}
{\infer[]{s \longrightarrow s}{}
&
\infer[]{s \longrightarrow s}{}
}}}}}}}}}}}}}}
\label{proof:BallHitWindow}
\end{equation}

Ellipsis is modelled similarly. Since we have only experimented with anaphoric reference in the current paper, we refer the reader   for the typing and proof tree of an example, e.g.  \textit{John plays guitar. Mary does too.} to previous work \cite{mcpheat2021LACL}. 

 \textit{The dog broke the vase. It was clumsy}, which naturally has type $s\cdot s$ in the following proof:
\[
\mbox{\textit{the dog}} : !\nabla n, \,
\mbox{\textit{broke}} : n\bs s /n, \,
\mbox{\textit{the vase}} : n,\,
\mbox{\textit{It}} : \nabla n \bs n,\,
\mbox{\textit{was}} : n\bs s /(n/n),\,
\mbox{\textit{clumsy}} : n/n,
\]

\begin{equation}
\infer[!_L]
{!\nabla n, n\bs s /n, n, \nabla n\bs n, n\bs s /(n/n), n/n \longrightarrow s\cdot s}
{\infer[perm]{\nabla n, \nabla n, n\bs s /n, n, \nabla n\bs n, n\bs s /(n/n), n/n \longrightarrow s\cdot s}
{\infer[perm]{\nabla n, n\bs s /n, \nabla n, n, \nabla n\bs n, n\bs s /(n/n), n/n \longrightarrow s\cdot s}
{\infer[\bs_L]{\nabla n, n\bs s /n, n, \nabla n, \nabla n\bs n, n\bs s /(n/n), n/n \longrightarrow s\cdot s}
{\infer[\nabla_R]{\nabla n \longrightarrow \nabla n}
{\infer[]{n \longrightarrow n}{}}
&
\infer[\nabla_L]{\nabla n, n\bs s /n, n, n, n\bs s /(n/n), n/n \longrightarrow s\cdot s}
{\infer[\bs_L]{n, n\bs s /n, n, n, n\bs s /(n/n), n/n \longrightarrow s\cdot s}
{\infer[]{n \longrightarrow n}{}
&
\infer[/_L]{s /n, n, n, n\bs s /(n/n), n/n \longrightarrow s\cdot s}
{\infer[]{n \longrightarrow n}{}
&
\infer[\bs_L]{s,n, n\bs s /(n/n), n/n \longrightarrow s\cdot s}
{\infer[]{n \longrightarrow n}{}
&
\infer[]{s,s /(n/n), n/n \longrightarrow s\cdot s}
{\infer[/_R]{n/n \longrightarrow n/n}
{\infer[/_L]{n/n,n\longrightarrow n}
{\infer[]{n \longrightarrow n}{}
&
\infer[]{n \longrightarrow n}{}}}
&
\infer[\cdot_R]{s,s \longrightarrow s\cdot s}
{\infer[]{s \longrightarrow s}{}
&
\infer[]{s \longrightarrow s}{}}}}}}}}}}}
\label{proof:JohnSleepsHeSnores}
\end{equation}

Similarly we can resolve object-anaphora, as in the example: \textit{The cat broke the glass. It was fragile.}

\[
\mbox{\textit{the cat}} : n, \,
\mbox{\textit{broke}} : n\bs s /n, \,
\mbox{\textit{the glass}} : !\nabla n,\,
\mbox{\textit{It}} : \nabla n \bs n,\,
\mbox{\textit{was}} : n\bs s /(n/n),\,
\mbox{\textit{fragile}} : n/n,
\]

\begin{equation}
\infer[!_L]
{n, n\bs s /n, !\nabla n, \nabla n\bs n, n\bs s /(n/n), n/n \longrightarrow s\cdot s}
{\infer[\bs_L]{n, n\bs s /n,\nabla n,\nabla n, \nabla n\bs n, n\bs s /(n/n), n/n \longrightarrow s\cdot s}
{\infer[]{\nabla n\longrightarrow \nabla n}{}
&
 \infer[\nabla_L]{n, n\bs s /n,\nabla n, n, n\bs s /(n/n), n/n \longrightarrow s\cdot s}
 {\infer[\bs_L]{n, n\bs s /n, n, n, n\bs s /(n/n), n/n \longrightarrow s\cdot s}
 {\infer[]{ n\longrightarrow n}{}
 &
 \infer[/_L]{s /n, n, n, n\bs s /(n/n), n/n \longrightarrow s\cdot s}
 {\infer[]{ n\longrightarrow n}{}
 &
 \infer[\bs_L]{s, n, n\bs s /(n/n), n/n \longrightarrow s\cdot s}
 {\infer[]{ n\longrightarrow n}{}
 &
 \infer[/_L]{s, s /(n/n), n/n \longrightarrow s\cdot s}
 {
 \infer[/_R]{n/n \longrightarrow n/n}
{\infer[/_L]{n/n,n\longrightarrow n}
{\infer[]{n \longrightarrow n}{}
&
\infer[]{n \longrightarrow n}{}}}
&
\infer[\cdot_R]{s,s \longrightarrow s\cdot s}
{\infer[]{s \longrightarrow s}{}
&
\infer[]{s \longrightarrow s}{}
 }}}}}}}}
\label{proof:fragileVase}
\end{equation}

\section{Truncated Fock space semantics of $\sllm$}\label{subsec:VSS}
In this section we recall the definition of the vector space semantics of $\sllm$, first defined in \cite{mcpheat2021LACL}. We define the semantics inductively on formulas, which are interpreted as vector spaces and proofs, which are interpreted as linear maps. We will use a semantic bracket notation $\semantics{\,} : \sllm \to \fdvect$ to denote the semantics of formulas $\semantics{A}$ and the semantics of proofs $\semantics{\pi}$, where if $\pi$ is a proof of a sequent $\Gamma \longrightarrow A$, then $\semantics{\pi}$ is a linear map from $\semantics{\Gamma}$ to $\semantics{A}$.

\begin{enumerate}
\item For the atomic formulas $n, s$ we interpret them as some fixed vector spaces $N := \semantics{n}$ and $S:= \semantics{s}$.
\item Formulas of the form $A\cdot B$ are interpreted using the tensor product $\semantics{A\cdot B} := \semantics{A}\otimes \semantics{B}$.
\item Formulas of the form $A\bs B$ are interpreted using the vector space dual $\semantics{A\bs B} := \semantics{A}^*\otimes \semantics{B}$. Similarly, we have $\semantics{B/ A} := \semantics{B}\otimes \semantics{A}^*$.
\item $\nabla$-formulas are interpreted trivially $\semantics{\nabla A} := \semantics{A}$.
\item $!$-formulas are interpreted using truncated Fock-spaces $\semantics{!A} := T_{k_0}\semantics{A}$, where 
 \[T_{k_0}\semantics{A} = \bigoplus_{i=0}^{k_0}\semantics{A}^{\otimes i} = k \oplus \semantics{A} \oplus (\semantics{A} \otimes \semantics{A}) \oplus (\semantics{A}\otimes
 \semantics{A}\otimes \semantics{A}) \oplus \cdots \oplus\semantics{A}^{\otimes k_0}.
 \]
\end{enumerate}

As a notational convenience, we will write $\semantics{\Gamma}$ for $\semantics{A_1}\otimes \semantics{A_2}\otimes \cdots \otimes \semantics{A_n}$ for $\Gamma = A_1, \ldots, A_n$.
 
 The vector space semantics of proofs is as follows:
 \begin{enumerate}
 \item The axiom $\infer[]{A\longrightarrow A}{}$ is simply interpreted as the identity matrix $I_{\semantics{A}}:\semantics{A}\to \semantics{A}$.

 \item The $\cdot_L$ rule is trivially interpretted, as the semantics does not distinguish between the $,$ and the $\cdot$.

 \item The $\cdot_R$ rule is interpreted as the tensor product of linear maps. That is, given proofs of the hypotheses of the $\cdot_R$-rule $\pi_1$ of $\Gamma_1 \longrightarrow A$ and $\pi_2$ of $\Gamma_2 \longrightarrow B$ we have the semantics of the new proof, ending with the $\cdot_R$ rule as $\semantics{\pi_1}\otimes \semantics{\pi_2}$.

 \item The $\bs_L$ and $/_L$-rules are interpreted as application. That is, given proofs $\pi$ of $\Gamma \longrightarrow A$ and $\tau$ of $\Sigma_1, B, \Sigma_2 \longrightarrow C$ we have the semantics of the proof ending with the $\bs_L$-rule being 
 \[\semantics{\tau}\circ
 (I_{\semantics{\Sigma_1}}\otimes \ev^l_{\semantics{A},\semantics{B}} \otimes I_{\semantics{\Sigma_2}}) \circ 
            (I_{\semantics{\Sigma_1}}\otimes\semantics{\pi} \otimes I_{\semantics{A \bs B}} \otimes I_{\semantics{\Sigma_2}} ).\] 
 Similarly, for the $/_L$-rule with the same $\pi$ and $\tau$ as above, we have that the semantics of the new proof, now ending with $/_L$ is
 \[\semantics{\tau}\circ
 (I_{\semantics{\Sigma_1}}\otimes \ev^r_{\semantics{A},\semantics{B}} \otimes I_{\semantics{\Sigma_2}}) \circ 
            (I_{\semantics{\Sigma_1}} \otimes I_{\semantics{B / A}}\otimes\semantics{\pi} \otimes I_{\semantics{\Sigma_2}} ).\]

 \item The $\bs_R$ and $/_R$-rules are interpreted as currying. That is, given a proof $\pi$ of $A,\Gamma \longrightarrow B$, the semantics of the proof beginning with $\pi$ and ending with $\bs_R$ is the curried version of $\semantics{\pi}$, which we denote as \[\Lambda^l\semantics{\pi} : \semantics{\Gamma}\to \semantics{A} \otimes \semantics{B}.\]
 Similarly for the $/_R$-rule, if we assume the sequent $\Gamma, A\longrightarrow B$ has proof $\tau$, the new proof ending in $/_R$ has semantics \[\Lambda_r\semantics{\tau} : \Gamma \to \semantics{B} \otimes \semantics{A}.
 \]
 
 \item Both the $\nabla_L$ and $\nabla_R$-rules are trivial, since we interpret $\nabla$ trivially in the vector space semantics.

 \item The $perm$ and $perm'$-rules are interpreted using the symmetry, $\sigma$, of the tensor product. Recall that for any two vector spaces $V,W$ we have that $\sigma_{V,W}: V\otimes W \cong W\otimes V$. Using this symmetry we can interpret a proof $\pi$ followed by the $perm$-rule as $\semantics{\pi}\circ (I_{\semantics{\Gamma_1}} \otimes \sigma_{\semantics{A},\semantics{\Gamma_2}} \otimes I_{\semantics{\Gamma_3}})$ 

 \item The $!_L$-rule, the one that lets us copy, is interpreted using the projections from the truncated Fock space. Recall that whenever you have a direct sum $V\oplus W$ of vector spaces $V,W$ we have two canonical projections, namely $p_V : V \oplus W \to V$ and $p_W: V\oplus W \to W$ defined as $p_V(v,w) = v$ and $p_W(v,w) = w$. These projections extend to any number of summands, which in our case is $k_0$. Thus, if we consider a proof $\pi$ of a sequent $\Gamma_1,A,A,\ldots, A, \Gamma_2 \longrightarrow B$ (with $n$ instances of $A$, say) which is followed by an application of the $!_L$-rule, the semantics of the whole proof becomes 
\[
\semantics{\pi}\circ (I_{\semantics{\Gamma_1}} \otimes p_n \otimes I_{\semantics{\Gamma_2}})
\]
where $p_n : T_{k_0}\semantics{A} \to \semantics{A}^{\otimes n}$, is the $n$th projection map.

\item The $!_R$-rule is interpreted as the application of $T_{k_0}$. That is, given a proof $\pi$ of the sequent $A\longrightarrow B$, which is followed by an application of $!_R$ to give $!A \longrightarrow !B$, the semantics of the whole proof is $T_{k_0}\semantics{\pi}$. By this notation we mean the following map 
\[
(T_{k_0}\semantics{\pi})(\bigotimes_{i=0}^{j}a_i^j)_{j=0}^{k_0}
:= 
(\bigotimes_{i=0}^{j}\semantics{\pi}a_i^j)_{j=0}^{k_0}
\]
where $a_0$ is some element in the ground field, and $a_i^j \in \semantics{A}$ for all $1\leq i\leq j\leq k_0$. That $T_{k_0}\semantics{\pi}$ yields a linear map is not obvious, but it is a well known fact proven in any text on universal enveloping algebras, for instance \cite{Humphreys}.
 \end{enumerate}

\section{Diagrammatic Computations}
\label{sec:diag}

\subsection{Existing String Diagrammatic Calculus}

\begin{figure}
\centering
\begin{subfigure}[c]{0.3\textwidth}
\[{%
\beginpgfgraphicnamed{objects}
\begin{tikzpicture}
	\begin{pgfonlayer}{nodelayer}
		\node [style=none] (0) at (0, 0.5) {};
		\node [style=none] (1) at (0, -0.5) {};
		\node [style=none] (2) at (0.5, 0) {$V$};
	\end{pgfonlayer}
	\begin{pgfonlayer}{edgelayer}
		\draw [style=downArrow] (0.center) to (1.center);
	\end{pgfonlayer}
\end{tikzpicture}}
\endpgfgraphicnamed}\]
\caption{Vector spaces}\label{fig:DiagramObjects}
\end{subfigure}
\begin{subfigure}[c]{0.4\textwidth}
\[{%
\beginpgfgraphicnamed{tensorObjects}
\begin{tikzpicture}
	\begin{pgfonlayer}{nodelayer}
		\node [style=none] (0) at (-0.5, 0.75) {};
		\node [style=none] (1) at (-0.5, -0.25) {};
		\node [style=none] (2) at (-1, 0.25) {$V$};
		\node [style=none] (3) at (0.5, 0.75) {};
		\node [style=none] (4) at (0.5, -0.25) {};
		\node [style=none] (5) at (1, 0.25) {$W$};
	\end{pgfonlayer}
	\begin{pgfonlayer}{edgelayer}
		\draw [style=downArrow] (0.center) to (1.center);
		\draw [style=downArrow] (3.center) to (4.center);
	\end{pgfonlayer}
\end{tikzpicture}}
\endpgfgraphicnamed}\]
\caption{Tensored vector spaces}\label{fig:DiagramTensorProduct}
\end{subfigure}
\caption{Standard graphical calculus for vector spaces.}
\end{figure}
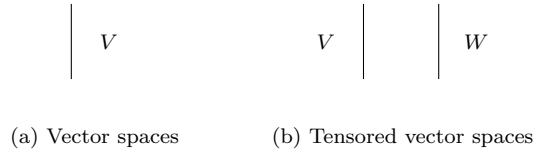

We recall the standard graphical language for finite dimensional vector spaces.
Vector spaces are denoted by labelled strings, as in figure \ref{fig:DiagramObjects} where we have drawn a vector space $V$. By convention, the 1-dimensional vector space is not drawn at all. 
The tensor product of two vector spaces is denoted by placing the corresponding strings side-by-side as in figure \ref{fig:DiagramTensorProduct}, where we have drawn $V\otimes W$.

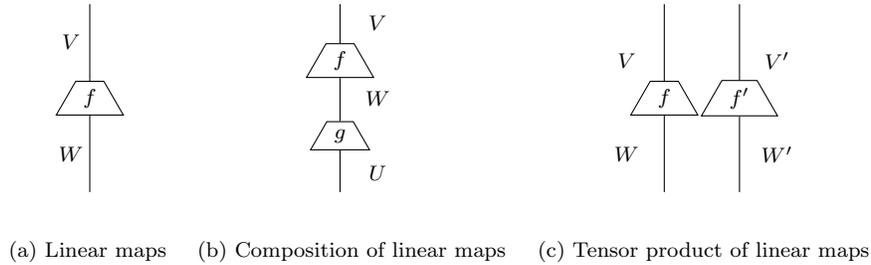
\begin{figure}
\begin{subfigure}[c]{0.2\textwidth}
\[{%
\beginpgfgraphicnamed{morphisms}
\begin{tikzpicture}
	\begin{pgfonlayer}{nodelayer}
		\node [style=none] (0) at (0, 1.25) {};
		\node [style=none] (1) at (0, -1.25) {};
		\node [style=none] (2) at (-0.25, 0.75) {$V$};
		\node [style=new style 0] (3) at (0, 0) {$f$};
		\node [style=none] (4) at (-0.25, -0.75) {$W$};
	\end{pgfonlayer}
	\begin{pgfonlayer}{edgelayer}
		\draw [style=downArrow] (0.center) to (3);
		\draw [style=downArrow] (3) to (1.center);
	\end{pgfonlayer}
\end{tikzpicture}}
\endpgfgraphicnamed}\]
\caption{Linear maps}\label{fig:DiagramMorphims}
\end{subfigure}\hfill
\begin{subfigure}[c]{0.4\textwidth}
\[{%
\beginpgfgraphicnamed{composition}
\begin{tikzpicture}
	\begin{pgfonlayer}{nodelayer}
		\node [style=none] (0) at (0, 0.75) {};
		\node [style=none] (2) at (0.5, 0.5) {$V$};
		\node [style=new style 0] (3) at (0, 0) {$f$};
		\node [style=none] (4) at (0.5, -0.5) {$W$};
		\node [style=none] (5) at (0, -1.75) {};
		\node [style=new style 0] (6) at (0, -1) {$g$};
		\node [style=none] (7) at (0.5, -1.5) {$U$};
	\end{pgfonlayer}
	\begin{pgfonlayer}{edgelayer}
		\draw [style=downArrow] (6) to (5.center);
		\draw [style=downArrow] (0.center) to (3);
		\draw [style=downArrow] (3) to (6);
	\end{pgfonlayer}
\end{tikzpicture}}
\endpgfgraphicnamed}\]
\caption{Composition of linear maps}\label{fig:DiagramComposition}
\end{subfigure}\hfill
\begin{subfigure}[c]{0.4\textwidth}
\[{%
\beginpgfgraphicnamed{tensorMorphisms}
\begin{tikzpicture}
	\begin{pgfonlayer}{nodelayer}
		\node [style=none] (0) at (-1, 0.5) {$V$};
		\node [style=none] (1) at (-0.5, -1.25) {};
		\node [style=new style 0] (2) at (-0.5, 0) {$f$};
		\node [style=none] (3) at (-1, -0.75) {$W$};
		\node [style=none] (4) at (-0.5, 1.25) {};
		\node [style=none] (5) at (1, 0.5) {$V'$};
		\node [style=none] (6) at (0.5, -1.25) {};
		\node [style=new style 0] (7) at (0.5, 0) {$f'$};
		\node [style=none] (8) at (1, -0.75) {$W'$};
		\node [style=none] (9) at (0.5, 1.25) {};
	\end{pgfonlayer}
	\begin{pgfonlayer}{edgelayer}
		\draw [style=downArrow] (2) to (1.center);
		\draw [style=downArrow] (4.center) to (2);
		\draw [style=downArrow] (7) to (6.center);
		\draw [style=downArrow] (9.center) to (7);
	\end{pgfonlayer}
\end{tikzpicture}}
\endpgfgraphicnamed}\]
\caption{Tensor product of linear maps}\label{fig:DiagramTensorMorphisms}
\end{subfigure}\hfill
\caption{Linear maps}
\end{figure}

Linear maps $f:V\to W$ are denoted as boxes on strings, with their domain feeding into the box from above, and the codomain coming out from below, as in figure \ref{fig:DiagramMorphims}.
The composition of linear maps is denoted by vertical superposition of boxes, as in \ref{fig:DiagramComposition}, where we have drawn the composition of maps $f:V\to W$ and $g: W\to U$.
The tensor product of linear maps is depicted by horizontal juxtaposition, as in figure \ref{fig:DiagramTensorMorphisms}

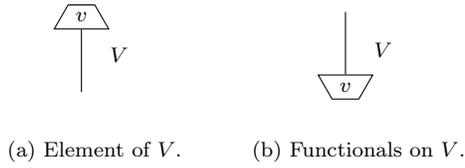
\begin{figure}
\centering
\begin{subfigure}[c]{0.3\textwidth}
\[{%
\beginpgfgraphicnamed{element}
\begin{tikzpicture}
	\begin{pgfonlayer}{nodelayer}
		\node [style=element] (0) at (0, 0.5) {$v$};
		\node [style=none] (1) at (0, -0.5) {};
		\node [style=none] (2) at (0.5, 0) {$V$};
	\end{pgfonlayer}
	\begin{pgfonlayer}{edgelayer}
		\draw [style=downArrow] (0) to (1.center);
	\end{pgfonlayer}
\end{tikzpicture}}
\endpgfgraphicnamed}\]
\caption{Element of $V$.}\label{fig:DiagramsVectors}
\end{subfigure}
\begin{subfigure}[c]{0.3\textwidth}
\[{%
\beginpgfgraphicnamed{functional}
\begin{tikzpicture}
	\begin{pgfonlayer}{nodelayer}
		\node [style=none] (1) at (0, 1) {};
		\node [style=none] (2) at (0.5, 0.5) {$V$};
		\node [style=functional] (3) at (0, 0) {$v$};
	\end{pgfonlayer}
	\begin{pgfonlayer}{edgelayer}
		\draw [style=downArrow] (1.center) to (3);
	\end{pgfonlayer}
\end{tikzpicture}}
\endpgfgraphicnamed}\]
\caption{Functionals on $V$.}\label{fig:DiagramFunctional}
\end{subfigure}
\caption{Vectors and linear functionals.}\label{fig:DiagramsVectorsFunctionals}
\end{figure}

Vectors $v\in V$ are in bijection with linear maps $k\to V$, so we may think of vectors as a special kind of linear functions. Since we do not draw the ground field at all we may draw vectors as boxes with no output, as in figure \ref{fig:DiagramsVectors}. We orient these boxes as we will need to manipulate diagrams with element-boxes and in doing so it is useful to keep track of which way round your diagram is drawn. Note that a linear functional, i.e. a linear map $V\to k$ is a box with no output, and so we may draw it as in figure \ref{fig:DiagramFunctional}. We draw it with the opposite orientation to illustrate the self-duality of $\fdvect$, where every vector $v\in V$ defines a functional via the inner product $(v,-) : V\to k$ which maps $w \mapsto (v,w)$.

The symmetry of the tensor product, i.e. $V\otimes W \cong W\otimes V$ for any vector spaces $V,W$, is drawn by crossing wires as in figure \ref{fig:DiagramSymmetry}.

\begin{center}
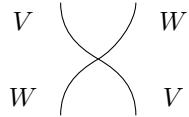

    {%
\beginpgfgraphicnamed{symmetry}
\begin{tikzpicture}
	\begin{pgfonlayer}{nodelayer}
		\node [style=none] (0) at (-0.5, 0.75) {};
		\node [style=none] (1) at (0.5, 0.75) {};
		\node [style=none] (2) at (-0.5, -0.75) {};
		\node [style=none] (3) at (0.5, -0.75) {};
		\node [style=none] (4) at (-1, 0.5) {$V$};
		\node [style=none] (5) at (1, 0.5) {$W$};
		\node [style=none] (6) at (0, 0) {};
		\node [style=none] (7) at (1, -0.5) {$V$};
		\node [style=none] (8) at (-1, -0.5) {$W$};
	\end{pgfonlayer}
	\begin{pgfonlayer}{edgelayer}
		\draw [style=downArrow, in=150, out=-90] (0.center) to (6.center);
		\draw [style=downArrow, in=90, out=-30] (6.center) to (3.center);
		\draw [style=downArrow, in=30, out=-90, looseness=0.75] (1.center) to (6.center);
		\draw [style=downArrow, in=90, out=-150] (6.center) to (2.center);
	\end{pgfonlayer}
\end{tikzpicture}}
\endpgfgraphicnamed}
\captionof{figure}{Symmetry of tensor}
    \label{fig:DiagramSymmetry}
\end{center}


 Inner products are depicted by  cups, as drawn in \ref{fig:DiagramCupsandCaps}. Dually, we have caps, which are maps $k \to V\otimes V$ corresponding to the unit map $1 \mapsto \sum_{i=1}^n v_i \otimes v_i$ where we take $V$ to have basis $\{v_i\}_{i=1,\ldots,n}$.

\begin{center}
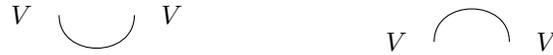

    {%
\beginpgfgraphicnamed{cups}
\begin{tikzpicture}
	\begin{pgfonlayer}{nodelayer}
		\node [style=none] (4) at (0.5, 0) {};
		\node [style=none] (5) at (-0.5, 0) {};
		\node [style=none] (6) at (-1, 0) {$V$};
		\node [style=none] (7) at (1, 0) {$V$};
	\end{pgfonlayer}
	\begin{pgfonlayer}{edgelayer}
		\draw [style=downArrow, bend left=90, looseness=1.50] (4.center) to (5.center);
	\end{pgfonlayer}
\end{tikzpicture}
}
\endpgfgraphicnamed}  \hspace{2cm} {%
\beginpgfgraphicnamed{caps}
\begin{tikzpicture}
	\begin{pgfonlayer}{nodelayer}
		\node [style=none] (0) at (-0.5, 0) {};
		\node [style=none] (1) at (0.5, 0) {};
		\node [style=none] (2) at (-1, 0) {$V$};
		\node [style=none] (3) at (1, 0) {$V$};
	\end{pgfonlayer}
	\begin{pgfonlayer}{edgelayer}
		\draw [style=downArrow, bend left=90, looseness=1.50] (0.center) to (1.center);
	\end{pgfonlayer}
\end{tikzpicture}
}
\endpgfgraphicnamed}
\captionof{figure}{Cups and Caps}
    \label{fig:DiagramCupsandCaps}
\end{center}

These diagrams satisfy the usual string diagrammatic equations, e.g. the most important of which is  the following known as \emph{yanking} \ref{fig:yanking}.

\begin{center}
    {%
\beginpgfgraphicnamed{yanking}
\begin{tikzpicture}
	\begin{pgfonlayer}{nodelayer}
		\node [style=none] (0) at (-3.75, 1) {};
		\node [style=none] (1) at (-3.75, 0) {};
		\node [style=none] (3) at (-2.75, 0) {};
		\node [style=none] (4) at (-1.75, 0) {};
		\node [style=none] (5) at (-1.75, -1) {};
		\node [style=none] (6) at (3.75, 1) {};
		\node [style=none] (7) at (3.75, 0) {};
		\node [style=none] (8) at (2.75, 0) {};
		\node [style=none] (9) at (1.75, 0) {};
		\node [style=none] (10) at (1.75, -1) {};
		\node [style=none] (11) at (0.25, 1) {};
		\node [style=none] (12) at (0.25, -1) {};
		\node [style=none] (13) at (-1, 0) {$=$};
		\node [style=none] (14) at (1, 0) {$=$};
		\node [style=none] (15) at (-0.25, 0) {$V$};
		\node [style=none] (16) at (-4.5, 0.75) {$V$};
		\node [style=none] (17) at (-2.25, -0.75) {$V$};
		\node [style=none] (18) at (2.25, -0.75) {$V$};
		\node [style=none] (19) at (4.25, 0.75) {$V$};
	\end{pgfonlayer}
	\begin{pgfonlayer}{edgelayer}
		\draw [style=downArrow] (0.center) to (1.center);
		\draw [style=downArrow, bend left=90, looseness=1.75] (3.center) to (4.center);
		\draw [style=downArrow] (4.center) to (5.center);
		\draw [style=downArrow, bend right=90, looseness=1.50] (1.center) to (3.center);
		\draw [style=downArrow] (6.center) to (7.center);
		\draw [style=downArrow, bend right=90, looseness=1.75] (8.center) to (9.center);
		\draw [style=downArrow] (9.center) to (10.center);
		\draw [style=downArrow, bend left=90, looseness=1.50] (7.center) to (8.center);
		\draw [style=downArrow] (11.center) to (12.center);
	\end{pgfonlayer}
\end{tikzpicture}}
\endpgfgraphicnamed}  
\captionof{figure}{Yanking}
    \label{fig:yanking}
\end{center}

%


\subsection{New Diagrams for Fock Spaces}

Next we introduce now diagrams  for  our Fock spaces. We depict Fock spaces, i.e. vector spaces of the form $T_{k_0}V$ by bold strings, labelled with a $V$, see figure \ref{fig:DiagramsProjection}. The special  diagrammatic structure we have on Fock spaces is the projection to the $n$-th layer, which is  denoted by the  usual linear map notation, which in this case we label by $p_n$, and call a $p_n$-box, as in figure \ref{fig:DiagramsProjection}.

\begin{center}
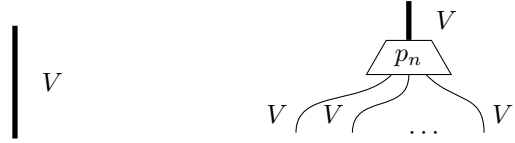

    {%
\beginpgfgraphicnamed{bangObjects}
\begin{tikzpicture}
	\begin{pgfonlayer}{nodelayer}
		\node [style=none] (3) at (0, 0.75) {};
		\node [style=none] (4) at (0, -0.75) {};
		\node [style=none] (5) at (0.5, 0) {$V$};
	\end{pgfonlayer}
	\begin{pgfonlayer}{edgelayer}
		\draw [style=thickArr] (3.center) to (4.center);
	\end{pgfonlayer}
\end{tikzpicture}
}
\endpgfgraphicnamed} \hspace{2cm} {%
\beginpgfgraphicnamed{projection}
\begin{tikzpicture}
	\begin{pgfonlayer}{nodelayer}
		\node [style=none] (0) at (0, 0.5) {};
		\node [style=none] (1) at (-1.5, -1.25) {};
		\node [style=none] (2) at (-0.75, -1.25) {};
		\node [style=none] (3) at (1, -1.25) {};
		\node [style=new style 0] (4) at (0, -0.25) {$p_n$};
		\node [style=none] (5) at (0.2, -1.25) {$\cdots$};
		\node [style=none] (6) at (-1.75, -1) {$V$};
		\node [style=none] (7) at (1.25, -1) {$V$};
		\node [style=none] (8) at (-1, -1) {$V$};
		\node [style=none] (9) at (0.5, 0.25) {$V$};
	\end{pgfonlayer}
	\begin{pgfonlayer}{edgelayer}
		\draw [style=thickArr] (0.center) to (4);
		\draw [style=downArrow, in=90, out=-135] (4) to (1.center);
		\draw [style=downArrow, in=90, out=-90, looseness=1.25] (4) to (2.center);
		\draw [style=downArrow, in=90, out=-45, looseness=1.25] (4) to (3.center);
	\end{pgfonlayer}
\end{tikzpicture}}
\endpgfgraphicnamed}
\captionof{figure}{ Fock  Spaces and Projections from them}
    \label{fig:DiagramsProjection}
\end{center}


Similar to vectors from vector spaces, vectors from Fock spaces $T_{k_0}V$  are depicted with linear maps  $v \colon k \to T_{k_0}V$, see figure \ref{fig:elementFock}. 

\begin{center}
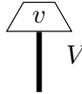

    {%
\beginpgfgraphicnamed{elementFock}

\begin{tikzpicture}
	\begin{pgfonlayer}{nodelayer}
		\node [style=element] (0) at (0, 0.5) {$v$};
		\node [style=none] (1) at (0, -0.5) {};
		\node [style=none] (2) at (0.5, 0) {$V$};
	\end{pgfonlayer}
	\begin{pgfonlayer}{edgelayer}
		\draw [style=thickArr] (0) to (1.center);
	\end{pgfonlayer}
\end{tikzpicture}
}
\endpgfgraphicnamed} 
\captionof{figure}{Elements of Fock Spaces}
    \label{fig:elementFock}
\end{center}

A series of operation often performed on Fock spaces is accessing an element via the above linear map, followed by a projection to the $n$-th layer, see \ref{fig:elementProjection}.

\begin{center}
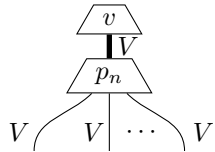

    {%
\beginpgfgraphicnamed{elementProjectionFock}
\begin{tikzpicture}
	\begin{pgfonlayer}{nodelayer}
		\node [style=element] (1) at (2.25, 0.25) {$v$};
		\node [style=new style 0] (2) at (2.25, -0.5) {$p_n$};
		\node [style=none] (3) at (1.25, -1.5) {};
		\node [style=none] (4) at (2.25, -1.5) {};
		\node [style=none] (5) at (2.675, -1.25) {$\cdots$};
		\node [style=none] (6) at (3.25, -1.5) {};
		\node [style=none] (10) at (-0.5, -1.5) {};
		\node [style=none] (12) at (2.5, -0.1) {$V$};
		\node [style=none] (14) at (3.5, -1.25) {$V$};
		\node [style=none] (18) at (1.05, -1.25) {$V$};
		\node [style=none] (19) at (2.05, -1.25) {$V$};
	\end{pgfonlayer}
	\begin{pgfonlayer}{edgelayer}
		\draw [style=downArrow, in=90, out=-135] (2) to (3.center);
		\draw [style=downArrow] (2) to (4.center);
		\draw [style=downArrow, in=90, out=-45] (2) to (6.center);
		\draw [style=thickArr] (1) to (2);
	\end{pgfonlayer}
\end{tikzpicture}}
\endpgfgraphicnamed} 
\captionof{figure}{Accessing an Elements of Fock Spaces followed by a Projection}
    \label{fig:elementProjection}
\end{center}

As an example, here is the diagram of the discourse `\textit{John slept. He snored.}'. 

\begin{center}
\scalebox{0.75}{
 {%
\beginpgfgraphicnamed{johnSleepsHeSnores}
\begin{tikzpicture}
	\begin{pgfonlayer}{nodelayer}
		\node [style=element] (0) at (-3.75, 2.25) {John};
		\node [style=element] (1) at (-1.75, 2.25) {slept};
		\node [style=element] (3) at (0, 2.25) {He};
		\node [style=element] (4) at (1.75, 2.25) {snored};
		\node [style=none] (6) at (-3.75, 2) {};
		\node [style=none] (7) at (-1.5, 2) {};
		\node [style=none] (9) at (0.25, 2) {};
		\node [style=none] (10) at (1.5, 2) {};
		\node [style=none] (11) at (2, 2) {};
		\node [style=none] (13) at (-2, 2) {};
		\node [style=none] (15) at (-0.25, 2) {};
		\node [style=none] (20) at (-1.5, -2.75) {};
		\node [style=none] (22) at (0.25, 1.25) {};
		\node [style=none] (23) at (1.5, 1.25) {};
		\node [style=none] (24) at (2, -2.75) {};
		\node [style=none] (26) at (-2, 0.25) {};
		\node [style=none] (28) at (-0.25, 0.25) {};
		\node [style=element] (31) at (-3.75, 1.25) {$p_2$};
		\node [style=none] (32) at (-4.25, 0.25) {};
		\node [style=none] (34) at (-4.5, 1.75) {$N$};
		\node [style=none] (36) at (-2.5, 1.5) {$N$};
		\node [style=none] (38) at (2.5, -2.25) {$S$};
		\node [style=none] (39) at (-1, -2.25) {$S$};
		\node [style=none] (40) at (-0.75, 1.5) {$N$};
		\node [style=none] (41) at (0.5, 1.5) {$N$};
		\node [style=none] (42) at (1.25, 1.5) {$N$};
		\node [style=none] (45) at (-4.75, 0.25) {$N$};
		\node [style=none] (46) at (-3, 0.75) {$N$};
		\node [style=none] (47) at (-2.75, -0.5) {$ N$};
		\node [style=none] (48) at (-4, 1) {};
		\node [style=none] (49) at (-3.5, 1) {};
		\node [style=none] (50) at (-3.25, 0.25) {};
	\end{pgfonlayer}
	\begin{pgfonlayer}{edgelayer}
		\draw [style=downArrow] (26.center) to (13.center);
		\draw [style=downArrow] (7.center) to (20.center);
		\draw [style=downArrow] (28.center) to (15.center);
		\draw [style=downArrow] (9.center) to (22.center);
		\draw [style=downArrow, bend right=90, looseness=1.25] (22.center) to (23.center);
		\draw [style=downArrow] (23.center) to (10.center);
		\draw [style=downArrow] (11.center) to (24.center);
		\draw [style=thickArr] (6.center) to (31);
		\draw [style=downArrow, bend right=90] (32.center) to (28.center);
		\draw [style=downArrow, in=90, out=-90, looseness=1.50] (48.center) to (32.center);
		\draw [style=downArrow, in=90, out=-90, looseness=1.25] (49.center) to (50.center);
		\draw [style=downArrow, bend right=90, looseness=1.25] (50.center) to (26.center);
	\end{pgfonlayer}
\end{tikzpicture}}
\endpgfgraphicnamed}} 
  \captionof{figure}{}
    \label{fig:johnSleepsHeSnores}
\end{center}

\section{Translating  String Diagrams to Quantum Circuits}

In \cite{QnlpInPractice}  the process of representing a sentence as a quantum circuit is described in four steps. We apply these steps to the process of representing a discourse as a quantum circuit, where they  become as follows:

\begin{enumerate}
\item A  discourse is parsed into a proof in $\sllm$. 
\item The proof tree is turned into a string diagram. 
\item The string diagram is simplified according to  the rewrite rules of Lambeq, where we use an existing rule to replace the verb \emph{to be} with a cap and introduce a new rule which also replaces the pronoun with a cap. 
\item The simplified diagram is then normalised, again using Lambeq, where cups are removed, wires are stretched and boxes are rearranged; all of these are for the purpose of a faster execution on the IBMQ simulator and backends. 
\item The resulting simplified and normalised diagrams are transformed into a  quantum circuit based on a specific parameterisation scheme and choice of ansätze\footnote{A map that determines choices such as the number of qubits that every wire of a string diagram is associated with and the concrete parameterised quantum states that correspond to each word. For our experiments, we represented the noun wires and sentence wires by a one-qubit system.}.
\end{enumerate}

We elaborate on this process. The translation between string diagrams and quantum circuits is summarised in figure \ref{fig:transform}. Here, a triangle with a $0$ in it, is a 0-state qubit. A box with an $H$ in it is a Hadamard gate. We have only two operations that can act on a $0$-state qubit after a Hadamard operation. One of them is a CNOT gate which is depicted with a dot connected to an $\oplus$. The other one is a controlled-Z-rotation on an angle, depicted as a box labelled as $R_{\alpha}(\theta_i)$ connected to a control qubit, where $\alpha$ can be $x,y$ or $x$ and $\theta$ any angle from $0$ to $2\pi$. These are the native gates of the IBMQ quantum computers. 

\begin{center}
\includegraphics[width=12cm]{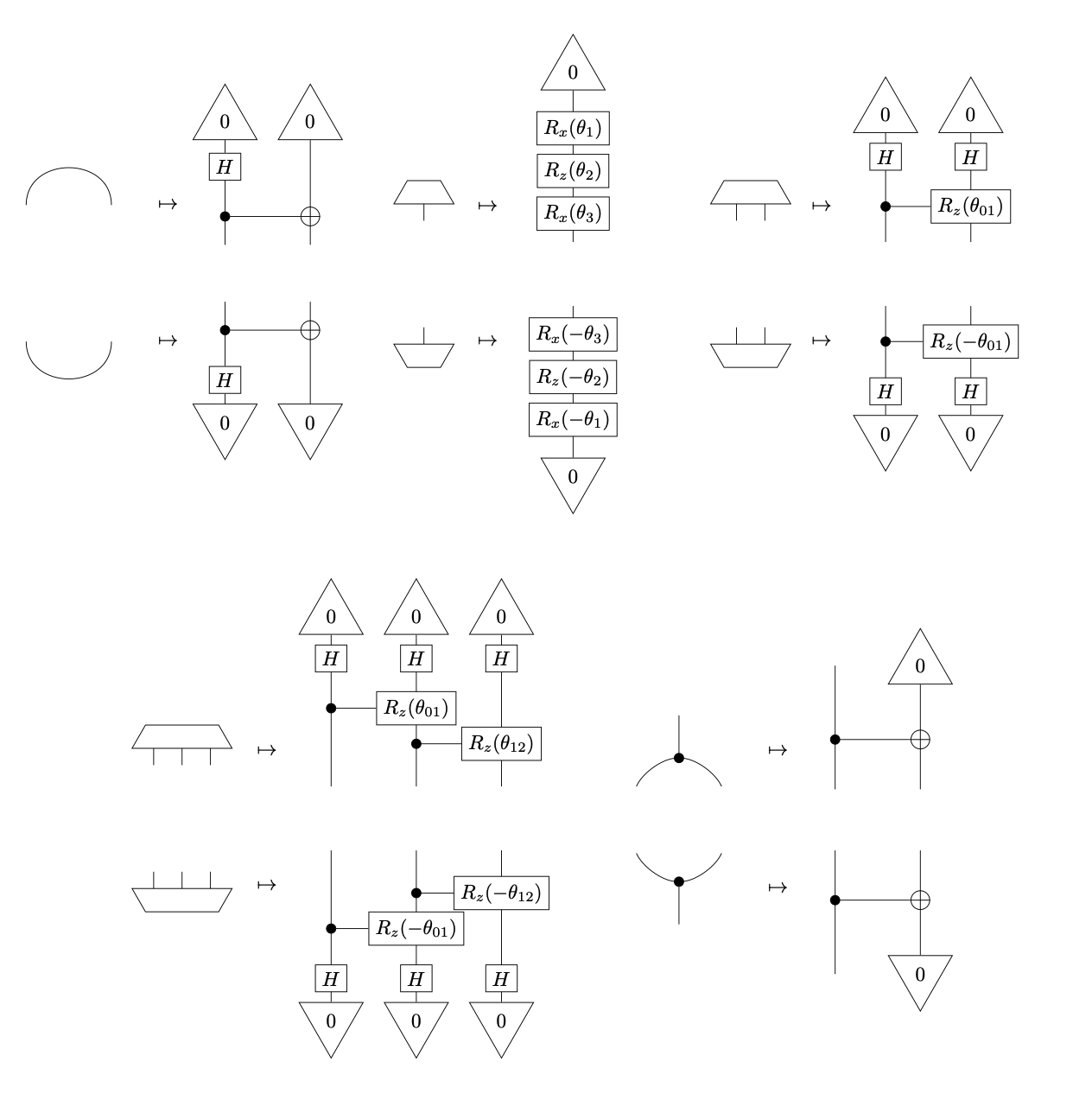}
\captionof{figure}{From String Diagrams to Quantum Circuits}
    \label{fig:transform}
\end{center}

The above steps only operate at the  sentence level. We  need Fock spaces to model the between-sentences discourse relations. Fock spaces do not have counterparts in the quantum circuits implemented by the IBMQ range.  We observe that in the string diagrams of  the discourse phenomena,  Fock spaces are  \emph{only} manipulated using a series of two operations: an access to an element of a Fock space, followed by a projection to an $n$-th layer. We replace this sequence of operations by an order $n$ tensor. This is depicted in Figure \ref{fig:shorthand}. Formally speaking, by doing so, we are restricting ourselves to only the $n$-th layer of a Fock space, which is nothing but an order $n$ tensor, for which we have a quantum circuit counterpart in IBMQ. 

\begin{center}
 {%
\beginpgfgraphicnamed{bangVector}
\begin{tikzpicture}
	\begin{pgfonlayer}{nodelayer}
		\node [style=element] (0) at (-1.5, 0.25) {$v$};
		\node [style=element] (1) at (2.25, 0.25) {$v$};
		\node [style=new style 0] (2) at (2.25, -0.5) {$p_n$};
		\node [style=none] (3) at (1.25, -1.5) {};
		\node [style=none] (4) at (2.25, -1.5) {};
		\node [style=none] (5) at (2.675, -1.25) {$\cdots$};
		\node [style=none] (6) at (3.25, -1.5) {};
		\node [style=none] (7) at (-2.5, -1.5) {};
		\node [style=none] (8) at (-1.5, -1.5) {};
		\node [style=none] (9) at (-1, -1) {$\cdots$};
		\node [style=none] (10) at (-0.5, -1.5) {};
		\node [style=none] (11) at (0.5, -0.5) {$\Leftarrow$};
		\node [style=none] (12) at (2.5, -0.1) {$V$};
		\node [style=none] (14) at (3.5, -1.25) {$V$};
		\node [style=none] (15) at (-2.75, -1) {$V$};
		\node [style=none] (16) at (-1.75, -1) {$V$};
		\node [style=none] (17) at (-0.25, -1) {$V$};
		\node [style=none] (18) at (1.05, -1.25) {$V$};
		\node [style=none] (19) at (2.05, -1.25) {$V$};
	\end{pgfonlayer}
	\begin{pgfonlayer}{edgelayer}
		\draw [style=downArrow, in=90, out=-105] (0) to (7.center);
		\draw [style=downArrow, in=90, out=-90] (0) to (8.center);
		\draw [style=downArrow, in=90, out=-75] (0) to (10.center);
		\draw [style=downArrow, in=90, out=-135] (2) to (3.center);
		\draw [style=downArrow] (2) to (4.center);
		\draw [style=downArrow, in=90, out=-45] (2) to (6.center);
		\draw [style=thickArr] (1) to (2);
	\end{pgfonlayer}
\end{tikzpicture}}
\endpgfgraphicnamed} 
 
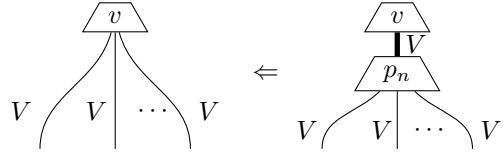
\captionof{figure}{Using the Projection }
    \label{fig:shorthand}
\end{center}

This is mitigated by simplifying the string diagrams.  For instance, a discourse with a subject anaphoric relation such as `\textit{The dog broke the vase. It was clumsy.}', is depicted into the  string diagram of Figure \ref{fig:dogBrokevase} while an object anaphoric relation is depicted as in Figure \ref{fig:fragileVase}.


\begin{center}
\scalebox{0.78}{
 {%
\beginpgfgraphicnamed{dogBrokeVase}
\begin{tikzpicture}
	\begin{pgfonlayer}{nodelayer}
		\node [style=element] (0) at (-6, 2.25) {The dog};
		\node [style=element] (1) at (-4, 2.25) {broke};
		\node [style=element] (2) at (-2, 2.25) {the vase};
		\node [style=element] (3) at (0, 2.25) {It};
		\node [style=element] (4) at (2, 2.25) {was};
		\node [style=element] (5) at (3.75, 2.25) {clumsy};
		\node [style=none] (6) at (-6, 2) {};
		\node [style=none] (7) at (-4, 2) {};
		\node [style=none] (8) at (-2, 2) {};
		\node [style=none] (9) at (0.25, 2) {};
		\node [style=none] (10) at (1.75, 2) {};
		\node [style=none] (11) at (2, 2) {};
		\node [style=none] (12) at (2.25, 2) {};
		\node [style=none] (13) at (-4.25, 2) {};
		\node [style=none] (14) at (-3.75, 2) {};
		\node [style=none] (15) at (-0.25, 2) {};
		\node [style=none] (17) at (3.75, 2) {};
		\node [style=none] (20) at (-4, -2.75) {};
		\node [style=none] (21) at (-2, 0.5) {};
		\node [style=none] (22) at (0.25, 1.25) {};
		\node [style=none] (23) at (1.75, 1.25) {};
		\node [style=none] (24) at (2, -2.75) {};
		\node [style=none] (25) at (2.25, 1.25) {};
		\node [style=none] (26) at (-4.25, 0.25) {};
		\node [style=none] (27) at (-3.75, 0.5) {};
		\node [style=none] (28) at (-0.25, 0.25) {};
		\node [style=none] (30) at (3.75, 1.25) {};
		\node [style=element] (31) at (-6, 1.25) {$p_2$};
		\node [style=none] (32) at (-6.5, 0.25) {};
		\node [style=none] (34) at (-6.75, 1.75) {$N$};
		\node [style=none] (35) at (-1.75, 1.5) {$N$};
		\node [style=none] (36) at (-4.75, 1.5) {$N$};
		\node [style=none] (37) at (-3.25, 1.5) {$N$};
		\node [style=none] (38) at (2.5, -2.25) {$S$};
		\node [style=none] (39) at (-3.5, -2.25) {$S$};
		\node [style=none] (40) at (-0.75, 1.5) {$N$};
		\node [style=none] (41) at (0.5, 1.5) {$N$};
		\node [style=none] (42) at (1.5, 1.5) {$N$};
		\node [style=none] (43) at (4.75, 1.5) {$N/N$};
		\node [style=none] (44) at (3, 1.5) {$N/N$};
		\node [style=none] (45) at (-7, 0.25) {$N$};
		\node [style=none] (46) at (-5.25, 0.75) {$N$};
		\node [style=none] (47) at (-5, -0.5) {$ N$};
		\node [style=none] (48) at (-6.25, 1) {};
		\node [style=none] (49) at (-5.75, 1) {};
		\node [style=none] (50) at (-5.5, 0.25) {};
	\end{pgfonlayer}
	\begin{pgfonlayer}{edgelayer}
		\draw [style=downArrow] (26.center) to (13.center);
		\draw [style=downArrow] (27.center) to (14.center);
		\draw [style=downArrow] (8.center) to (21.center);
		\draw [style=downArrow, bend left=90, looseness=1.25] (21.center) to (27.center);
		\draw [style=downArrow] (7.center) to (20.center);
		\draw [style=downArrow] (28.center) to (15.center);
		\draw [style=downArrow] (9.center) to (22.center);
		\draw [style=downArrow, bend right=90, looseness=1.25] (22.center) to (23.center);
		\draw [style=downArrow] (23.center) to (10.center);
		\draw [style=downArrow] (11.center) to (24.center);
		\draw [style=downArrow] (17.center) to (30.center);
		\draw [style=downArrow] (25.center) to (12.center);
		\draw [style=downArrow, bend left=90, looseness=1.50] (30.center) to (25.center);
		\draw [style=thickArr] (6.center) to (31);
		\draw [style=downArrow, bend right=90] (32.center) to (28.center);
		\draw [style=downArrow, in=90, out=-90, looseness=1.50] (48.center) to (32.center);
		\draw [style=downArrow, in=90, out=-90, looseness=1.25] (49.center) to (50.center);
		\draw [style=downArrow, bend right=90, looseness=1.25] (50.center) to (26.center);
	\end{pgfonlayer}
\end{tikzpicture}}
\endpgfgraphicnamed}} 
  \captionof{figure}{A discourse with a subject anaphoric relation: the pronoun `it' in the second sentence refers to the subject `dog' of the first sentence.}
    \label{fig:dogBrokevase}
\end{center}

\begin{center}
\scalebox{0.8}{
{%
\beginpgfgraphicnamed{catBrokeGlass}
\begin{tikzpicture}
	\begin{pgfonlayer}{nodelayer}
		\node [style=element] (0) at (-1.75, 2.5) {The cat};
		\node [style=element] (1) at (0, 2.5) {broke};
		\node [style=element] (2) at (1.75, 2.5) {the glass};
		\node [style=element] (3) at (3.25, 2.5) {It};
		\node [style=element] (4) at (4.75, 2.5) {was};
		\node [style=element] (5) at (6.25, 2.5) {fragile};
		\node [style=none] (6) at (1.75, 2.25) {};
		\node [style=none] (7) at (0, 2.25) {};
		\node [style=none] (8) at (-1.75, 2.25) {};
		\node [style=none] (9) at (3.5, 2.25) {};
		\node [style=none] (10) at (4.5, 2.25) {};
		\node [style=none] (11) at (4.75, 2.25) {};
		\node [style=none] (12) at (5, 2.25) {};
		\node [style=none] (13) at (-0.25, 2.25) {};
		\node [style=none] (14) at (0.25, 2.25) {};
		\node [style=none] (15) at (3, 2.25) {};
		\node [style=none] (16) at (6.25, 2.25) {};
		\node [style=none] (17) at (0, -0.75) {};
		\node [style=none] (18) at (-1.75, 1.25) {};
		\node [style=none] (19) at (3.5, 1.5) {};
		\node [style=none] (20) at (4.5, 1.5) {};
		\node [style=none] (21) at (4.75, -0.75) {};
		\node [style=none] (22) at (5, 1.5) {};
		\node [style=none] (23) at (-0.25, 1.25) {};
		\node [style=none] (24) at (0.25, 0.5) {};
		\node [style=none] (25) at (3, 0.5) {};
		\node [style=none] (26) at (6.25, 1.5) {};
		\node [style=element] (27) at (1.75, 1.5) {$p_2$};
		\node [style=none] (28) at (1.25, 0.5) {};
		\node [style=none] (29) at (1.5, 2) {$N$};
		\node [style=none] (30) at (-1.5, 1.75) {$N$};
		\node [style=none] (31) at (-0.5, 1.75) {$N$};
		\node [style=none] (32) at (0.5, 1.75) {$N$};
		\node [style=none] (33) at (5, -0.25) {$S$};
		\node [style=none] (34) at (0.25, -0.25) {$S$};
		\node [style=none] (35) at (2.75, 1.75) {$N$};
		\node [style=none] (36) at (3.75, 1.75) {$N$};
		\node [style=none] (37) at (4.25, 1.75) {$N$};
		\node [style=none] (38) at (7, 1.75) {$N/N$};
		\node [style=none] (39) at (5.5, 1.75) {$N/N$};
		\node [style=none] (40) at (1, 0.5) {$N$};
		\node [style=none] (42) at (2.5, 0.5) {$ N$};
		\node [style=none] (43) at (1.5, 1.25) {};
		\node [style=none] (44) at (2, 1.25) {};
		\node [style=none] (45) at (2.25, 0.5) {};
	\end{pgfonlayer}
	\begin{pgfonlayer}{edgelayer}
		\draw [style=downArrow] (23.center) to (13.center);
		\draw [style=downArrow] (24.center) to (14.center);
		\draw [style=downArrow] (8.center) to (18.center);
		\draw [style=downArrow] (7.center) to (17.center);
		\draw [style=downArrow] (25.center) to (15.center);
		\draw [style=downArrow] (9.center) to (19.center);
		\draw [style=downArrow, bend right=90, looseness=1.25] (19.center) to (20.center);
		\draw [style=downArrow] (20.center) to (10.center);
		\draw [style=downArrow] (11.center) to (21.center);
		\draw [style=downArrow] (16.center) to (26.center);
		\draw [style=downArrow] (22.center) to (12.center);
		\draw [style=downArrow, bend left=90, looseness=1.50] (26.center) to (22.center);
		\draw [style=thickArr] (6.center) to (27);
		\draw [style=downArrow, in=90, out=-90, looseness=1.50] (43.center) to (28.center);
		\draw [style=downArrow, in=90, out=-90, looseness=1.25] (44.center) to (45.center);
		\draw [bend left=90, looseness=1.50] (28.center) to (24.center);
		\draw [bend right=90, looseness=1.75] (45.center) to (25.center);
		\draw [bend right=90, looseness=1.25] (18.center) to (23.center);
	\end{pgfonlayer}
\end{tikzpicture}}
\endpgfgraphicnamed}}

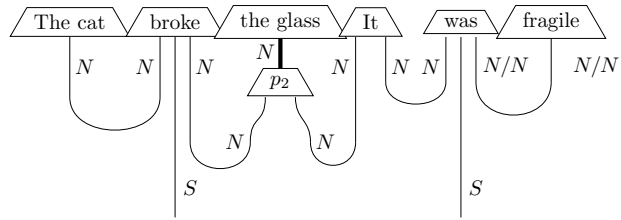
\captionof{figure}{A discourse with an object anaphoric relation: the pronoun `it' in the second sentence refers to the object `glass' of the first sentence.}
\label{fig:fragileVase}
\end{center}

The diagrams in Figures \ref{fig:dogBrokevase} and \ref{fig:fragileVase} are then simplified and normalised  into the diagrams in Figure \ref{fig:dogBrokevaseSimplified} and Figure \ref{fig:catBrokeGlassSimplified} 
\footnote{Although there are still cups in Figure \ref{fig:catBrokeGlassSimplified}, this is the most optimised version of the diagram. Removing the remaining two cups,  stretching the wires, and rearranging the boxes will result in Figure \ref{fig:catBrokeGlass_v2}, which not only has a cap and a cup, but also a crossing, thus is far from optimised.}. 

\begin{figure}
    \centering
    \begin{minipage}{.5\textwidth}
        \centering
        \scalebox{0.45}{
 {%
\beginpgfgraphicnamed{dogBrokevaseSimplified}
\begin{tikzpicture}
	\begin{pgfonlayer}{nodelayer}
		\node [style=none] (0) at (3.5, -28.25) {};
		\node [style=none] (1) at (5.25, -28.25) {};
		\node [style=none] (2) at (6, -29.25) {};
		\node [style=none] (3) at (2.75, -29.25) {};
		\node [style=none] (4) at (4.5, -28.75) {\huge broke};
		\node [style=none] (5) at (2.25, -26.75) {};
		\node [style=none] (6) at (8.5, -26.75) {};
		\node [style=none] (7) at (9, -27.75) {};
		\node [style=none] (8) at (1.5, -27.75) {};
		\node [style=none] (9) at (5.25, -27.25) {\huge clumsy};
		\node [style=none] (10) at (4.75, -30.75) {};
		\node [style=none] (11) at (7.75, -30.75) {};
		\node [style=none] (12) at (7, -31.75) {};
		\node [style=none] (13) at (5.5, -31.75) {};
		\node [style=none] (14) at (6.25, -31.25) {\huge vase};
		\node [style=none] (17) at (0.75, -30.75) {};
		\node [style=none] (18) at (3.75, -30.75) {};
		\node [style=none] (19) at (3, -31.75) {};
		\node [style=none] (20) at (1.5, -31.75) {};
		\node [style=none] (21) at (2.25, -31.25) {\huge dog};
		\node [style=none] (24) at (4.25, -29.25) {};
		\node [style=none] (25) at (4.25, -33.25) {};
		\node [style=none] (26) at (5.25, -29.25) {};
		\node [style=none] (27) at (5.25, -30.75) {};
		\node [style=none] (28) at (3.25, -29.25) {};
		\node [style=none] (29) at (3.25, -30.75) {};
		\node [style=none] (30) at (2.25, -27.75) {};
		\node [style=none] (31) at (8.25, -27.75) {};
		\node [style=none] (32) at (2.25, -30.75) {};
		\node [style=none] (33) at (8.25, -33.25) {};
		\node [style=none] (73) at (1.75, -28.25) {\huge $N$};
		\node [style=none] (74) at (9, -32.75) {\huge $S$};
		\node [style=none] (75) at (2.75, -30) {\huge $N$};
		\node [style=none] (76) at (5, -32.75) {\huge $S$};
		\node [style=none] (77) at (5.75, -30) {\huge $N$};
	\end{pgfonlayer}
	\begin{pgfonlayer}{edgelayer}
		\draw [line width=1pt] (0.center) to (3.center);
		\draw [line width=1pt] (3.center) to (2.center);
		\draw [line width=1pt] (2.center) to (1.center);
		\draw [line width=1pt] (1.center) to (0.center);
		\draw [line width=1pt] (5.center) to (8.center);
		\draw [line width=1pt] (8.center) to (7.center);
		\draw [line width=1pt] (7.center) to (6.center);
		\draw [line width=1pt] (6.center) to (5.center);
		\draw [line width=1pt] (10.center) to (13.center);
		\draw [line width=1pt] (13.center) to (12.center);
		\draw [line width=1pt] (12.center) to (11.center);
		\draw [line width=1pt] (11.center) to (10.center);
		\draw [line width=1pt] (17.center) to (20.center);
		\draw [line width=1pt] (20.center) to (19.center);
		\draw [line width=1pt] (19.center) to (18.center);
		\draw [line width=1pt] (18.center) to (17.center);
		\draw [line width=1pt] (24.center) to (25.center);
		\draw [line width=1pt] (26.center) to (27.center);
		\draw [line width=1pt] (28.center) to (29.center);
		\draw [line width=1pt] (30.center) to (32.center);
		\draw [line width=1pt] (31.center) to (33.center);
	\end{pgfonlayer}
\end{tikzpicture}}
\endpgfgraphicnamed}} 
        \caption{}
        \label{fig:dogBrokevaseSimplified}
    \end{minipage}%
    \begin{minipage}{0.5\textwidth}
        \centering
        \scalebox{0.45}{
        {%
\beginpgfgraphicnamed{catBrokeGlassSimplified}
\begin{tikzpicture}
	\begin{pgfonlayer}{nodelayer}
		\node [style=none] (78) at (12.75, -27) {};
		\node [style=none] (79) at (14.75, -27) {};
		\node [style=none] (80) at (15.25, -28) {};
		\node [style=none] (81) at (12.25, -28) {};
		\node [style=none] (82) at (13.75, -27.5) {\huge broke};
		\node [style=none] (83) at (10, -29.5) {};
		\node [style=none] (84) at (13.5, -29.5) {};
		\node [style=none] (85) at (13, -30.5) {};
		\node [style=none] (86) at (10.5, -30.5) {};
		\node [style=none] (87) at (11.75, -30) {\huge cat};
		\node [style=none] (88) at (13.75, -28) {};
		\node [style=none] (89) at (13.75, -33) {};
		\node [style=none] (90) at (14.5, -28) {};
		\node [style=none] (91) at (13, -28) {};
		\node [style=none] (92) at (13, -29.5) {};
		\node [style=none] (93) at (15.5, -29.25) {};
		\node [style=none] (94) at (17.5, -29.25) {};
		\node [style=none] (95) at (18, -30.25) {};
		\node [style=none] (96) at (15, -30.25) {};
		\node [style=none] (97) at (16.5, -29.75) {\huge glass};
		\node [style=none] (98) at (17.5, -30.25) {};
		\node [style=none] (99) at (15.75, -30.25) {};
		\node [style=none] (100) at (18.75, -30.25) {};
		\node [style=none] (101) at (20.75, -30.25) {};
		\node [style=none] (102) at (21.25, -31.25) {};
		\node [style=none] (103) at (18.25, -31.25) {};
		\node [style=none] (104) at (19.75, -30.75) {\huge fragile};
		\node [style=none] (105) at (20.5, -31.25) {};
		\node [style=none] (106) at (18.75, -31.25) {};
		\node [style=none] (107) at (14.5, -30.25) {};
		\node [style=none] (108) at (17.5, -31.25) {};
		\node [style=none] (109) at (12.5, -28.75) {\huge $N$};
		\node [style=none] (110) at (15, -28.75) {\huge $N$};
		\node [style=none] (111) at (14.25, -32.5) {\huge $S$};
		\node [style=none] (112) at (16.25, -30.75) {\huge $N$};
		\node [style=none] (113) at (17, -30.75) {\huge $N$};
		\node [style=none] (114) at (19.25, -31.75) {\huge $N$};
		\node [style=none] (115) at (20.5, -33) {};
		\node [style=none] (116) at (21, -32.5) {\huge $S$};
	\end{pgfonlayer}
	\begin{pgfonlayer}{edgelayer}
		\draw [line width=1pt] [line width=1pt] (78.center) to (81.center);
		\draw [line width=1pt] [line width=1pt] (81.center) to (80.center);
		\draw [line width=1pt] [line width=1pt] (80.center) to (79.center);
		\draw [line width=1pt] [line width=1pt] (79.center) to (78.center);
		\draw [line width=1pt] [line width=1pt] (83.center) to (86.center);
		\draw [line width=1pt] [line width=1pt] (86.center) to (85.center);
		\draw [line width=1pt] [line width=1pt] (85.center) to (84.center);
		\draw [line width=1pt] [line width=1pt] (84.center) to (83.center);
		\draw [line width=1pt] [line width=1pt] (88.center) to (89.center);
		\draw [line width=1pt] [line width=1pt] (91.center) to (92.center);
		\draw [line width=1pt] [line width=1pt] (93.center) to (96.center);
		\draw [line width=1pt] [line width=1pt] (96.center) to (95.center);
		\draw [line width=1pt] [line width=1pt] (95.center) to (94.center);
		\draw [line width=1pt] [line width=1pt] (94.center) to (93.center);
		\draw [line width=1pt] [line width=1pt] (100.center) to (103.center);
		\draw [line width=1pt] [line width=1pt] (103.center) to (102.center);
		\draw [line width=1pt] [line width=1pt] (102.center) to (101.center);
		\draw [line width=1pt] [line width=1pt] (101.center) to (100.center);
		\draw [line width=1pt] [line width=1pt] (90.center) to (107.center);
		\draw [line width=1pt] [line width=1pt] [bend left=90, looseness=2.25] (99.center) to (107.center);
		\draw [line width=1pt] [line width=1pt] (98.center) to (108.center);
		\draw [line width=1pt] [line width=1pt] [bend right=90, looseness=2.00] (108.center) to (106.center);
		\draw [line width=1pt] [line width=1pt] (105.center) to (115.center);
	\end{pgfonlayer}
\end{tikzpicture}}
\endpgfgraphicnamed}} 
        \caption{}
        \label{fig:catBrokeGlassSimplified}
    \end{minipage}
\end{figure}

\begin{center}
\scalebox{0.45}{
{%
\beginpgfgraphicnamed{catBrokeGlass_v2}
\begin{tikzpicture}
	\begin{pgfonlayer}{nodelayer}
		\node [style=none] (0) at (2.75, -27) {};
		\node [style=none] (1) at (8.75, -27) {};
		\node [style=none] (2) at (9.25, -28) {};
		\node [style=none] (3) at (2.25, -28) {};
		\node [style=none] (4) at (6, -27.5) {\huge broke};
		\node [style=none] (5) at (0.5, -29.5) {};
		\node [style=none] (6) at (3.25, -29.5) {};
		\node [style=none] (7) at (2.75, -30.5) {};
		\node [style=none] (8) at (1, -30.5) {};
		\node [style=none] (9) at (1.75, -30) {\huge cat};
		\node [style=none] (12) at (7.75, -28) {};
		\node [style=none] (13) at (2.75, -28) {};
		\node [style=none] (14) at (2.75, -29.5) {};
		\node [style=none] (15) at (5, -29) {};
		\node [style=none] (16) at (7, -29) {};
		\node [style=none] (17) at (7.5, -30) {};
		\node [style=none] (18) at (4.5, -30) {};
		\node [style=none] (19) at (6, -29.5) {\huge fragile};
		\node [style=none] (20) at (6, -30) {};
		\node [style=none] (29) at (7.75, -32) {};
		\node [style=none] (31) at (2.25, -28.5) {\huge $N$};
		\node [style=none] (32) at (9, -28.5) {\huge $N$};
		\node [style=none] (39) at (5.5, -32) {};
		\node [style=none] (40) at (8.5, -32) {};
		\node [style=none] (41) at (8, -33) {};
		\node [style=none] (42) at (6, -33) {};
		\node [style=none] (43) at (7, -32.5) {\huge glass};
		\node [style=none] (44) at (6, -32) {};
		\node [style=none] (45) at (6.75, -30) {};
		\node [style=none] (46) at (6.75, -31) {};
		\node [style=none] (47) at (8.25, -31) {};
		\node [style=none] (48) at (8.25, -29.5) {};
		\node [style=none] (49) at (9.25, -29.5) {};
		\node [style=none] (50) at (9.25, -33) {};
		\node [style=none] (51) at (5.5, -31.5) {\huge $N$};
		\node [style=none] (52) at (7.25, -30.75) {\huge $S$};
		\node [style=none] (53) at (9.75, -32.75) {\huge $S$};
		\node [style=none] (54) at (4, -33) {};
		\node [style=none] (55) at (4, -28) {};
		\node [style=none] (56) at (4.5, -32.75) {\huge $S$};
	\end{pgfonlayer}
	\begin{pgfonlayer}{edgelayer}
		\draw [line width=1pt]  (0.center) to (3.center);
		\draw [line width=1pt]  (3.center) to (2.center);
		\draw [line width=1pt]  (2.center) to (1.center);
		\draw [line width=1pt]  (1.center) to (0.center);
		\draw [line width=1pt]  (5.center) to (8.center);
		\draw [line width=1pt]  (8.center) to (7.center);
		\draw [line width=1pt]  (7.center) to (6.center);
		\draw [line width=1pt]  (6.center) to (5.center);
		\draw [line width=1pt]  (13.center) to (14.center);
		\draw [line width=1pt]  (15.center) to (18.center);
		\draw [line width=1pt]  (18.center) to (17.center);
		\draw [line width=1pt]  (17.center) to (16.center);
		\draw [line width=1pt]  (16.center) to (15.center);
		\draw [line width=1pt]  (12.center) to (29.center);
		\draw [line width=1pt]  (39.center) to (42.center);
		\draw [line width=1pt]  (42.center) to (41.center);
		\draw [line width=1pt]  (41.center) to (40.center);
		\draw [line width=1pt]  (40.center) to (39.center);
		\draw [line width=1pt]  (20.center) to (44.center);
		\draw [line width=1pt]  (45.center) to (46.center);
		\draw [line width=1pt]  [bend right=90] (46.center) to (47.center);
		\draw [line width=1pt]  (47.center) to (48.center);
		\draw [line width=1pt] [bend left=90, looseness=0.75] (48.center) to (49.center);
		\draw [line width=1pt]  (49.center) to (50.center);
		\draw [line width=1pt]  (55.center) to (54.center);
	\end{pgfonlayer}
\end{tikzpicture}}
\endpgfgraphicnamed}}
\captionof{figure}{}
\label{fig:catBrokeGlass_v2}
\end{center}


\noindent
The simplified and normalised diagrams are  then translated into the circuits in Figure \ref{fig:subjectCircuit} and in Figure \ref{fig:objectCircuit} respectively.

\hspace{-1cm}\begin{minipage}{4cm}
\scalebox{0.28}{
{%
\beginpgfgraphicnamed{subjectCircuit}
\InputIfFileExists{subjectCircuit.tikz}{}{\input{./tikz/subjectCircuit.tikz}}
\endpgfgraphicnamed}}
\captionof{figure}{}
\label{fig:subjectCircuit}
\end{minipage}
\hspace{2cm}
\begin{minipage}{4cm}
\scalebox{0.28}{
{%
\beginpgfgraphicnamed{objectCircuit}
\InputIfFileExists{objectCircuit.tikz}{}{\input{./tikz/objectCircuit.tikz}}
\endpgfgraphicnamed}}
\captionof{figure}{}
\label{fig:objectCircuit}
\end{minipage}

\section{Experiments}
We train binary classification models to predict whether a pronoun refers to a subject or an object of a sentence. This training is a classical-quantum hybrid training scheme where a quantum computer is responsible for computing the meaning of the sentence by connecting the quantum states in a quantum circuit. A classical computer is used to compute the loss function of the training. In each iteration, a new set of quantum states is computed based on the loss function of the previous iteration. We use DisCoPy\cite{de_Felice_2021} to produce diagrams and circuits and lambeq\cite{lambeq_paper} to train our models. The code and the dataset are publicly available here\footnote{\url{https://github.com/hwazni/anaphora}}\footnote{\url{https://github.com/kinianlo/discopro/}}.

\subsection{Dataset}
We use the BERT\footnote{\url{https://huggingface.co/bert-large-cased-whole-word-masking}} large model with whole word masking to generate our dataset. This model uses the context words surrounding a mask token to predict what the masked word should be. We use it to to generate the words of the entries of our dataset. 

Each entry of the dataset contains a pair of sentences $(S_1, S_2)$. The first sentence $S_1$  is of the form  \textit{Subject1 Verb1 Object} and the second sentence $S_2$  is of the form \textit{Subject2 Verb2 Adjective}.  The first sentence contains two candidate referents (\textit{Subject1, Object}) and the second sentence contains one referring pronoun (\textit{Subject2}). The  pronoun agrees in gender, number, and semantic class with both referents and can in principle refer to either, but only one of the two reference relations is correct.  If the referent relation is between the pronoun and \textit{Subject1}, we have a case of \emph{subject anaphora} and if it is between the pronoun and the  \textit{Object}, it is \emph{object anaphora}. These two cases are the two classes of our binary classification task. 

We start with an initial sentence \textit{The girls ate the apples. They were hungry.} from the definite pronoun resolution dataset of \cite{rahman-ng-2012-resolving}. We first mask the subject (\textit{girls}) and let the model predict new subjects (\textit{men, children}).  We then mask the object (\textit{apples}) and let the model predict new objects (\textit{cookies, pancakes}). Next, we mask the  verbs \textit{ate, were} and the adjective (\textit{hungry})  and apply a similar process to these to obtain more verbs and adjectives. 

The final dataset has a vocabulary of size 16, It consists of 144 pairs of sentences. We divide these to 72 pairs for training, 36 for testing, and 36 for validation.  Sentences with subject anaphora are labeled 0, while sentences with object anaphora are labeled 1.  The dataset is  balanced with respect to the two classes. A snapshot of the entries of the dataset is presented in Table \ref{tab:table2}.

\begin{table}[]
\centering
\begin{tabular}{@{}clll@{}}
\toprule
Sentence                                            & referent & pronoun & label \\ \midrule
The girls ate the cookies. They looked hungry.      & girls    & they    & 0     \\
The men enjoyed the pancakes. They were tasty.      & pancakes & they    & 1     \\
The children loved the cookies. They were starving. & children & they    & 0     \\
\multicolumn{1}{l}{The girls enjoyed the pancakes. They looked delicious.} & pancakes & they & 1 \\ \bottomrule
\end{tabular}
\caption{\label{tab:table2} An example of sentences from the dataset}
\end{table}

\subsection{Preprocessing for binary classification}
 
The two sentences $(S_1, S_2)$ of each entry of the dataset are combined together to create a single output quantum state. 
This single state will be used as the input to our binary classifier.   
In principle, this can be any quantum map that takes two sentences as inputs and gives a sentence as the output.  
We experimented with two choices for this quantum map:
\begin{enumerate}
\item A spider, implemented as a CNOT gate on a quantum circuit, which encodes a commutative Frobenius multiplication. The resulting combined quantum circuit is denoted by $S_1 \odot S_2$. 
\item A general quantum gate parameterised by the IQP ansatz which contains a controlled-Z rotation whose angle is learnt in the training. The resulting combined quantum circuit is denoted by $S_1 R_Z S_2$. 
\end{enumerate}
When there is no need to specify the combination operation, we use the general symbol $\bigcirc$ to denote either.   

In the language of string diagrams, the Frobenius multiplication is  depicted by a black circle, see for example \cite{Sadretal2013Frob,Sadrzadeh2016,Moortgatetal2019,Wijnholds2018}. The controlled-Z rotation is a depicted by a box, which we label as $R_Z$, see Figures \ref{fig:exampleWithSpider} and \ref{fig:exampleWithBox}.

\begin{center}
    \begin{minipage}{.5\textwidth}
        \centering
        \scalebox{0.45}{
 {%
\beginpgfgraphicnamed{exampleWithSpider}
\begin{tikzpicture}
	\begin{pgfonlayer}{nodelayer}
		\node [style=none] (0) at (12.25, -25.75) {};
		\node [style=none] (1) at (14.25, -25.75) {};
		\node [style=none] (2) at (14.75, -26.75) {};
		\node [style=none] (3) at (11.75, -26.75) {};
		\node [style=none] (4) at (13.25, -26.25) {\huge ate};
		\node [style=none] (5) at (11, -24.25) {};
		\node [style=none] (6) at (17.5, -24.25) {};
		\node [style=none] (7) at (18, -25.25) {};
		\node [style=none] (8) at (10.5, -25.25) {};
		\node [style=none] (9) at (14.25, -24.75) {\huge hungry};
		\node [style=none] (10) at (13.5, -28.25) {};
		\node [style=none] (11) at (17, -28.25) {};
		\node [style=none] (12) at (16.5, -29.25) {};
		\node [style=none] (13) at (14, -29.25) {};
		\node [style=none] (14) at (15.25, -28.75) {\huge cookies};
		\node [style=none] (15) at (15.5, -29.25) {};
		\node [style=none] (16) at (14.75, -29.25) {};
		\node [style=none] (17) at (9.5, -28.25) {};
		\node [style=none] (18) at (13, -28.25) {};
		\node [style=none] (19) at (12.5, -29.25) {};
		\node [style=none] (20) at (10, -29.25) {};
		\node [style=none] (21) at (11.25, -28.75) {\huge children};
		\node [style=none] (22) at (11.5, -29.25) {};
		\node [style=none] (23) at (10.75, -29.25) {};
		\node [style=none] (24) at (13.25, -26.75) {};
		\node [style=none] (25) at (13.25, -30) {};
		\node [style=none] (26) at (14, -26.75) {};
		\node [style=none] (27) at (14, -28.25) {};
		\node [style=none] (28) at (12.5, -26.75) {};
		\node [style=none] (29) at (12.5, -28.25) {};
		\node [style=none] (30) at (11.25, -25.25) {};
		\node [style=none] (31) at (17.5, -25.25) {};
		\node [style=none] (32) at (11.25, -28.25) {};
		\node [style=none] (33) at (17.5, -30) {};
		\node [style=none] (34) at (10.75, -25.75) {\huge $N$};
		\node [style=none] (35) at (18, -25.75) {\huge $S$};
		\node [style=none] (36) at (12, -27.25) {\huge $N$};
		\node [style=none] (37) at (13.5, -27.25) {\huge $S$};
		\node [style=none] (38) at (14.5, -27.25) {\huge $N$};
		\node [style={fill=black, draw=black, shape=circle}] (86) at (15.5, -30.75) {};
		\node [style=none] (87) at (15.5, -31.75) {};
		\node [style=none] (88) at (17, -31.8) {\huge $S_1 \odot S_2$};
	\end{pgfonlayer}
	\begin{pgfonlayer}{edgelayer}
		\draw [line width=1pt] (0.center) to (3.center);
		\draw [line width=1pt] (3.center) to (2.center);
		\draw [line width=1pt] (2.center) to (1.center);
		\draw [line width=1pt] (1.center) to (0.center);
		\draw [line width=1pt] (5.center) to (8.center);
		\draw [line width=1pt] (8.center) to (7.center);
		\draw [line width=1pt] (7.center) to (6.center);
		\draw [line width=1pt] (6.center) to (5.center);
		\draw [line width=1pt] (10.center) to (13.center);
		\draw [line width=1pt] (13.center) to (12.center);
		\draw [line width=1pt] (12.center) to (11.center);
		\draw [line width=1pt] (11.center) to (10.center);
		\draw [line width=1pt] (17.center) to (20.center);
		\draw [line width=1pt] (20.center) to (19.center);
		\draw [line width=1pt] (19.center) to (18.center);
		\draw [line width=1pt] (18.center) to (17.center);
		\draw [line width=1pt] (24.center) to (25.center);
		\draw [line width=1pt] (26.center) to (27.center);
		\draw [line width=1pt] (28.center) to (29.center);
		\draw [line width=1pt] (30.center) to (32.center);
		\draw [line width=1pt] (31.center) to (33.center);
		\draw [line width=1pt] [bend right=15] (25.center) to (86);
		\draw [line width=1pt] [bend right=15, looseness=0.75] (86) to (33.center);
		\draw [line width=1pt] (86) to (87.center);
	\end{pgfonlayer}
\end{tikzpicture}}
\endpgfgraphicnamed}} 
        
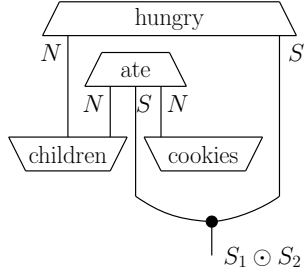
\captionof{figure}{Frobenius combination}
        \label{fig:exampleWithSpider}
    \end{minipage}%
    \begin{minipage}{0.5\textwidth}
        \centering
        \scalebox{0.43}{
        {%
\beginpgfgraphicnamed{exampleWithBox}
\begin{tikzpicture}
	\begin{pgfonlayer}{nodelayer}
		\node [style=none] (39) at (26, -25.75) {};
		\node [style=none] (40) at (28, -25.75) {};
		\node [style=none] (41) at (28.5, -26.75) {};
		\node [style=none] (42) at (25.5, -26.75) {};
		\node [style=none] (43) at (27, -26.25) {\huge ate};
		\node [style=none] (44) at (24.75, -24.25) {};
		\node [style=none] (45) at (31.25, -24.25) {};
		\node [style=none] (46) at (31.75, -25.25) {};
		\node [style=none] (47) at (24.25, -25.25) {};
		\node [style=none] (48) at (28, -24.75) {\huge hungry};
		\node [style=none] (49) at (27.25, -28.25) {};
		\node [style=none] (50) at (30.75, -28.25) {};
		\node [style=none] (51) at (30.25, -29.25) {};
		\node [style=none] (52) at (27.75, -29.25) {};
		\node [style=none] (53) at (29, -28.75) {\huge cookies};
		\node [style=none] (56) at (23.25, -28.25) {};
		\node [style=none] (57) at (26.75, -28.25) {};
		\node [style=none] (58) at (26.25, -29.25) {};
		\node [style=none] (59) at (23.75, -29.25) {};
		\node [style=none] (60) at (25, -28.75) {\huge children};
		\node [style=none] (63) at (27, -26.75) {};
		\node [style=none] (64) at (27, -30) {};
		\node [style=none] (65) at (27.75, -26.75) {};
		\node [style=none] (66) at (27.75, -28.25) {};
		\node [style=none] (67) at (26.25, -26.75) {};
		\node [style=none] (68) at (26.25, -28.25) {};
		\node [style=none] (69) at (25, -25.25) {};
		\node [style=none] (70) at (31.25, -25.25) {};
		\node [style=none] (71) at (25, -28.25) {};
		\node [style=none] (72) at (31.25, -30) {};
		\node [style=none] (73) at (24.5, -25.75) {\huge $N$};
		\node [style=none] (74) at (31.75, -25.75) {\huge $S$};
		\node [style=none] (75) at (25.75, -27.25) {\huge $N$};
		\node [style=none] (76) at (27.25, -27.25) {\huge $S$};
		\node [style=none] (77) at (28.25, -27.25) {\huge $N$};
		\node [style=none] (78) at (26.5, -30) {};
		\node [style=none] (79) at (31.75, -30) {};
		\node [style=none] (80) at (32.25, -31) {};
		\node [style=none] (81) at (26, -31) {};
		\node [style=none] (82) at (29, -30.5) {\huge $R_Z$};
		\node [style=none] (83) at (29.25, -32.25) {};
		\node [style=none] (84) at (30.65, -31.8) {\huge $S_1 R_Z S_2$};
		\node [style=none] (85) at (29.25, -31) {};
	\end{pgfonlayer}
	\begin{pgfonlayer}{edgelayer}
		\draw [line width=1pt] (39.center) to (42.center);
		\draw [line width=1pt] (42.center) to (41.center);
		\draw [line width=1pt] (41.center) to (40.center);
		\draw [line width=1pt] (40.center) to (39.center);
		\draw [line width=1pt] (44.center) to (47.center);
		\draw [line width=1pt] (47.center) to (46.center);
		\draw [line width=1pt] (46.center) to (45.center);
		\draw [line width=1pt] (45.center) to (44.center);
		\draw [line width=1pt] (49.center) to (52.center);
		\draw [line width=1pt] (52.center) to (51.center);
		\draw [line width=1pt] (51.center) to (50.center);
		\draw [line width=1pt] (50.center) to (49.center);
		\draw [line width=1pt] (56.center) to (59.center);
		\draw [line width=1pt] (59.center) to (58.center);
		\draw [line width=1pt] (58.center) to (57.center);
		\draw [line width=1pt] (57.center) to (56.center);
		\draw [line width=1pt] (63.center) to (64.center);
		\draw [line width=1pt] (65.center) to (66.center);
		\draw [line width=1pt] (67.center) to (68.center);
		\draw [line width=1pt] (69.center) to (71.center);
		\draw [line width=1pt] (70.center) to (72.center);
		\draw [line width=1pt] (78.center) to (81.center);
		\draw [line width=1pt] (81.center) to (80.center);
		\draw [line width=1pt] (80.center) to (79.center);
		\draw [line width=1pt] (79.center) to (78.center);
		\draw [line width=1pt] (85.center) to (83.center);
	\end{pgfonlayer}
\end{tikzpicture}}
\endpgfgraphicnamed}} 
        
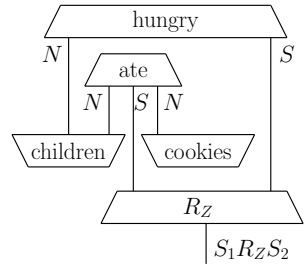
\captionof{figure}{IQP sansatz Combination}
        \label{fig:exampleWithBox}
    \end{minipage}
\end{center}

As before, we use Lambeq to produce the circuits corresponding to these diagrams  and these are shown in Figures \ref{fig:subjectCircuitSpider} and \ref{fig:subjectCircuitBox} respectively.

\begin{center}
    \centering
    \begin{minipage}{.5\textwidth}
        \centering
        \scalebox{0.3}{
 {%
\beginpgfgraphicnamed{subjectCircuitSpider}
\InputIfFileExists{subjectCircuitSpider.tikz}{}{\input{./tikz/subjectCircuitSpider.tikz}}
\endpgfgraphicnamed}} 
        \captionof{figure}{ Frobenius  Combination}
        \label{fig:subjectCircuitSpider}
    \end{minipage}%
    \begin{minipage}{0.5\textwidth}
        \centering
        \scalebox{0.3}{
        {%
\beginpgfgraphicnamed{subjectCircuitBox}
\InputIfFileExists{subjectCircuitBox.tikz}{}{\input{./tikz/subjectCircuitBox.tikz}}
\endpgfgraphicnamed}} 
        \captionof{figure}{IQP ansatz Combination}
        \label{fig:subjectCircuitBox}
    \end{minipage}
\end{center}

\subsection{Binary classification training}
\label{subsec:training}

 We represent the combined sentences  $S_1 \bigcirc S_2$ of each entry of the dataset by  a single parametrised quantum circuit, where the combination circuit is the combined version of the circuits of $S_1$ and $S_2$ according to the operations listed  above, in the previous subsection. 

We first randomly initialise a set of parameters $\Theta = \left ( \theta_1, \theta_2, ..., \theta_k \right )$, such that every parameter used by the IQP ansatz \cite{Shepherd_2009,Havl_ek_2019} of every word in the vocabulary is included. IQP stands for \emph{Instantaneous Quantum Polynomial}, it is a method of developing  circuits which interleaves layers of Hadamard quantum gates with diagonal unitaries.

Every combined circuit from the test dataset is then evaluated against $\Theta$. That means if the same word appears in two different circuits, it is represented as the same quantum state in both circuits.
We denote the output state from the circuit $ S_1 \bigcirc S_2$ as $\left | S_1 \bigcirc S_2 \left ( \Theta \right ) \right\rangle$.
The expected prediction of the binary class of each entry of the dataset is then given by Born rule, i.e. as follows 
\[
l^i_\Theta (S_1 \bigcirc S_2):= \left | \left \langle i\left | S_1 \bigcirc S_2 \left ( \Theta \right ) \right | \right \rangle \right |^2 + \epsilon 
\]
\noindent
In the above, $i \in \left \{ 0, 1 \right \}$, $\epsilon = 10^{-9}$ 
and $l_\Theta (S_1 \bigcirc S_2)$ is a probability distribution defined as follows
\[
l_\Theta (S_1 \bigcirc S_2):= (l^0_\Theta (S_1 \bigcirc S_2), l^1_\Theta (S_1 \bigcirc S_2))/{\sum_{i}^{}l^i_\Theta (S_1 \bigcirc S_2)}
\]
 \noindent
Note that the sum $\sum_i \left | \left \langle i\left | S_1 \bigcirc S_2 \left ( \Theta \right ) \right | \right \rangle \right |^2$ need not equal to $1$ as there can be post-selections in the circuits, rendering some circuit runs are discarded.
In some extreme cases, almost all circuit runs could be discarded. To avoid the possibility of dividing by zero, the small constant $\epsilon = 10^{-9}$ is added.
The predicted binary class from the model is then obtained by rounding to the nearest integer $\lfloor l_\Theta (S_1 \bigcirc S_2)\rceil$. 

The class labels are represented as one-hot encoding [0,1] and [1,0], corresponding to \textit{subject} anaphora and \textit{object} anaphora respectively. The model is trained by comparing the predicted label with the training label using a binary cross-entropy loss function and minimized using a non-gradient based optimisation algorithm known as SPSA (Simultaneous Perturbation Stochastic Approximation) \cite{705889}. As a result, the system learns to classify the sentences by adjusting the parameters.

\subsection{Models}
First and foremost, we implement the $\sllm$ model. This model encodes both the grammatical and discourse structures of pairs of sentences of our dataset. In other words, in the $\sllm$ model, each sentence of the dataset has a circuit built according to the string diagram of its syntactic structure. Further, the two sentences of the dataset are connected to each other according to the discourse structure between them. For comparison reasons, we implement and experiment with three other models: (1) a model where there is notion of grammar and no notion of discourse, (2) a model where there is no notion of grammar, but anaphora is resolved, and (3) a model where we have a notion of grammar, but the anaphora relation between the sentence are not resolved. 

We train a binary classification task for each of these four models (the $\sllm$ model is our model no. 4), In order to do so, we have to combine the sentences of each entry of the dataset to obtain one single output state. We experiment with both the Frobenius and the Z Rotation operations, described in the previous susbection. As a result, in total we train eight binary classification models.

We elaborate on each of these models below and draw the string diagrams corresponding to the subject and object anaphora relations.. For reasons of space, we only provide the Frobenius combinations of sentences. The Z rotation combinations are obtained in a similar way, by only replacing the final (lower most) Frobenius with a Z rotation box.

\begin{itemize}
\item Model 1: this model is a bag of words model where the words are connected to each other using the Frobenius multiplication operation. As this operation is commutative, we have no notion of grammar in the model. Further, we do not resolve the anaphora, i.e. we do not connect any of the noun phrases of the first sentence to the pronoun of the second sentence. We refer to this model as the \emph{no-grammar, no-discourse} model. This model is depicted in Figure \ref{fig:model1}. 

\begin{center}
\scalebox{0.36}{
 {%
\beginpgfgraphicnamed{model1_new}
\begin{tikzpicture}
	\begin{pgfonlayer}{nodelayer}
		\node [style=none] (112) at (21.75, -29.5) {};
		\node [style=none] (113) at (23.25, -29.5) {};
		\node [style=none] (114) at (24, -30.5) {};
		\node [style=none] (115) at (21, -30.5) {};
		\node [style=none] (116) at (22.5, -30) {\huge children};
		\node [style=none] (117) at (25.25, -29.5) {};
		\node [style=none] (118) at (26.75, -29.5) {};
		\node [style=none] (119) at (27.5, -30.5) {};
		\node [style=none] (120) at (24.5, -30.5) {};
		\node [style=none] (121) at (26, -30) {\huge ate};
		\node [style=none] (122) at (28.75, -29.5) {};
		\node [style=none] (123) at (30.25, -29.5) {};
		\node [style=none] (124) at (31, -30.5) {};
		\node [style=none] (125) at (28, -30.5) {};
		\node [style=none] (126) at (29.5, -30) {\huge cookies};
		\node [style=none] (127) at (32.75, -29.5) {};
		\node [style=none] (128) at (34.25, -29.5) {};
		\node [style=none] (129) at (35, -30.5) {};
		\node [style=none] (130) at (32, -30.5) {};
		\node [style=none] (131) at (33.5, -30) {\huge they};
		\node [style=none] (132) at (36.25, -29.5) {};
		\node [style=none] (133) at (37.75, -29.5) {};
		\node [style=none] (134) at (38.5, -30.5) {};
		\node [style=none] (135) at (35.5, -30.5) {};
		\node [style=none] (136) at (37, -30) {\huge were};
		\node [style=none] (137) at (39.75, -29.5) {};
		\node [style=none] (138) at (41.25, -29.5) {};
		\node [style=none] (139) at (42, -30.5) {};
		\node [style=none] (140) at (39, -30.5) {};
		\node [style=none] (141) at (40.5, -30) {\huge hungry};
		\node [style=none] (142) at (22.5, -30.5) {};
		\node [style=none] (143) at (26, -30.5) {};
		\node [style=none] (144) at (29.5, -30.5) {};
		\node [style=none] (145) at (33.5, -30.5) {};
		\node [style=none] (146) at (37, -30.5) {};
		\node [style=none] (147) at (40.5, -30.5) {};
		\node [style={fill=black, draw=black, shape=circle}] (148) at (26, -32.5) {};
		\node [style={fill=black, draw=black, shape=circle}] (149) at (37, -32.5) {};
		\node [style={fill=black, draw=black, shape=circle}] (150) at (31.5, -34.5) {};
		\node [style=none] (151) at (31.5, -36) {};
		\node [style=none] (152) at (23.5, -31) {\huge $S$};
		\node [style=none] (153) at (26.75, -31) {\huge $S$};
		\node [style=none] (154) at (30, -31) {\huge $S$};
		\node [style=none] (155) at (34.5, -31) {\huge $S$};
		\node [style=none] (156) at (37.75, -31) {\huge $S$};
		\node [style=none] (157) at (41, -31) {\huge $S$};
		\node [style=none] (158) at (37, -33.25) {\huge $S$};
		\node [style=none] (159) at (25.75, -33.25) {\huge $S$};
		\node [style=none] (160) at (32.25, -35) {\huge $S$};
		\node [style=none] (161) at (21.75, -37) {};
		\node [style=none] (162) at (23.25, -37) {};
		\node [style=none] (163) at (24, -38) {};
		\node [style=none] (164) at (21, -38) {};
		\node [style=none] (165) at (22.5, -37.5) {\huge children};
		\node [style=none] (166) at (25.25, -37) {};
		\node [style=none] (167) at (26.75, -37) {};
		\node [style=none] (168) at (27.5, -38) {};
		\node [style=none] (169) at (24.5, -38) {};
		\node [style=none] (170) at (26, -37.5) {\huge ate};
		\node [style=none] (171) at (28.75, -37) {};
		\node [style=none] (172) at (30.25, -37) {};
		\node [style=none] (173) at (31, -38) {};
		\node [style=none] (174) at (28, -38) {};
		\node [style=none] (175) at (29.5, -37.5) {\huge cookies};
		\node [style=none] (176) at (32.75, -37) {};
		\node [style=none] (177) at (34.25, -37) {};
		\node [style=none] (178) at (35, -38) {};
		\node [style=none] (179) at (32, -38) {};
		\node [style=none] (180) at (33.5, -37.5) {\huge they};
		\node [style=none] (181) at (36.25, -37) {};
		\node [style=none] (182) at (37.75, -37) {};
		\node [style=none] (183) at (38.5, -38) {};
		\node [style=none] (184) at (35.5, -38) {};
		\node [style=none] (185) at (37, -37.5) {\huge were};
		\node [style=none] (186) at (39.75, -37) {};
		\node [style=none] (187) at (41.25, -37) {};
		\node [style=none] (188) at (42, -38) {};
		\node [style=none] (189) at (39, -38) {};
		\node [style=none] (190) at (40.5, -37.5) {\huge tasty};
		\node [style=none] (191) at (22.5, -38) {};
		\node [style=none] (192) at (26, -38) {};
		\node [style=none] (193) at (29.5, -38) {};
		\node [style=none] (194) at (33.5, -38) {};
		\node [style=none] (195) at (37, -38) {};
		\node [style=none] (196) at (40.5, -38) {};
		\node [style={fill=black, draw=black, shape=circle}] (197) at (26, -40) {};
		\node [style={fill=black, draw=black, shape=circle}] (198) at (37, -40) {};
		\node [style={fill=black, draw=black, shape=circle}] (199) at (31.5, -42) {};
		\node [style=none] (200) at (31.5, -43.5) {};
		\node [style=none] (201) at (23.5, -38.5) {\huge $S$};
		\node [style=none] (202) at (26.75, -38.5) {\huge $S$};
		\node [style=none] (203) at (30, -38.5) {\huge $S$};
		\node [style=none] (204) at (34.5, -38.5) {\huge $S$};
		\node [style=none] (205) at (37.75, -38.5) {\huge $S$};
		\node [style=none] (206) at (41, -38.5) {\huge $S$};
		\node [style=none] (207) at (37, -40.75) {\huge $S$};
		\node [style=none] (208) at (25.75, -40.75) {\huge $S$};
		\node [style=none] (209) at (32.25, -42.5) {\huge $S$};
	\end{pgfonlayer}
	\begin{pgfonlayer}{edgelayer}
		\draw [line width=1pt] (112.center) to (115.center);
		\draw [line width=1pt] (115.center) to (114.center);
		\draw [line width=1pt] (114.center) to (113.center);
		\draw [line width=1pt] (113.center) to (112.center);
		\draw [line width=1pt] (117.center) to (120.center);
		\draw [line width=1pt] (120.center) to (119.center);
		\draw [line width=1pt] (119.center) to (118.center);
		\draw [line width=1pt] (118.center) to (117.center);
		\draw [line width=1pt] (122.center) to (125.center);
		\draw [line width=1pt] (125.center) to (124.center);
		\draw [line width=1pt] (124.center) to (123.center);
		\draw [line width=1pt] (123.center) to (122.center);
		\draw [line width=1pt] (127.center) to (130.center);
		\draw [line width=1pt] (130.center) to (129.center);
		\draw [line width=1pt] (129.center) to (128.center);
		\draw [line width=1pt] (128.center) to (127.center);
		\draw [line width=1pt] (132.center) to (135.center);
		\draw [line width=1pt] (135.center) to (134.center);
		\draw [line width=1pt] (134.center) to (133.center);
		\draw [line width=1pt] (133.center) to (132.center);
		\draw [line width=1pt] (137.center) to (140.center);
		\draw [line width=1pt] (140.center) to (139.center);
		\draw [line width=1pt] (139.center) to (138.center);
		\draw [line width=1pt] (138.center) to (137.center);
		\draw [line width=1pt] (143.center) to (148);
		\draw [line width=1pt] (146.center) to (149);
		\draw [line width=1pt] (150) to (151.center);
		\draw [line width=1pt] [bend right] (142.center) to (148);
		\draw [line width=1pt] [bend left] (144.center) to (148);
		\draw [line width=1pt] [bend right] (145.center) to (149);
		\draw [line width=1pt] [bend right] (149) to (147.center);
		\draw [line width=1pt] [bend left=15] (149) to (150);
		\draw [line width=1pt] [bend left=15] (150) to (148);
		\draw [line width=1pt] (161.center) to (164.center);
		\draw [line width=1pt] (164.center) to (163.center);
		\draw [line width=1pt] (163.center) to (162.center);
		\draw [line width=1pt] (162.center) to (161.center);
		\draw [line width=1pt] (166.center) to (169.center);
		\draw [line width=1pt] (169.center) to (168.center);
		\draw [line width=1pt] (168.center) to (167.center);
		\draw [line width=1pt] (167.center) to (166.center);
		\draw [line width=1pt] (171.center) to (174.center);
		\draw [line width=1pt] (174.center) to (173.center);
		\draw [line width=1pt] (173.center) to (172.center);
		\draw [line width=1pt] (172.center) to (171.center);
		\draw [line width=1pt] (176.center) to (179.center);
		\draw [line width=1pt] (179.center) to (178.center);
		\draw [line width=1pt] (178.center) to (177.center);
		\draw [line width=1pt] (177.center) to (176.center);
		\draw [line width=1pt] (181.center) to (184.center);
		\draw [line width=1pt] (184.center) to (183.center);
		\draw [line width=1pt] (183.center) to (182.center);
		\draw [line width=1pt] (182.center) to (181.center);
		\draw [line width=1pt] (186.center) to (189.center);
		\draw [line width=1pt] (189.center) to (188.center);
		\draw [line width=1pt] (188.center) to (187.center);
		\draw [line width=1pt] (187.center) to (186.center);
		\draw [line width=1pt] (192.center) to (197);
		\draw [line width=1pt] (195.center) to (198);
		\draw [line width=1pt] (199) to (200.center);
		\draw [line width=1pt] [bend right] (191.center) to (197);
		\draw [line width=1pt] [bend left] (193.center) to (197);
		\draw [line width=1pt] [bend right] (194.center) to (198);
		\draw [line width=1pt] [bend right] (198) to (196.center);
		\draw [line width=1pt] [bend left=15] (198) to (199);
		\draw [line width=1pt] [bend left=15] (199) to (197);
	\end{pgfonlayer}
\end{tikzpicture}}
\endpgfgraphicnamed}} 
  
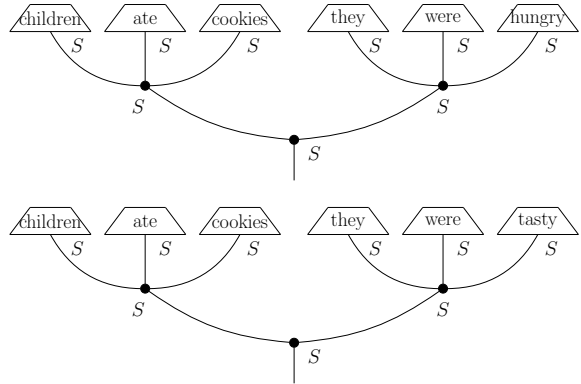
\captionof{figure}{Bag-of-words Model (no-grammar, no-discourse)}
    \label{fig:model1}
\end{center}

\item Model 2: this model is similar to the first model in that, there is still no notion of grammar: the words are connected to each other using the Frobenius multiplication. It differs from it in that either the subject or object of the first sentence is connected to the pronoun of the second sentence. So we resolve the anaphora. We refer to this model as \emph{no grammar, discourse} model. This model is depicted in Figure \ref{fig:model2}.

\begin{center}
\scalebox{0.4}{
 {%
\beginpgfgraphicnamed{model2_final}
\begin{tikzpicture}
	\begin{pgfonlayer}{nodelayer}
		\node [style=none] (0) at (1.25, -28.25) {};
		\node [style=none] (1) at (10.5, -28.25) {};
		\node [style=none] (2) at (11, -29.25) {};
		\node [style=none] (3) at (0.5, -29.25) {};
		\node [style=none] (4) at (6, -28.75) {\huge children};
		\node [style=none] (5) at (4.5, -30.25) {};
		\node [style=none] (6) at (6, -30.25) {};
		\node [style=none] (7) at (6.75, -31.25) {};
		\node [style=none] (8) at (3.75, -31.25) {};
		\node [style=none] (9) at (5.25, -30.75) {\huge ate};
		\node [style=none] (10) at (7.75, -30.25) {};
		\node [style=none] (11) at (9.25, -30.25) {};
		\node [style=none] (12) at (10, -31.25) {};
		\node [style=none] (13) at (7, -31.25) {};
		\node [style=none] (14) at (8.5, -30.75) {\huge cookies};
		\node [style=none] (15) at (1.5, -29.25) {};
		\node [style=none] (16) at (5.25, -31.25) {};
		\node [style=none] (17) at (8.5, -31.25) {};
		\node [style={fill=black, draw=black, shape=circle}] (18) at (5.25, -32.75) {};
		\node [style=none] (19) at (2.25, -30) {\huge $S$};
		\node [style=none] (20) at (9.25, -31.75) {\huge $S$};
		\node [style=none] (21) at (13.25, -30.25) {};
		\node [style=none] (22) at (14.75, -30.25) {};
		\node [style=none] (23) at (15.5, -31.25) {};
		\node [style=none] (24) at (12.5, -31.25) {};
		\node [style=none] (25) at (14, -30.75) {\huge were};
		\node [style=none] (26) at (16.5, -30.25) {};
		\node [style=none] (27) at (18, -30.25) {};
		\node [style=none] (28) at (18.75, -31.25) {};
		\node [style=none] (29) at (15.75, -31.25) {};
		\node [style=none] (30) at (17.25, -30.75) {\huge hungry};
		\node [style=none] (31) at (10.25, -29.25) {};
		\node [style=none] (32) at (14, -31.25) {};
		\node [style=none] (33) at (17.25, -31.25) {};
		\node [style={fill=black, draw=black, shape=circle}] (34) at (14, -32.75) {};
		\node [style=none] (35) at (14.75, -31.75) {\huge $S$};
		\node [style=none] (36) at (18, -31.75) {\huge $S$};
		\node [style={fill=black, draw=black, shape=circle}] (37) at (9.75, -35) {};
		\node [style=none] (38) at (9.75, -36.25) {};
		\node [style=none] (39) at (6, -31.75) {\huge $S$};
		\node [style=none] (40) at (11, -30) {\huge $S$};
		\node [style=none] (41) at (14.75, -33.25) {\huge $S$};
		\node [style=none] (42) at (6.5, -33) {\huge $S$};
		\node [style=none] (43) at (10.5, -35.5) {\huge $S$};
		\node [style=none] (88) at (5, -38) {};
		\node [style=none] (89) at (6.5, -38) {};
		\node [style=none] (90) at (7.25, -39) {};
		\node [style=none] (91) at (4.25, -39) {};
		\node [style=none] (92) at (5.75, -38.5) {\huge ate};
		\node [style=none] (93) at (8.5, -38) {};
		\node [style=none] (94) at (10, -38) {};
		\node [style=none] (95) at (10.75, -39) {};
		\node [style=none] (96) at (7.75, -39) {};
		\node [style=none] (97) at (9.25, -38.5) {\huge cookies};
		\node [style=none] (98) at (2.25, -39) {};
		\node [style=none] (99) at (5.75, -39) {};
		\node [style=none] (100) at (9, -39) {};
		\node [style={fill=black, draw=black, shape=circle}] (101) at (5.75, -41) {};
		\node [style=none] (102) at (3, -39.5) {\huge $S$};
		\node [style=none] (103) at (8, -39.5) {\huge $S$};
		\node [style=none] (104) at (12.5, -38.75) {};
		\node [style=none] (105) at (14, -38.75) {};
		\node [style=none] (106) at (14.75, -39.75) {};
		\node [style=none] (107) at (11.75, -39.75) {};
		\node [style=none] (108) at (13.25, -39.25) {\huge were};
		\node [style=none] (109) at (15.75, -38.75) {};
		\node [style=none] (110) at (17.25, -38.75) {};
		\node [style=none] (111) at (18, -39.75) {};
		\node [style=none] (112) at (15, -39.75) {};
		\node [style=none] (113) at (16.5, -39.25) {\huge tasty};
		\node [style=none] (114) at (13.25, -39.75) {};
		\node [style=none] (115) at (16.5, -39.75) {};
		\node [style={fill=black, draw=black, shape=circle}] (116) at (13.25, -41) {};
		\node [style=none] (117) at (14, -40.25) {\huge $S$};
		\node [style=none] (118) at (17.25, -40.25) {\huge $S$};
		\node [style={fill=black, draw=black, shape=circle}] (119) at (9.75, -43) {};
		\node [style=none] (120) at (9.75, -43.75) {};
		\node [style=none] (121) at (6.5, -39.5) {\huge $S$};
		\node [style=none] (122) at (10.75, -39.5) {\huge $S$};
		\node [style=none] (123) at (14, -41.5) {\huge $S$};
		\node [style=none] (124) at (7, -41.5) {\huge $S$};
		\node [style=none] (125) at (10.75, -43.5) {\huge $S$};
		\node [style=none] (126) at (1.5, -38) {};
		\node [style=none] (127) at (3, -38) {};
		\node [style=none] (128) at (3.75, -39) {};
		\node [style=none] (129) at (0.75, -39) {};
		\node [style=none] (130) at (2.25, -38.5) {\huge children};
		\node [style=none] (131) at (9.75, -39) {};
	\end{pgfonlayer}
	\begin{pgfonlayer}{edgelayer}
		\draw [line width=0.5pt] (0.center) to (3.center);
		\draw [line width=0.5pt] (3.center) to (2.center);
		\draw [line width=0.5pt] (2.center) to (1.center);
		\draw [line width=0.5pt] (1.center) to (0.center);
		\draw [line width=0.5pt] (5.center) to (8.center);
		\draw [line width=0.5pt] (8.center) to (7.center);
		\draw [line width=0.5pt] (7.center) to (6.center);
		\draw [line width=0.5pt] (6.center) to (5.center);
		\draw [line width=0.5pt] (10.center) to (13.center);
		\draw [line width=0.5pt] (13.center) to (12.center);
		\draw [line width=0.5pt] (12.center) to (11.center);
		\draw [line width=0.5pt] (11.center) to (10.center);
		\draw [line width=1pt] (16.center) to (18);
		\draw [line width=1pt] [bend right=45] (15.center) to (18);
		\draw [line width=1pt] [bend left] (17.center) to (18);
		\draw [line width=0.5pt] (21.center) to (24.center);
		\draw [line width=0.5pt] (24.center) to (23.center);
		\draw [line width=0.5pt] (23.center) to (22.center);
		\draw [line width=0.5pt] (22.center) to (21.center);
		\draw [line width=0.5pt] (26.center) to (29.center);
		\draw [line width=0.5pt] (29.center) to (28.center);
		\draw [line width=0.5pt] (28.center) to (27.center);
		\draw [line width=0.5pt] (27.center) to (26.center);
		\draw [line width=1pt] (32.center) to (34);
		\draw [line width=1pt] [bend right=45] (31.center) to (34);
		\draw [line width=1pt] [bend left] (33.center) to (34);
		\draw [line width=1pt] [bend right, looseness=0.75] (18) to (37);
		\draw [line width=1pt] [bend right, looseness=0.75] (37) to (34);
		\draw [line width=1pt] (37) to (38.center);
		\draw [line width=0.5pt] (88.center) to (91.center);
		\draw [line width=0.5pt] (91.center) to (90.center);
		\draw [line width=0.5pt] (90.center) to (89.center);
		\draw [line width=0.5pt] (89.center) to (88.center);
		\draw [line width=0.5pt] (93.center) to (96.center);
		\draw [line width=0.5pt] (96.center) to (95.center);
		\draw [line width=0.5pt] (95.center) to (94.center);
		\draw [line width=0.5pt] (94.center) to (93.center);
		\draw [line width=1pt] (99.center) to (101);
		\draw [line width=1pt] [bend right] (98.center) to (101);
		\draw [line width=1pt] [bend left] (100.center) to (101);
		\draw [line width=0.5pt] (104.center) to (107.center);
		\draw [line width=0.5pt] (107.center) to (106.center);
		\draw [line width=0.5pt] (106.center) to (105.center);
		\draw [line width=0.5pt] (105.center) to (104.center);
		\draw [line width=0.5pt] (109.center) to (112.center);
		\draw [line width=0.5pt] (112.center) to (111.center);
		\draw [line width=0.5pt] (111.center) to (110.center);
		\draw [line width=0.5pt] (110.center) to (109.center);
		\draw [line width=1pt] (114.center) to (116);
		\draw [line width=1pt] [bend left] (115.center) to (116);
		\draw [line width=1pt] [bend right, looseness=0.75] (101) to (119);
		\draw [line width=1pt] [bend right, looseness=0.75] (119) to (116);
		\draw [line width=1pt] (119) to (120.center);
		\draw [line width=0.5pt] (126.center) to (129.center);
		\draw [line width=0.5pt] (129.center) to (128.center);
		\draw [line width=0.5pt] (128.center) to (127.center);
		\draw [line width=0.5pt] (127.center) to (126.center);
		\draw [line width=1pt] [bend right] (131.center) to (116);
	\end{pgfonlayer}
\end{tikzpicture}}
\endpgfgraphicnamed}} 
  
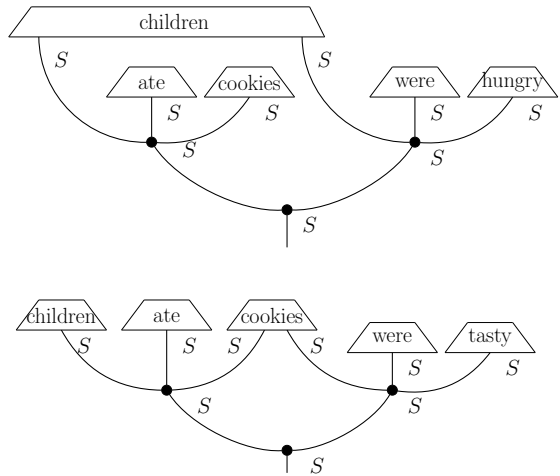
\captionof{figure}{Bag-of-words Model with discourse (no-grammar, discourse)}
    \label{fig:model2}
\end{center}

\item Model 3: Here we have a notion of grammar; each sentence has a circuit obtained from the string diagram of its syntactic structure. There is, however, no notion of discourse. The anaphora is not resolved and neither the subject or object of the first sentence is connected to the pronoun of the second sentence. We refer to this model as the \emph{grammar, no-discourse} model. This model is depicted in Figure \ref{fig:model3}.

\begin{center}
\scalebox{0.4}{
 {%
\beginpgfgraphicnamed{model4}
\begin{tikzpicture}
	\begin{pgfonlayer}{nodelayer}
		\node [style=none] (0) at (16.5, -27.25) {};
		\node [style=none] (1) at (19, -27.25) {};
		\node [style=none] (2) at (18.5, -28.25) {};
		\node [style=none] (3) at (17, -28.25) {};
		\node [style=none] (4) at (17.75, -27.75) {\huge they};
		\node [style=none] (5) at (11, -24.25) {};
		\node [style=none] (6) at (20.75, -24.25) {};
		\node [style=none] (7) at (21.25, -25.25) {};
		\node [style=none] (8) at (10.5, -25.25) {};
		\node [style=none] (9) at (14.25, -24.75) {\huge ate};
		\node [style=none] (10) at (15, -29) {};
		\node [style=none] (11) at (20.25, -29) {};
		\node [style=none] (12) at (19.75, -30) {};
		\node [style=none] (13) at (15.5, -30) {};
		\node [style=none] (14) at (17.5, -29.5) {\huge hungry};
		\node [style=none] (17) at (9.5, -26.25) {};
		\node [style=none] (18) at (13, -26.25) {};
		\node [style=none] (19) at (12.5, -27.25) {};
		\node [style=none] (20) at (10, -27.25) {};
		\node [style=none] (21) at (11.25, -26.75) {\huge children};
		\node [style=none] (22) at (11.5, -27.25) {};
		\node [style=none] (23) at (10.75, -27.25) {};
		\node [style=none] (24) at (17.75, -28.25) {};
		\node [style=none] (26) at (18.5, -28.25) {};
		\node [style=none] (27) at (15.5, -29) {};
		\node [style=none] (28) at (17, -28.25) {};
		\node [style=none] (29) at (12.5, -26.25) {};
		\node [style=none] (30) at (11.25, -25.25) {};
		\node [style=none] (31) at (20.75, -25.25) {};
		\node [style=none] (32) at (11.25, -26.25) {};
		\node [style=none] (33) at (20.75, -30) {};
		\node [style=none] (73) at (10.75, -25.75) {\huge $N$};
		\node [style={fill=black, draw=black, shape=circle}] (78) at (14, -32.75) {};
		\node [style=none] (79) at (14, -33.5) {};
		\node [style=none] (84) at (14.5, -33) {\huge $S$};
		\node [style=none] (85) at (13.5, -25.25) {};
		\node [style=none] (86) at (13.5, -32.25) {};
		\node [style=none] (87) at (19.75, -29) {};
		\node [style=none] (88) at (18, -27.25) {};
		\node [style=none] (89) at (18, -26.75) {};
		\node [style=none] (90) at (19.75, -26.75) {};
		\node [style=none] (91) at (14.5, -27.75) {};
		\node [style=none] (92) at (15.5, -27.75) {};
		\node [style=none] (93) at (14.5, -32.25) {};
		\node [style=none] (94) at (19, -31.25) {};
		\node [style=none] (95) at (22.5, -31.25) {};
		\node [style=none] (96) at (22, -32.25) {};
		\node [style=none] (97) at (19.5, -32.25) {};
		\node [style=none] (98) at (20.75, -31.75) {\huge cookies};
		\node [style=none] (99) at (21, -32.25) {};
		\node [style=none] (100) at (20.25, -32.25) {};
		\node [style=none] (101) at (22, -31.25) {};
		\node [style=none] (102) at (20.75, -31.25) {};
		\node [style=none] (103) at (21.5, -25.75) {\huge $N$};
		\node [style=none] (104) at (18.5, -26.75) {\huge $N$};
		\node [style=none] (105) at (15, -28) {\huge $S$};
		\node [style=none] (106) at (14, -25.75) {\huge $S$};
		\node [style=none] (107) at (31.5, -27.25) {};
		\node [style=none] (108) at (34.25, -27.25) {};
		\node [style=none] (109) at (33.75, -28.25) {};
		\node [style=none] (110) at (32, -28.25) {};
		\node [style=none] (111) at (32.75, -27.75) {\huge they};
		\node [style=none] (112) at (25, -24.25) {};
		\node [style=none] (113) at (35.75, -24.25) {};
		\node [style=none] (114) at (36.25, -25.25) {};
		\node [style=none] (115) at (24.5, -25.25) {};
		\node [style=none] (116) at (28.25, -24.75) {\huge ate};
		\node [style=none] (117) at (30, -29.5) {};
		\node [style=none] (118) at (35.25, -29.5) {};
		\node [style=none] (119) at (34.75, -30.5) {};
		\node [style=none] (120) at (30.5, -30.5) {};
		\node [style=none] (121) at (32.5, -30) {\huge tasty};
		\node [style=none] (122) at (23.5, -26.25) {};
		\node [style=none] (123) at (27, -26.25) {};
		\node [style=none] (124) at (26.5, -27.25) {};
		\node [style=none] (125) at (24, -27.25) {};
		\node [style=none] (126) at (25.25, -26.75) {\huge children};
		\node [style=none] (127) at (25.5, -27.25) {};
		\node [style=none] (128) at (24.75, -27.25) {};
		\node [style=none] (131) at (30.5, -29.5) {};
		\node [style=none] (133) at (26.5, -26.25) {};
		\node [style=none] (134) at (25.25, -25.25) {};
		\node [style=none] (135) at (35.75, -25.25) {};
		\node [style=none] (136) at (25.25, -26.25) {};
		\node [style=none] (137) at (35.75, -30) {};
		\node [style=none] (138) at (24.75, -25.75) {\huge $N$};
		\node [style={fill=black, draw=black, shape=circle}] (139) at (28, -32.75) {};
		\node [style=none] (140) at (28, -33.5) {};
		\node [style=none] (141) at (28.5, -33) {\huge $S$};
		\node [style=none] (142) at (27.5, -25.25) {};
		\node [style=none] (143) at (27.5, -32.25) {};
		\node [style=none] (144) at (34.75, -29.5) {};
		\node [style=none] (145) at (33, -27.25) {};
		\node [style=none] (146) at (33, -26.75) {};
		\node [style=none] (147) at (34.75, -26.75) {};
		\node [style=none] (150) at (28.5, -32.25) {};
		\node [style=none] (151) at (34, -31.25) {};
		\node [style=none] (152) at (37.5, -31.25) {};
		\node [style=none] (153) at (37, -32.25) {};
		\node [style=none] (154) at (34.5, -32.25) {};
		\node [style=none] (155) at (35.75, -31.75) {\huge cookies};
		\node [style=none] (156) at (36, -32.25) {};
		\node [style=none] (157) at (35.25, -32.25) {};
		\node [style=none] (158) at (37, -31.25) {};
		\node [style=none] (159) at (35.75, -31.25) {};
		\node [style=none] (160) at (36.5, -25.75) {\huge $N$};
		\node [style=none] (161) at (33.5, -26.75) {\huge $N$};
		\node [style=none] (163) at (28, -25.75) {\huge $S$};
		\node [style=none] (164) at (28.5, -27.25) {};
		\node [style=none] (165) at (30.5, -27.25) {};
		\node [style=none] (166) at (31, -28.25) {};
		\node [style=none] (167) at (28, -28.25) {};
		\node [style=none] (168) at (29.5, -27.75) {\huge looked};
		\node [style=none] (170) at (30.5, -28.25) {};
		\node [style=none] (171) at (28.5, -28.25) {};
		\node [style=none] (172) at (29, -28.75) {\huge $S$};
		\node [style=none] (173) at (31, -28.75) {\huge $S$};
	\end{pgfonlayer}
	\begin{pgfonlayer}{edgelayer}
		\draw (0.center) to (3.center);
		\draw (3.center) to (2.center);
		\draw (2.center) to (1.center);
		\draw (1.center) to (0.center);
		\draw (5.center) to (8.center);
		\draw (8.center) to (7.center);
		\draw (7.center) to (6.center);
		\draw (6.center) to (5.center);
		\draw (10.center) to (13.center);
		\draw (13.center) to (12.center);
		\draw (12.center) to (11.center);
		\draw (11.center) to (10.center);
		\draw (17.center) to (20.center);
		\draw (20.center) to (19.center);
		\draw (19.center) to (18.center);
		\draw (18.center) to (17.center);
		\draw (30.center) to (32.center);
		\draw (31.center) to (33.center);
		\draw (78) to (79.center);
		\draw (85.center) to (86.center);
		\draw (87.center) to (90.center);
		\draw [bend left=270] (90.center) to (89.center);
		\draw (89.center) to (88.center);
		\draw [bend left=270] (92.center) to (91.center);
		\draw (92.center) to (27.center);
		\draw (91.center) to (93.center);
		\draw [bend right=90, looseness=2.00] (86.center) to (93.center);
		\draw (94.center) to (97.center);
		\draw (97.center) to (96.center);
		\draw (96.center) to (95.center);
		\draw (95.center) to (94.center);
		\draw (33.center) to (102.center);
		\draw (107.center) to (110.center);
		\draw (110.center) to (109.center);
		\draw (109.center) to (108.center);
		\draw (108.center) to (107.center);
		\draw (112.center) to (115.center);
		\draw (115.center) to (114.center);
		\draw (114.center) to (113.center);
		\draw (113.center) to (112.center);
		\draw (117.center) to (120.center);
		\draw (120.center) to (119.center);
		\draw (119.center) to (118.center);
		\draw (118.center) to (117.center);
		\draw (122.center) to (125.center);
		\draw (125.center) to (124.center);
		\draw (124.center) to (123.center);
		\draw (123.center) to (122.center);
		\draw (134.center) to (136.center);
		\draw (135.center) to (137.center);
		\draw (139) to (140.center);
		\draw (142.center) to (143.center);
		\draw (144.center) to (147.center);
		\draw [bend left=270] (147.center) to (146.center);
		\draw (146.center) to (145.center);
		\draw [bend right=90, looseness=2.00] (143.center) to (150.center);
		\draw (151.center) to (154.center);
		\draw (154.center) to (153.center);
		\draw (153.center) to (152.center);
		\draw (152.center) to (151.center);
		\draw (137.center) to (159.center);
		\draw (164.center) to (167.center);
		\draw (167.center) to (166.center);
		\draw (166.center) to (165.center);
		\draw (165.center) to (164.center);
		\draw (171.center) to (150.center);
		\draw (170.center) to (131.center);
	\end{pgfonlayer}
\end{tikzpicture}}
\endpgfgraphicnamed}} 
 
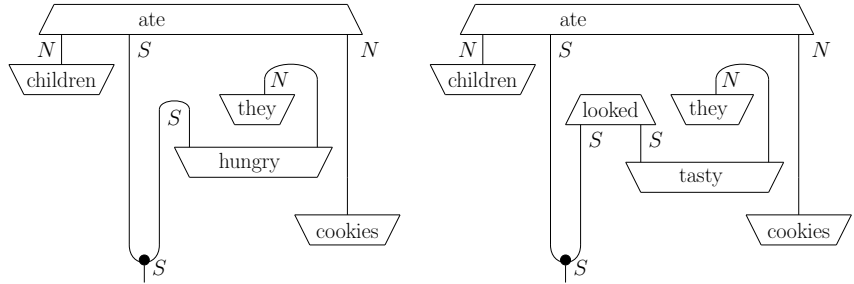
\captionof{figure}{Grammar with no discourse Model }
    \label{fig:model3}
\end{center}

\item Model 4: Finally, in this model we have the full $\sllm$ syntax turned into string diagrams and then into circuits. Here, we both encode the grammatical structure of each sentence, as well as resolve the anaphora. We refer to this model as \emph{grammar, discourse} model and depicted in Figure \ref{fig:model4}.

\begin{center}
\scalebox{0.45}{
 {%
\beginpgfgraphicnamed{model3}
\begin{tikzpicture}
	\begin{pgfonlayer}{nodelayer}
		\node [style=none] (0) at (12.25, -25.75) {};
		\node [style=none] (1) at (14.25, -25.75) {};
		\node [style=none] (2) at (14.75, -26.75) {};
		\node [style=none] (3) at (11.75, -26.75) {};
		\node [style=none] (4) at (13.25, -26.25) {\huge ate};
		\node [style=none] (5) at (11, -24.25) {};
		\node [style=none] (6) at (17.5, -24.25) {};
		\node [style=none] (7) at (18, -25.25) {};
		\node [style=none] (8) at (10.5, -25.25) {};
		\node [style=none] (9) at (14.25, -24.75) {\huge hungry};
		\node [style=none] (10) at (13.5, -28.25) {};
		\node [style=none] (11) at (17, -28.25) {};
		\node [style=none] (12) at (16.5, -29.25) {};
		\node [style=none] (13) at (14, -29.25) {};
		\node [style=none] (14) at (15.25, -28.75) {\huge cookies};
		\node [style=none] (15) at (15.5, -29.25) {};
		\node [style=none] (16) at (14.75, -29.25) {};
		\node [style=none] (17) at (9.5, -28.25) {};
		\node [style=none] (18) at (13, -28.25) {};
		\node [style=none] (19) at (12.5, -29.25) {};
		\node [style=none] (20) at (10, -29.25) {};
		\node [style=none] (21) at (11.25, -28.75) {\huge children};
		\node [style=none] (22) at (11.5, -29.25) {};
		\node [style=none] (23) at (10.75, -29.25) {};
		\node [style=none] (24) at (13.25, -26.75) {};
		\node [style=none] (25) at (13.25, -30) {};
		\node [style=none] (26) at (14, -26.75) {};
		\node [style=none] (27) at (14, -28.25) {};
		\node [style=none] (28) at (12.5, -26.75) {};
		\node [style=none] (29) at (12.5, -28.25) {};
		\node [style=none] (30) at (11.25, -25.25) {};
		\node [style=none] (31) at (17.5, -25.25) {};
		\node [style=none] (32) at (11.25, -28.25) {};
		\node [style=none] (33) at (17.5, -30) {};
		\node [style=none] (73) at (10.75, -25.75) {\huge $N$};
		\node [style=none] (74) at (18, -25.75) {\huge $S$};
		\node [style=none] (75) at (12, -27.25) {\huge $N$};
		\node [style=none] (76) at (13.5, -27.25) {\huge $S$};
		\node [style=none] (77) at (14.5, -27.25) {\huge $N$};
		\node [style={fill=black, draw=black, shape=circle}] (78) at (15.5, -30.75) {};
		\node [style=none] (79) at (15.5, -31.75) {};
		\node [style=none] (84) at (16, -31) {\huge $S$};
		\node [style=none] (120) at (22.25, -24.25) {};
		\node [style=none] (121) at (23.75, -24.25) {};
		\node [style=none] (122) at (24.25, -25.25) {};
		\node [style=none] (123) at (21.75, -25.25) {};
		\node [style=none] (124) at (23, -24.75) {\huge ate};
		\node [style=none] (125) at (19.25, -26.75) {};
		\node [style=none] (126) at (22.75, -26.75) {};
		\node [style=none] (127) at (22.25, -27.75) {};
		\node [style=none] (128) at (19.75, -27.75) {};
		\node [style=none] (129) at (21, -27.25) {\huge children};
		\node [style=none] (130) at (23, -25.25) {};
		\node [style=none] (131) at (23, -30) {};
		\node [style=none] (132) at (23.75, -25.25) {};
		\node [style=none] (133) at (22.25, -25.25) {};
		\node [style=none] (134) at (22.25, -26.75) {};
		\node [style=none] (135) at (24.75, -26.5) {};
		\node [style=none] (136) at (26.75, -26.5) {};
		\node [style=none] (137) at (27.25, -27.5) {};
		\node [style=none] (138) at (24.25, -27.5) {};
		\node [style=none] (139) at (25.75, -27) {\huge cookies};
		\node [style=none] (140) at (26.75, -27.5) {};
		\node [style=none] (141) at (25, -27.5) {};
		\node [style=none] (142) at (28, -27.5) {};
		\node [style=none] (143) at (30, -27.5) {};
		\node [style=none] (144) at (30.5, -28.5) {};
		\node [style=none] (145) at (27.5, -28.5) {};
		\node [style=none] (146) at (29, -28) {\huge tasty};
		\node [style=none] (147) at (29.75, -28.5) {};
		\node [style=none] (148) at (28, -28.5) {};
		\node [style=none] (149) at (23.75, -27.5) {};
		\node [style=none] (150) at (26.75, -28.5) {};
		\node [style=none] (151) at (21.75, -25.75) {\huge $N$};
		\node [style=none] (152) at (24.25, -25.75) {\huge $N$};
		\node [style=none] (153) at (23.25, -25.75) {\huge $S$};
		\node [style=none] (154) at (25.25, -28) {\huge $N$};
		\node [style=none] (155) at (27.25, -28) {\huge $N$};
		\node [style=none] (156) at (28.5, -29) {\huge $N$};
		\node [style=none] (157) at (29.75, -30) {};
		\node [style=none] (158) at (30.25, -29) {\huge $S$};
		\node [style=none] (166) at (23, -30) {};
		\node [style=none] (167) at (29.75, -30) {};
		\node [style={fill=black, draw=black, shape=circle}] (168) at (26.5, -31) {};
		\node [style=none] (169) at (26.5, -32.25) {};
		\node [style=none] (171) at (27, -31.5) {\huge $S$};
	\end{pgfonlayer}
	\begin{pgfonlayer}{edgelayer}
		\draw [line width=1pt] (0.center) to (3.center);
		\draw [line width=1pt] (3.center) to (2.center);
		\draw [line width=1pt] (2.center) to (1.center);
		\draw [line width=1pt] (1.center) to (0.center);
		\draw [line width=1pt] (5.center) to (8.center);
		\draw [line width=1pt] (8.center) to (7.center);
		\draw [line width=1pt] (7.center) to (6.center);
		\draw [line width=1pt] (6.center) to (5.center);
		\draw [line width=1pt] (10.center) to (13.center);
		\draw [line width=1pt] (13.center) to (12.center);
		\draw [line width=1pt] (12.center) to (11.center);
		\draw [line width=1pt] (11.center) to (10.center);
		\draw [line width=1pt] (17.center) to (20.center);
		\draw [line width=1pt] (20.center) to (19.center);
		\draw [line width=1pt] (19.center) to (18.center);
		\draw [line width=1pt] (18.center) to (17.center);
		\draw [line width=1pt] (24.center) to (25.center);
		\draw [line width=1pt] (26.center) to (27.center);
		\draw [line width=1pt] (28.center) to (29.center);
		\draw [line width=1pt] (30.center) to (32.center);
		\draw [line width=1pt] (31.center) to (33.center);
		\draw [line width=1pt] [bend right=15] (25.center) to (78);
		\draw [line width=1pt] [bend right=15, looseness=0.75] (78) to (33.center);
		\draw [line width=1pt] (78) to (79.center);
		\draw [line width=1pt] (120.center) to (123.center);
		\draw [line width=1pt] (123.center) to (122.center);
		\draw [line width=1pt] (122.center) to (121.center);
		\draw [line width=1pt] (121.center) to (120.center);
		\draw [line width=1pt] (125.center) to (128.center);
		\draw [line width=1pt] (128.center) to (127.center);
		\draw [line width=1pt] (127.center) to (126.center);
		\draw [line width=1pt] (126.center) to (125.center);
		\draw [line width=1pt] (130.center) to (131.center);
		\draw [line width=1pt] (133.center) to (134.center);
		\draw [line width=1pt] (135.center) to (138.center);
		\draw [line width=1pt] (138.center) to (137.center);
		\draw [line width=1pt] (137.center) to (136.center);
		\draw [line width=1pt] (136.center) to (135.center);
		\draw [line width=1pt] (142.center) to (145.center);
		\draw [line width=1pt] (145.center) to (144.center);
		\draw [line width=1pt] (144.center) to (143.center);
		\draw [line width=1pt] (143.center) to (142.center);
		\draw [line width=1pt] (132.center) to (149.center);
		\draw [line width=1pt] [bend left=90, looseness=2.25] (141.center) to (149.center);
		\draw [line width=1pt] (140.center) to (150.center);
		\draw [line width=1pt] [bend right=90, looseness=2.00] (150.center) to (148.center);
		\draw [line width=1pt] (147.center) to (157.center);
		\draw [line width=1pt] [bend right=15] (166.center) to (168);
		\draw [line width=1pt] [bend right=15, looseness=0.75] (168) to (167.center);
		\draw [line width=1pt] (168) to (169.center);
	\end{pgfonlayer}
\end{tikzpicture}}
\endpgfgraphicnamed}} 
  
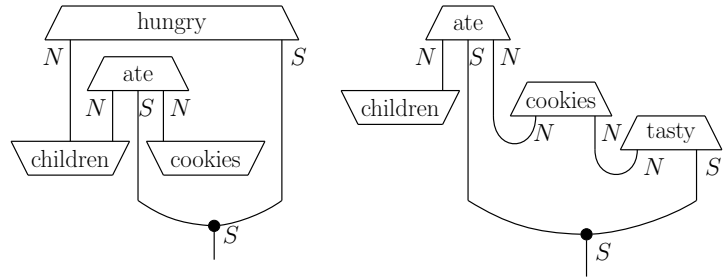
\captionof{figure}{Grammar with discourse Model}
    \label{fig:model4}
\end{center}

\end{itemize}

\subsection{Results}
\begin{figure}
\centering
\begin{subfigure}[b]{0.48\textwidth}
\begin{tikzpicture}[scale=0.75]
  \pgfplotsset{
      scale only axis,
  }

  \begin{axis}[
    title= Model 4a: $S_1 \odot S_2$,
    axis y line*=left,
    xlabel=iterations,
    ylabel=$y_1$,
  ]
    \addplot[red, mark=*]
      coordinates{
      (1,2.75752)
(2,2.869485)
(3,2.45094)
(4,2.391485)
(5,1.91237)
(6,1.894505)
(7,2.036315)
(8,2.104325)
(9,2.271785)
(10,1.95666)
(11,1.58983)
(12,1.43234)
(13,1.37862)
(14,1.468945)
(15,1.38974)
(16,1.396725)
(17,1.37607)
(18,1.541335)
(19,1.429355)
(20,1.345985)
(21,1.37429)
(22,1.419885)
(23,1.273375)
(24,1.194125)
(25,1.09988)
(26,1.18661)
(27,1.23588)
(28,1.144155)
(29,1.300955)
(30,1.17081)
(31,1.0993)
(32,1.04038)
(33,1.032935)
(34,0.91881)
(35,0.873295)
(36,0.84508)
(37,0.77443)
(38,0.768305)
(39,0.78993)
(40,0.82251)
(41,0.753245)
(42,0.72867)
(43,0.644465)
(44,0.640115)
(45,0.68573)
(46,0.632945)
(47,0.55087)
(48,0.49188)
(49,0.500765)
(50,0.50218)
(51,0.58485)
(52,0.4827)
(53,0.480935)
(54,0.470555)
(55,0.438965)
(56,0.385985)
(57,0.380405)
(58,0.39038)
(59,0.39275)
(60,0.39005)
(61,0.389335)
(62,0.34224)
(63,0.369405)
(64,0.34671)
(65,0.367105)
(66,0.32508)
(67,0.380425)
(68,0.331355)
(69,0.306615)
(70,0.28824)
(71,0.3074)
(72,0.291015)
(73,0.29048)
(74,0.27382)
(75,0.304535)
(76,0.279595)
(77,0.274955)
(78,0.275645)
(79,0.26505)
(80,0.264155)
(81,0.2578)
(82,0.257665)
(83,0.23849)
(84,0.249435)
(85,0.25736)
(86,0.240635)
(87,0.244595)
(88,0.22779)
(89,0.240255)
(90,0.22951)
(91,0.221325)
(92,0.23261)
(93,0.22646)
(94,0.212575)
(95,0.206)
(96,0.212395)
(97,0.22306)
(98,0.20905)
(99,0.20726)
(100,0.20814)
      }; \label{plot_1_y1}

    \end{axis}

    \begin{axis}[
      axis y line*=right,
      axis x line=none,
      ylabel=$y_2$,
    ]

    \addplot[blue, mark=x]
      coordinates{
      (1,0.528815)
(2,0.53159)
(3,0.511805)
(4,0.520145)
(5,0.57569)
(6,0.584725)
(7,0.544445)
(8,0.57569)
(9,0.552775)
(10,0.54444)
(11,0.573955)
(12,0.60522)
(13,0.61042)
(14,0.58958)
(15,0.617705)
(16,0.607635)
(17,0.63715)
(18,0.605905)
(19,0.61528)
(20,0.638545)
(21,0.640965)
(22,0.635065)
(23,0.630205)
(24,0.679855)
(25,0.688185)
(26,0.683685)
(27,0.67638)
(28,0.67744)
(29,0.671175)
(30,0.67848)
(31,0.682645)
(32,0.69028)
(33,0.689585)
(34,0.70139)
(35,0.73645)
(36,0.741665)
(37,0.769785)
(38,0.76701)
(39,0.76701)
(40,0.763555)
(41,0.762155)
(42,0.76909)
(43,0.80485)
(44,0.804155)
(45,0.798605)
(46,0.809025)
(47,0.827425)
(48,0.83993)
(49,0.84652)
(50,0.84097)
(51,0.85312)
(52,0.85278)
(53,0.86354)
(54,0.86806)
(55,0.872925)
(56,0.886105)
(57,0.883685)
(58,0.88959)
(59,0.88959)
(60,0.87604)
(61,0.87291)
(62,0.89028)
(63,0.89306)
(64,0.88749)
(65,0.890975)
(66,0.90278)
(67,0.90382)
(68,0.897915)
(69,0.91319)
(70,0.920835)
(71,0.916315)
(72,0.91527)
(73,0.91597)
(74,0.92327)
(75,0.914585)
(76,0.92396)
(77,0.92327)
(78,0.92466)
(79,0.93021)
(80,0.92743)
(81,0.93472)
(82,0.9375)
(83,0.935765)
(84,0.937845)
(85,0.94132)
(86,0.941315)
(87,0.946525)
(88,0.942005)
(89,0.94167)
(90,0.94513)
(91,0.94861)
(92,0.948605)
(93,0.949645)
(94,0.95729)
(95,0.95312)
(96,0.952435)
(97,0.95382)
(98,0.949995)
(99,0.953815)
(100,0.95451)  
      }; \label{plot_1_y2}
      
  \end{axis}
\end{tikzpicture}
\end{subfigure}%
\hfill
\begin{subfigure}[b]{0.48\textwidth}
\begin{tikzpicture}[scale=0.75]
  \pgfplotsset{
      scale only axis,
  }

  \begin{axis}[
    title= Model 4b: $S_1 R_Z S_2$,
    axis y line*=left,
    xlabel=iterations,
    ylabel=$y_1$,
  ]
    \addplot[red, mark=*]
      coordinates{
       (1,2.183635)
(2,2.20851)
(3,1.952655)
(4,1.741295)
(5,1.916645)
(6,1.733285)
(7,1.63513)
(8,1.740245)
(9,1.52542)
(10,1.68851)
(11,1.505665)
(12,1.37869)
(13,1.380845)
(14,1.363435)
(15,1.364895)
(16,1.368395)
(17,1.45551)
(18,1.344175)
(19,1.220055)
(20,1.17922)
(21,1.194855)
(22,1.11046)
(23,1.055155)
(24,1.101355)
(25,1.13376)
(26,1.12472)
(27,1.075405)
(28,1.045515)
(29,1.03874)
(30,1.040215)
(31,1.084725)
(32,1.051665)
(33,1.021375)
(34,1.034545)
(35,1.00639)
(36,0.887105)
(37,0.898675)
(38,0.82724)
(39,0.86182)
(40,0.84793)
(41,0.81832)
(42,0.85315)
(43,0.797615)
(44,0.833695)
(45,0.860595)
(46,0.926785)
(47,0.946805)
(48,0.85678)
(49,0.79579)
(50,0.785095)
(51,0.712585)
(52,0.73529)
(53,0.766615)
(54,0.710695)
(55,0.72063)
(56,0.699575)
(57,0.71324)
(58,0.705125)
(59,0.681095)
(60,0.695265)
(61,0.663985)
(62,0.668)
(63,0.64988)
(64,0.64934)
(65,0.606665)
(66,0.65674)
(67,0.629935)
(68,0.65142)
(69,0.613325)
(70,0.64967)
(71,0.57319)
(72,0.583245)
(73,0.59135)
(74,0.569315)
(75,0.5725)
(76,0.594325)
(77,0.545405)
(78,0.558795)
(79,0.564245)
(80,0.543445)
(81,0.532005)
(82,0.5815)
(83,0.513995)
(84,0.52113)
(85,0.50833)
(86,0.513925)
(87,0.53683)
(88,0.510005)
(89,0.507295)
(90,0.52527)
(91,0.54417)
(92,0.502245)
(93,0.51314)
(94,0.52112)
(95,0.51028)
(96,0.54359)
(97,0.504015)
(98,0.527435)
(99,0.536855)
(100,0.530725) 
      }; \label{plot_1_y1}

    \end{axis}

    \begin{axis}[
      axis y line*=right,
      axis x line=none,
      ylabel=$y_2$,
    ]

    \addplot[blue, mark=x]
      coordinates{
       (1,0.52187)
(2,0.504525)
(3,0.505895)
(4,0.538885)
(5,0.54236)
(6,0.540615)
(7,0.543045)
(8,0.53924)
(9,0.539585)
(10,0.559365)
(11,0.559705)
(12,0.577085)
(13,0.581935)
(14,0.577075)
(15,0.56389)
(16,0.565275)
(17,0.56736)
(18,0.561795)
(19,0.58193)
(20,0.575345)
(21,0.57812)
(22,0.59305)
(23,0.591675)
(24,0.58959)
(25,0.59618)
(26,0.61042)
(27,0.61215)
(28,0.63265)
(29,0.625)
(30,0.62499)
(31,0.61006)
(32,0.623615)
(33,0.6264)
(34,0.627085)
(35,0.63125)
(36,0.635065)
(37,0.650345)
(38,0.64272)
(39,0.644785)
(40,0.651735)
(41,0.664925)
(42,0.667)
(43,0.67534)
(44,0.657975)
(45,0.64374)
(46,0.653815)
(47,0.665975)
(48,0.673955)
(49,0.669445)
(50,0.688185)
(51,0.69654)
(52,0.688885)
(53,0.695825)
(54,0.699305)
(55,0.690975)
(56,0.7073)
(57,0.70695)
(58,0.71771)
(59,0.724655)
(60,0.72083)
(61,0.722915)
(62,0.730215)
(63,0.72778)
(64,0.72639)
(65,0.728815)
(66,0.7243)
(67,0.727775)
(68,0.733675)
(69,0.72536)
(70,0.717005)
(71,0.73543)
(72,0.73715)
(73,0.74027)
(74,0.7434)
(75,0.751395)
(76,0.74479)
(77,0.73714)
(78,0.75451)
(79,0.749305)
(80,0.75451)
(81,0.75659)
(82,0.759385)
(83,0.7618)
(84,0.762145)
(85,0.76041)
(86,0.762505)
(87,0.75973)
(88,0.77395)
(89,0.77638)
(90,0.781945)
(91,0.777075)
(92,0.77292)
(93,0.780545)
(94,0.78333)
(95,0.77917)
(96,0.77673)
(97,0.79271)
(98,0.77465)
(99,0.775345)
(100,0.777095)         
      }; \label{plot_1_y2}
      
  \end{axis}
\end{tikzpicture}
\end{subfigure}
\caption{Training loss and accuracy for Model 4}
\label{ref:resultModel4}
\end{figure}
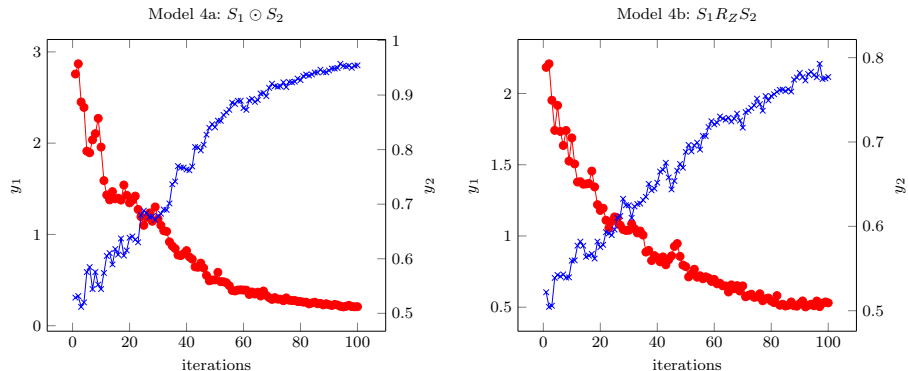

Figs. \ref{ref:resultModel1}, \ref{ref:resultModel2}, \ref{ref:resultModel3}, and \ref{ref:resultModel4} present the convergence of the models on the training dataset. Shown is the training loss (red, y1 axis) and the training accuracy (blue, y2 axis) where each line is from averaging over 20 runs of the optimisation with a random initial parameter point. For all the models, we fixed the hyper-parameters\footnote{The initial learning rate a = 0.05, the initial parameter shift scaling factor c = 0.06, a stability constant A = 0.01 * number of training steps.} and the number of iterations (100) to have a fair comparison. As is clear from the plots, the training converges smoothly in all cases. We notice that in Models 2, 3 and 4, the cost function converges faster when sentences $S_1$ and $S_2$ are connected with a Frobenius rather than the controlled-$R_z$ gate. For instance, after 60 iterations, Model 2a converges to a loss of (0.775) while Model 2b to (0.824); Model 3a to (0.749) while Model 3b to (0.861); Model 4a to (0.39) while Model 4b to (0.695). 

\begin{table}[t!]
\centering
\begin{tabular}{@{}lccccl@{}}
\toprule
 & No. & Model                    & $S_1 \odot S_2$ & $S_1 R_Z S_2$ &  \\ \midrule
 & 1   & No-grammar, no-discourse & 64.44\%         & 51.52\%       &  \\
 & 2   & No-grammar, discourse    & 100\%           & 67.70\%       &  \\
 & 3   & Grammar, no-discourse    & 72.84\%         & 48.97\%       &  \\
 & 4   & Grammar, discourse       & 91.38\%         & 76.67\%       &  \\ \bottomrule
\end{tabular}
\caption{Average test accuracy results for our 8 models.}
\label{tab:results}
\end{table}

Table \ref{tab:results} shows the test accuracy results for our 8 models. It's  clear that models with $S_1$ and $S_2$ being connected with a Frobenius performed better. Model 1 achieved a 12\% increase in accuracy; Model 2 a 32\%; Model 3 a 23\%; Model 4 a 14.71\%.  On the contrary, in the plots of models 1b and 3b show, the training accuracy remained around 0.5 without any increase.  Models with a controlled-$R_z$ gate may need more time and iterations to train. 

On the other hand, we observe a remarkable improvement in test accuracy after connecting the pronoun with its anaphora. In Model 1a, the recorded value was (0.64) after 100 iterations. This accuracy increased about 35\% in Model 2a. Similarly, the accuracy in Model 1b fluctuated around (0.51) while it increased to reach (0.67) in Model 2b. The same scenario is repeated when we compare Model 3a (0.72) with Model 4a (0.91) and Model 3b (0.48) with Model 4b (0.76). Also, we can see a boost in performance when we introduce the notion of grammar to the models. For instance, the accuracy increased 8.4\% between Model 1a and Model 3a, and around 8.9\% between Model 2b and Model 4b. However this wasn't the case between Models 2a and 4a even though both models performed well. 

\begin{figure}
\centering
\begin{subfigure}[b]{0.48\textwidth}
\begin{tikzpicture}[scale=0.75]
  \pgfplotsset{
      scale only axis,
  }

  \begin{axis}[
    title= Model 1a: $S_1 \odot S_2$,
    axis y line*=left,
    xlabel=iterations,
    ylabel=$y_1$,
  ]
    \addplot[red, mark=*]
      coordinates{
        (1,5.373705)
(2,5.100625)
(3,5.14622)
(4,4.93256)
(5,4.82098)
(6,5.17322)
(7,5.487325)
(8,5.67548)
(9,5.25987)
(10,4.99268)
(11,5.16295)
(12,4.823495)
(13,5.09242)
(14,5.03314)
(15,5.11022)
(16,5.11569)
(17,5.52281)
(18,5.00383)
(19,4.889135)
(20,4.83725)
(21,5.159105)
(22,4.121245)
(23,4.033325)
(24,4.645845)
(25,4.389865)
(26,4.62538)
(27,4.63519)
(28,4.453195)
(29,4.598775)
(30,4.46284)
(31,4.253055)
(32,4.412405)
(33,4.326585)
(34,4.638385)
(35,4.850865)
(36,4.179595)
(37,4.573905)
(38,4.66112)
(39,4.127865)
(40,4.313375)
(41,4.64199)
(42,4.255245)
(43,4.279825)
(44,3.942995)
(45,4.05055)
(46,4.00731)
(47,4.05581)
(48,4.229195)
(49,4.03384)
(50,3.83109)
(51,3.822085)
(52,3.949875)
(53,3.895425)
(54,4.029295)
(55,4.17646)
(56,4.32479)
(57,4.45641)
(58,4.48115)
(59,4.38569)
(60,4.178315)
(61,3.358275)
(62,3.61463)
(63,3.86581)
(64,3.54149)
(65,3.702675)
(66,3.741055)
(67,3.323715)
(68,3.184015)
(69,3.562335)
(70,3.541585)
(71,4.036765)
(72,4.32542)
(73,4.247755)
(74,3.816005)
(75,3.82274)
(76,3.68845)
(77,3.764145)
(78,3.290065)
(79,3.496545)
(80,3.38648)
(81,3.797175)
(82,3.63153)
(83,3.732645)
(84,3.666885)
(85,3.87465)
(86,3.72461)
(87,3.92233)
(88,3.68924)
(89,3.558325)
(90,3.636795)
(91,3.070615)
(92,3.07515)
(93,2.71238)
(94,2.475995)
(95,2.572835)
(96,3.208235)
(97,3.252175)
(98,3.094155)
(99,3.23262)
(100,3.017865)
      }; \label{plot_1_y1}

    \end{axis}

    \begin{axis}[
      axis y line*=right,
      axis x line=none,
      ylabel=$y_2$,
    ]

    \addplot[blue, mark=x]
      coordinates{
        (1,0.505565)
(2,0.49723)
(3,0.51354)
(4,0.53784)
(5,0.51736)
(6,0.520135)
(7,0.49862)
(8,0.48646)
(9,0.486455)
(10,0.48473)
(11,0.481595)
(12,0.50139)
(13,0.51424)
(14,0.49131)
(15,0.49096)
(16,0.4986)
(17,0.474655)
(18,0.485765)
(19,0.527775)
(20,0.50451)
(21,0.51458)
(22,0.53854)
(23,0.548945)
(24,0.545485)
(25,0.553825)
(26,0.52431)
(27,0.51631)
(28,0.530895)
(29,0.53576)
(30,0.50348)
(31,0.53818)
(32,0.54999)
(33,0.50832)
(34,0.51632)
(35,0.536115)
(36,0.53125)
(37,0.537155)
(38,0.55694)
(39,0.585415)
(40,0.552425)
(41,0.563895)
(42,0.55175)
(43,0.556945)
(44,0.588195)
(45,0.581255)
(46,0.562135)
(47,0.53265)
(48,0.548265)
(49,0.557635)
(50,0.5736)
(51,0.55799)
(52,0.532645)
(53,0.55034)
(54,0.55972)
(55,0.54791)
(56,0.552085)
(57,0.56493)
(58,0.55556)
(59,0.54271)
(60,0.537495)
(61,0.56423)
(62,0.5743)
(63,0.572915)
(64,0.587145)
(65,0.570485)
(66,0.559025)
(67,0.59827)
(68,0.59652)
(69,0.58612)
(70,0.599655)
(71,0.557985)
(72,0.57257)
(73,0.54653)
(74,0.58334)
(75,0.58229)
(76,0.583675)
(77,0.55277)
(78,0.578125)
(79,0.57987)
(80,0.60347)
(81,0.583675)
(82,0.586805)
(83,0.58194)
(84,0.563195)
(85,0.58298)
(86,0.59131)
(87,0.567355)
(88,0.59167)
(89,0.58785)
(90,0.60938)
(91,0.6191)
(92,0.624305)
(93,0.66597)
(94,0.641325)
(95,0.65416)
(96,0.60937)
(97,0.61597)
(98,0.647215)
(99,0.64132)
(100,0.64756)
      }; \label{plot_1_y2}
      
  \end{axis}
\end{tikzpicture}
\end{subfigure}%
\hfill
\begin{subfigure}[b]{0.48\textwidth}
\begin{tikzpicture}[scale=0.75]
  \pgfplotsset{
      scale only axis,
  }

  \begin{axis}[
    title= Model 1b: $S_1 R_Z S_2$,
    axis y line*=left,
    xlabel=iterations,
    ylabel=$y_1$,
  ]
    \addplot[red, mark=*]
      coordinates{
        (1,1.702701661)
(2,1.609060393)
(3,1.654414938)
(4,1.678650577)
(5,1.684397954)
(6,1.567432692)
(7,1.406033403)
(8,1.549080318)
(9,1.397952401)
(10,1.331146971)
(11,1.392035724)
(12,1.383242198)
(13,1.416372506)
(14,1.341767883)
(15,1.423750308)
(16,1.340588229)
(17,1.284691966)
(18,1.336836999)
(19,1.243857446)
(20,1.207783199)
(21,1.134641303)
(22,1.115011207)
(23,1.133380708)
(24,1.103165432)
(25,1.124839629)
(26,1.157189284)
(27,1.071112391)
(28,1.037650531)
(29,1.053527082)
(30,0.978771724)
(31,0.999837007)
(32,1.059160975)
(33,1.003738594)
(34,0.95245757)
(35,0.945810876)
(36,0.96543314)
(37,0.967899735)
(38,0.969460846)
(39,0.901753253)
(40,0.917338498)
(41,0.907680657)
(42,0.907315873)
(43,0.836362115)
(44,0.865016906)
(45,0.874571114)
(46,0.908297221)
(47,0.812519406)
(48,0.818336092)
(49,0.859996775)
(50,0.818411148)
(51,0.779946079)
(52,0.824412847)
(53,0.7981019)
(54,0.768469052)
(55,0.791848795)
(56,0.796273393)
(57,0.766587223)
(58,0.747645659)
(59,0.767587298)
(60,0.730382948)
(61,0.728191716)
(62,0.722579084)
(63,0.726556742)
(64,0.734591046)
(65,0.720573563)
(66,0.716412923)
(67,0.718804515)
(68,0.714902439)
(69,0.714465968)
(70,0.712560308)
(71,0.718897299)
(72,0.72006741)
(73,0.706322475)
(74,0.708998388)
(75,0.706091237)
(76,0.710198389)
(77,0.712616866)
(78,0.715460814)
(79,0.712900307)
(80,0.704732273)
(81,0.70880513)
(82,0.712713689)
(83,0.713963377)
(84,0.707525989)
(85,0.710828773)
(86,0.708826377)
(87,0.704469927)
(88,0.713798045)
(89,0.713068475)
(90,0.707717962)
(91,0.708146578)
(92,0.714244243)
(93,0.711735296)
(94,0.703543228)
(95,0.70775593)
(96,0.705010343)
(97,0.707032578)
(98,0.705064933)
(99,0.706552238)
(100,0.707420528)
      }; \label{plot_1_y1}

    \end{axis}

    \begin{axis}[
      axis y line*=right,
      axis x line=none,
      ylabel=$y_2$,
    ]

    \addplot[blue, mark=x]
      coordinates{
        (1,0.506944444)
(2,0.529166667)
(3,0.520138889)
(4,0.504861111)
(5,0.502430556)
(6,0.493402778)
(7,0.492013889)
(8,0.502083333)
(9,0.49375)
(10,0.498263889)
(11,0.4875)
(12,0.496527778)
(13,0.505555556)
(14,0.495833333)
(15,0.517361111)
(16,0.509375)
(17,0.502430556)
(18,0.503472222)
(19,0.507638889)
(20,0.521180556)
(21,0.503125)
(22,0.504861111)
(23,0.503125)
(24,0.498611111)
(25,0.508333333)
(26,0.495486111)
(27,0.493055556)
(28,0.494444444)
(29,0.502777778)
(30,0.513541667)
(31,0.497916667)
(32,0.481597222)
(33,0.500347222)
(34,0.497222222)
(35,0.518055556)
(36,0.499652778)
(37,0.497569444)
(38,0.495486111)
(39,0.517361111)
(40,0.495138889)
(41,0.497916667)
(42,0.490972222)
(43,0.497222222)
(44,0.5125)
(45,0.498263889)
(46,0.48125)
(47,0.500694444)
(48,0.498611111)
(49,0.509722222)
(50,0.491666667)
(51,0.480902778)
(52,0.490277778)
(53,0.490625)
(54,0.494444444)
(55,0.481597222)
(56,0.491666667)
(57,0.480555556)
(58,0.5)
(59,0.498611111)
(60,0.506944444)
(61,0.476041667)
(62,0.487847222)
(63,0.491319444)
(64,0.494097222)
(65,0.499652778)
(66,0.504861111)
(67,0.502777778)
(68,0.525694444)
(69,0.485416667)
(70,0.503125)
(71,0.489236111)
(72,0.523611111)
(73,0.514236111)
(74,0.505208333)
(75,0.516319444)
(76,0.5)
(77,0.484375)
(78,0.495833333)
(79,0.493402778)
(80,0.498958333)
(81,0.502430556)
(82,0.490277778)
(83,0.485416667)
(84,0.494097222)
(85,0.494444444)
(86,0.523611111)
(87,0.506944444)
(88,0.494791667)
(89,0.494444444)
(90,0.499652778)
(91,0.477430556)
(92,0.503819444)
(93,0.511458333)
(94,0.506944444)
(95,0.501388889)
(96,0.503472222)
(97,0.486805556)
(98,0.529166667)
(99,0.496180556)
(100,0.507291667)
      }; \label{plot_1_y2}
      
  \end{axis}
\end{tikzpicture}
\end{subfigure}
\caption{Training loss and accuracy for Model 1}
\label{ref:resultModel1}
\end{figure}
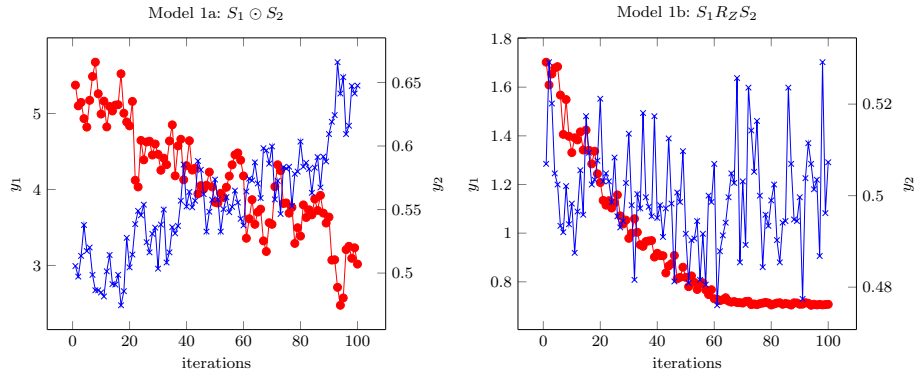


\begin{figure}
\centering
\begin{subfigure}[b]{0.48\textwidth}
\begin{tikzpicture}[scale=0.75]
  \pgfplotsset{
      scale only axis,
  }

  \begin{axis}[
    title= Model 2a: $S_1 \odot S_2$,
    axis y line*=left,
    xlabel=iterations,
    ylabel=$y_1$,
  ]
    \addplot[red, mark=*]
      coordinates{
      (1,4.337960975)
(2,4.072742794)
(3,4.624509707)
(4,4.407607209)
(5,3.392812174)
(6,3.356061234)
(7,3.833909711)
(8,4.29493371)
(9,3.26642563)
(10,3.337923565)
(11,3.6125442)
(12,3.791658987)
(13,4.099865735)
(14,4.058164286)
(15,3.74602071)
(16,3.618394713)
(17,3.673114759)
(18,3.820887873)
(19,3.299356107)
(20,3.209244883)
(21,3.078984535)
(22,2.591524821)
(23,2.498875363)
(24,2.528195388)
(25,3.020649341)
(26,2.885931418)
(27,2.457573706)
(28,2.491676994)
(29,1.971897274)
(30,1.928145663)
(31,2.26695621)
(32,2.479193144)
(33,2.204452397)
(34,2.342993642)
(35,1.694248005)
(36,2.117541636)
(37,1.858130598)
(38,2.521679122)
(39,2.334149256)
(40,2.155264339)
(41,2.16713428)
(42,2.194633866)
(43,2.447917184)
(44,1.904587398)
(45,1.881857217)
(46,1.69096116)
(47,1.475665908)
(48,1.472592999)
(49,1.417090954)
(50,1.036028637)
(51,0.964859151)
(52,0.828226587)
(53,0.822586915)
(54,0.583882605)
(55,0.558814074)
(56,0.656942852)
(57,0.732583405)
(58,0.740353764)
(59,0.773044194)
(60,0.775490134)
(61,0.521329366)
(62,0.359854516)
(63,0.313935259)
(64,0.34437979)
(65,0.295966059)
(66,0.346585529)
(67,0.2196915)
(68,0.327274071)
(69,0.352869342)
(70,0.367596966)
(71,0.286028839)
(72,0.184760894)
(73,0.165614489)
(74,0.129356089)
(75,0.121779334)
(76,0.190772196)
(77,0.185832082)
(78,0.173319919)
(79,0.138133677)
(80,0.203929282)
(81,0.202310502)
(82,0.135822413)
(83,0.134332072)
(84,0.172341028)
(85,0.193914313)
(86,0.128254581)
(87,0.107145958)
(88,0.105895344)
(89,0.182151623)
(90,0.294174047)
(91,0.265466705)
(92,0.089748748)
(93,0.064164532)
(94,0.075273525)
(95,0.084850424)
(96,0.073799589)
(97,0.062780458)
(98,0.054808506)
(99,0.049622938)
(100,0.054225731)  
      }; \label{plot_1_y1}

    \end{axis}

    \begin{axis}[
      axis y line*=right,
      axis x line=none,
      ylabel=$y_2$,
    ]

    \addplot[blue, mark=x]
      coordinates{
        (1,0.528125)
(2,0.48125)
(3,0.456597222)
(4,0.476388889)
(5,0.538194444)
(6,0.547569444)
(7,0.523263889)
(8,0.458333333)
(9,0.539236111)
(10,0.588541667)
(11,0.539236111)
(12,0.544097222)
(13,0.53125)
(14,0.551736111)
(15,0.551041667)
(16,0.543055556)
(17,0.576388889)
(18,0.531944444)
(19,0.560763889)
(20,0.554166667)
(21,0.599305556)
(22,0.632986111)
(23,0.634027778)
(24,0.651388889)
(25,0.64375)
(26,0.633333333)
(27,0.677083333)
(28,0.669097222)
(29,0.707986111)
(30,0.7)
(31,0.684027778)
(32,0.677777778)
(33,0.697222222)
(34,0.695486111)
(35,0.729513889)
(36,0.705902778)
(37,0.761805556)
(38,0.686111111)
(39,0.694097222)
(40,0.696527778)
(41,0.716666667)
(42,0.689236111)
(43,0.684027778)
(44,0.734027778)
(45,0.752430556)
(46,0.742708333)
(47,0.7625)
(48,0.774305556)
(49,0.781597222)
(50,0.805555556)
(51,0.847916667)
(52,0.838541667)
(53,0.839583333)
(54,0.882638889)
(55,0.896875)
(56,0.889930556)
(57,0.889236111)
(58,0.884722222)
(59,0.876041667)
(60,0.896180556)
(61,0.924652778)
(62,0.934027778)
(63,0.961805556)
(64,0.947222222)
(65,0.956597222)
(66,0.950694444)
(67,0.970833333)
(68,0.966666667)
(69,0.955555556)
(70,0.961805556)
(71,0.967013889)
(72,0.978472222)
(73,0.983333333)
(74,0.988194444)
(75,0.9875)
(76,0.982638889)
(77,0.972916667)
(78,0.965972222)
(79,0.968055556)
(80,0.971527778)
(81,0.977430556)
(82,0.973611111)
(83,0.973611111)
(84,0.972916667)
(85,0.975)
(86,0.971527778)
(87,0.974652778)
(88,0.979861111)
(89,0.974305556)
(90,0.969444444)
(91,0.967361111)
(92,0.985416667)
(93,0.988194444)
(94,0.990972222)
(95,0.988888889)
(96,0.988888889)
(97,0.992013889)
(98,0.994791667)
(99,0.998611111)
(100,0.997222222)
      }; \label{plot_1_y2}
      
  \end{axis}
\end{tikzpicture}
\end{subfigure}%
\hfill
\begin{subfigure}[b]{0.48\textwidth}
\begin{tikzpicture}[scale=0.75]
  \pgfplotsset{
      scale only axis,
  }

  \begin{axis}[
    title= Model 2b: $S_1 R_Z S_2$,
    axis y line*=left,
    xlabel=iterations,
    ylabel=$y_1$,
  ]
    \addplot[red, mark=*]
      coordinates{
        (1,1.63004399)
(2,1.607469732)
(3,1.547432604)
(4,1.458675505)
(5,1.354429335)
(6,1.394240349)
(7,1.478064311)
(8,1.363230591)
(9,1.458517957)
(10,1.412372752)
(11,1.391241595)
(12,1.379996467)
(13,1.290582816)
(14,1.251252467)
(15,1.273144113)
(16,1.288605202)
(17,1.284764205)
(18,1.171027245)
(19,1.171023814)
(20,1.118416182)
(21,1.086720598)
(22,1.165440931)
(23,1.14238846)
(24,1.200717471)
(25,1.188062412)
(26,1.161771632)
(27,1.126359462)
(28,1.186806332)
(29,1.163053344)
(30,1.136763641)
(31,1.109705239)
(32,1.211168106)
(33,1.135649775)
(34,1.109481547)
(35,1.126728608)
(36,1.068120097)
(37,1.080120336)
(38,1.02137685)
(39,0.975391474)
(40,1.014528087)
(41,0.998252511)
(42,1.009378238)
(43,0.874893783)
(44,1.010926217)
(45,0.942516269)
(46,0.907035411)
(47,0.886111787)
(48,0.961988094)
(49,0.960272281)
(50,0.928738762)
(51,0.899949292)
(52,0.893199229)
(53,0.912999673)
(54,0.980294465)
(55,0.928442691)
(56,0.836924628)
(57,0.890728955)
(58,0.90362514)
(59,0.865720077)
(60,0.824675191)
(61,0.86120905)
(62,0.840589107)
(63,0.84636391)
(64,0.852087655)
(65,0.807198209)
(66,0.770123909)
(67,0.789653858)
(68,0.791883407)
(69,0.841871175)
(70,0.833290929)
(71,0.803574772)
(72,0.836185625)
(73,0.807920139)
(74,0.801943531)
(75,0.799607664)
(76,0.825886161)
(77,0.763784533)
(78,0.716766567)
(79,0.757496846)
(80,0.714285989)
(81,0.726413078)
(82,0.728922708)
(83,0.710495337)
(84,0.732297295)
(85,0.71513931)
(86,0.696268622)
(87,0.681423105)
(88,0.67711903)
(89,0.685338068)
(90,0.662570191)
(91,0.609128381)
(92,0.630239982)
(93,0.643503093)
(94,0.631533306)
(95,0.641960261)
(96,0.605005871)
(97,0.582659533)
(98,0.613363968)
(99,0.622907626)
(100,0.606927912)
      }; \label{plot_1_y1}

    \end{axis}

    \begin{axis}[
      axis y line*=right,
      axis x line=none,
      ylabel=$y_2$,
    ]

    \addplot[blue, mark=x]
      coordinates{
(1,0.480208333)
(2,0.496180556)
(3,0.505555556)
(4,0.501736111)
(5,0.518055556)
(6,0.534722222)
(7,0.523958333)
(8,0.514583333)
(9,0.519097222)
(10,0.514930556)
(11,0.531944444)
(12,0.524652778)
(13,0.530208333)
(14,0.536458333)
(15,0.533333333)
(16,0.520486111)
(17,0.517013889)
(18,0.495138889)
(19,0.517013889)
(20,0.517361111)
(21,0.511805556)
(22,0.529166667)
(23,0.539583333)
(24,0.539930556)
(25,0.535416667)
(26,0.523263889)
(27,0.546527778)
(28,0.555902778)
(29,0.535763889)
(30,0.538541667)
(31,0.551736111)
(32,0.521875)
(33,0.55625)
(34,0.55625)
(35,0.546180556)
(36,0.549305556)
(37,0.564930556)
(38,0.570486111)
(39,0.560416667)
(40,0.567013889)
(41,0.564930556)
(42,0.590972222)
(43,0.584375)
(44,0.575694444)
(45,0.583680556)
(46,0.587152778)
(47,0.590625)
(48,0.568402778)
(49,0.578125)
(50,0.589583333)
(51,0.590972222)
(52,0.60625)
(53,0.607986111)
(54,0.610069444)
(55,0.602083333)
(56,0.607291667)
(57,0.598263889)
(58,0.603819444)
(59,0.604513889)
(60,0.626041667)
(61,0.623958333)
(62,0.61875)
(63,0.611458333)
(64,0.621180556)
(65,0.606597222)
(66,0.614236111)
(67,0.623958333)
(68,0.630208333)
(69,0.61875)
(70,0.609375)
(71,0.621527778)
(72,0.624652778)
(73,0.641666667)
(74,0.632291667)
(75,0.643402778)
(76,0.622569444)
(77,0.646527778)
(78,0.652777778)
(79,0.641666667)
(80,0.648958333)
(81,0.659375)
(82,0.643402778)
(83,0.666666667)
(84,0.668402778)
(85,0.670486111)
(86,0.661111111)
(87,0.656597222)
(88,0.670138889)
(89,0.676388889)
(90,0.678125)
(91,0.686111111)
(92,0.667361111)
(93,0.682638889)
(94,0.677430556)
(95,0.682986111)
(96,0.680555556)
(97,0.6875)
(98,0.688541667)
(99,0.679513889)
(100,0.687152778)                
      }; \label{plot_1_y2}
      
  \end{axis}
\end{tikzpicture}
\end{subfigure}
\caption{Training loss and accuracy for Model 2}
\label{ref:resultModel2}
\end{figure}
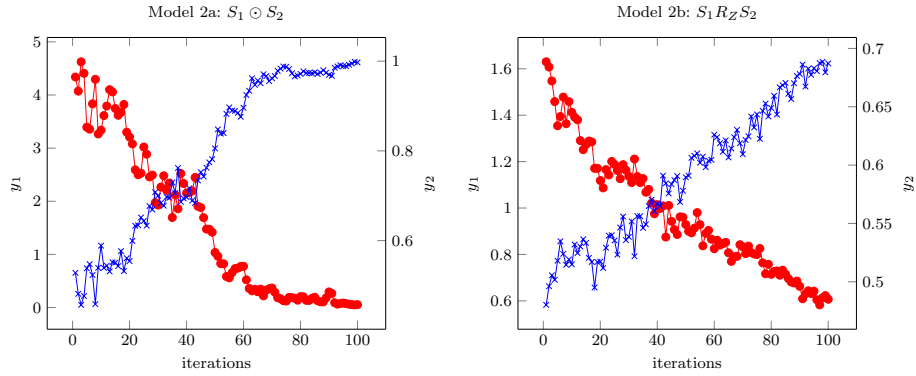


\begin{figure}
\centering
\begin{subfigure}[b]{0.48\textwidth}
\begin{tikzpicture}[scale=0.75]
  \pgfplotsset{
      scale only axis,
  }

  \begin{axis}[
    title= Model 3a: $S_1 \odot S_2$,
    axis y line*=left,
    xlabel=iterations,
    ylabel=$y_1$,
  ]
    \addplot[red, mark=*]
      coordinates{
      (1,2.99424766)
(2,2.96232728)
(3,2.8211767)
(4,2.86465037)
(5,2.56843667)
(6,2.58248851)
(7,2.89466613)
(8,2.62713844)
(9,2.44356646)
(10,2.71407275)
(11,2.46628852)
(12,2.38205028)
(13,2.16682316)
(14,2.03919011)
(15,2.01796626)
(16,1.96178002)
(17,1.91633697)
(18,1.92140474)
(19,1.81338541)
(20,1.66318149)
(21,1.77347055)
(22,1.83131267)
(23,1.67080126)
(24,1.54059251)
(25,1.78707789)
(26,1.81335745)
(27,1.54927898)
(28,1.47368059)
(29,1.51677985)
(30,1.64465961)
(31,1.43864223)
(32,1.42028999)
(33,1.34021868)
(34,1.52078924)
(35,1.30909829)
(36,1.29863791)
(37,1.24394473)
(38,1.07083847)
(39,1.12939656)
(40,1.20509871)
(41,1.1795425)
(42,1.13247186)
(43,1.09632843)
(44,1.01926627)
(45,1.00474813)
(46,0.9334696)
(47,0.84259192)
(48,0.85450771)
(49,0.86635266)
(50,0.94179825)
(51,0.76469034)
(52,0.77017037)
(53,0.80259411)
(54,0.80837597)
(55,0.75560883)
(56,0.80118621)
(57,0.88659296)
(58,0.95133531)
(59,0.79161723)
(60,0.74956037)
(61,0.66721713)
(62,0.64871065)
(63,0.64530139)
(64,0.67684547)
(65,0.67782235)
(66,0.60294864)
(67,0.62656133)
(68,0.62161034)
(69,0.60261983)
(70,0.62487297)
(71,0.600252)
(72,0.60685985)
(73,0.64374615)
(74,0.63378024)
(75,0.60962216)
(76,0.60063876)
(77,0.56718645)
(78,0.62158013)
(79,0.59217643)
(80,0.59577554)
(81,0.63041906)
(82,0.63440604)
(83,0.5823677)
(84,0.57306239)
(85,0.60755253)
(86,0.57281541)
(87,0.5762487)
(88,0.56277371)
(89,0.57515521)
(90,0.56506854)
(91,0.57095413)
(92,0.56168228)
(93,0.56696754)
(94,0.5542246)
(95,0.55954501)
(96,0.52789839)
(97,0.5388842)
(98,0.52808629)
(99,0.55529489)
(100,0.54458014)
      }; \label{plot_1_y1}

    \end{axis}

    \begin{axis}[
      axis y line*=right,
      axis x line=none,
      ylabel=$y_2$,
    ]

    \addplot[blue, mark=x]
      coordinates{
      (1,0.504166667)
(2,0.504861111)
(3,0.514583333)
(4,0.510416667)
(5,0.533333333)
(6,0.523958333)
(7,0.513194444)
(8,0.523958333)
(9,0.538541667)
(10,0.505555556)
(11,0.503819444)
(12,0.539583333)
(13,0.541319444)
(14,0.546527778)
(15,0.543402778)
(16,0.554166667)
(17,0.555208333)
(18,0.579166667)
(19,0.552430556)
(20,0.557638889)
(21,0.525)
(22,0.548958333)
(23,0.561111111)
(24,0.551736111)
(25,0.540625)
(26,0.553819444)
(27,0.578125)
(28,0.564583333)
(29,0.583680556)
(30,0.593055556)
(31,0.583680556)
(32,0.594444444)
(33,0.6125)
(34,0.615277778)
(35,0.624652778)
(36,0.650694444)
(37,0.612847222)
(38,0.644097222)
(39,0.642708333)
(40,0.640277778)
(41,0.639236111)
(42,0.639583333)
(43,0.633333333)
(44,0.635416667)
(45,0.646527778)
(46,0.653125)
(47,0.641319444)
(48,0.633333333)
(49,0.638541667)
(50,0.665277778)
(51,0.674652778)
(52,0.658680556)
(53,0.662847222)
(54,0.665972222)
(55,0.670138889)
(56,0.670486111)
(57,0.666666667)
(58,0.656944444)
(59,0.674305556)
(60,0.666666667)
(61,0.677430556)
(62,0.696180556)
(63,0.681597222)
(64,0.684375)
(65,0.687847222)
(66,0.694097222)
(67,0.7)
(68,0.713888889)
(69,0.709722222)
(70,0.697916667)
(71,0.713888889)
(72,0.709722222)
(73,0.717361111)
(74,0.707638889)
(75,0.709027778)
(76,0.713888889)
(77,0.729166667)
(78,0.720833333)
(79,0.730902778)
(80,0.725)
(81,0.732986111)
(82,0.735763889)
(83,0.748958333)
(84,0.744444444)
(85,0.726736111)
(86,0.729166667)
(87,0.747222222)
(88,0.738888889)
(89,0.740972222)
(90,0.733333333)
(91,0.740625)
(92,0.742708333)
(93,0.738888889)
(94,0.738194444)
(95,0.746180556)
(96,0.747222222)
(97,0.74375)
(98,0.760763889)
(99,0.754861111)
(100,0.752083333)  
      }; \label{plot_1_y2}
      
  \end{axis}
\end{tikzpicture}
\end{subfigure}%
\hfill
\begin{subfigure}[b]{0.48\textwidth}
\begin{tikzpicture}[scale=0.75]
  \pgfplotsset{
      scale only axis,
  }

  \begin{axis}[
    title= Model 3b: $S_1 R_Z S_2$,
    axis y line*=left,
    xlabel=iterations,
    ylabel=$y_1$,
  ]
    \addplot[red, mark=*]
      coordinates{
       (1,2.510062539)
(2,2.362245896)
(3,2.623185123)
(4,2.299381248)
(5,2.04970748)
(6,2.456837299)
(7,2.248534551)
(8,2.123259993)
(9,2.082568567)
(10,2.171451197)
(11,2.011119142)
(12,2.016848456)
(13,1.84898461)
(14,1.82273046)
(15,1.640701334)
(16,1.709147583)
(17,1.676797299)
(18,1.636656883)
(19,1.776783419)
(20,1.861573082)
(21,1.840516196)
(22,1.716011458)
(23,1.543907336)
(24,1.598139725)
(25,1.584259802)
(26,1.366955046)
(27,1.388614749)
(28,1.381828431)
(29,1.433454436)
(30,1.397982458)
(31,1.329783423)
(32,1.509309856)
(33,1.353177874)
(34,1.419702746)
(35,1.359259013)
(36,1.275324778)
(37,1.210776128)
(38,1.206628679)
(39,1.22556536)
(40,1.202320938)
(41,1.215600963)
(42,1.145261945)
(43,1.115444559)
(44,1.148232131)
(45,1.072493057)
(46,1.156500714)
(47,0.987728166)
(48,0.995698742)
(49,0.926566365)
(50,0.941084873)
(51,0.935456194)
(52,0.931144075)
(53,0.92377098)
(54,0.943326996)
(55,0.9204462)
(56,0.952213076)
(57,0.91211928)
(58,0.880818428)
(59,0.922266335)
(60,0.861084371)
(61,0.846398923)
(62,0.883679478)
(63,0.949425291)
(64,0.919229551)
(65,0.88353434)
(66,0.906114371)
(67,0.88528293)
(68,0.843140108)
(69,0.86020296)
(70,0.869222392)
(71,0.903991835)
(72,0.874419137)
(73,0.884827509)
(74,0.840640491)
(75,0.844326914)
(76,0.860470734)
(77,0.866416993)
(78,0.819142672)
(79,0.870972054)
(80,0.840549841)
(81,0.834203134)
(82,0.875383667)
(83,0.876967122)
(84,0.823858105)
(85,0.798803574)
(86,0.806316069)
(87,0.827744813)
(88,0.814118023)
(89,0.800494855)
(90,0.809275961)
(91,0.801691722)
(92,0.845566702)
(93,0.820716692)
(94,0.809349029)
(95,0.80215869)
(96,0.783461184)
(97,0.815991857)
(98,0.796422307)
(99,0.784908686)
(100,0.829683608) 
      }; \label{plot_1_y1}

    \end{axis}

    \begin{axis}[
      axis y line*=right,
      axis x line=none,
      ylabel=$y_2$,
    ]

    \addplot[blue, mark=x]
      coordinates{
      (1,0.504166667)
(2,0.504166667)
(3,0.500347222)
(4,0.503472222)
(5,0.509027778)
(6,0.497222222)
(7,0.480902778)
(8,0.491319444)
(9,0.493402778)
(10,0.500347222)
(11,0.506944444)
(12,0.495486111)
(13,0.495486111)
(14,0.509375)
(15,0.507986111)
(16,0.505902778)
(17,0.506944444)
(18,0.490277778)
(19,0.490277778)
(20,0.501736111)
(21,0.502083333)
(22,0.484027778)
(23,0.511111111)
(24,0.500694444)
(25,0.489930556)
(26,0.510069444)
(27,0.494791667)
(28,0.511111111)
(29,0.502777778)
(30,0.49375)
(31,0.502777778)
(32,0.481597222)
(33,0.508680556)
(34,0.507291667)
(35,0.502083333)
(36,0.505208333)
(37,0.501736111)
(38,0.502777778)
(39,0.485069444)
(40,0.506944444)
(41,0.493055556)
(42,0.488194444)
(43,0.498263889)
(44,0.499652778)
(45,0.493055556)
(46,0.50625)
(47,0.492013889)
(48,0.500694444)
(49,0.503819444)
(50,0.49375)
(51,0.491666667)
(52,0.502083333)
(53,0.506944444)
(54,0.500347222)
(55,0.495833333)
(56,0.501388889)
(57,0.516319444)
(58,0.510416667)
(59,0.489583333)
(60,0.49375)
(61,0.495833333)
(62,0.507291667)
(63,0.495833333)
(64,0.495833333)
(65,0.509027778)
(66,0.486111111)
(67,0.492361111)
(68,0.490972222)
(69,0.501388889)
(70,0.513194444)
(71,0.492708333)
(72,0.490625)
(73,0.501041667)
(74,0.498611111)
(75,0.517708333)
(76,0.492361111)
(77,0.498263889)
(78,0.494791667)
(79,0.497569444)
(80,0.514583333)
(81,0.50625)
(82,0.488194444)
(83,0.487847222)
(84,0.51875)
(85,0.498263889)
(86,0.482291667)
(87,0.492708333)
(88,0.503472222)
(89,0.489930556)
(90,0.497916667)
(91,0.487152778)
(92,0.507291667)
(93,0.480208333)
(94,0.491319444)
(95,0.480902778)
(96,0.484722222)
(97,0.513541667)
(98,0.493402778)
(99,0.510069444)
(100,0.497916667)          
      }; \label{plot_1_y2}
  \end{axis}
\end{tikzpicture}
\end{subfigure}
\caption{Training loss and accuracy for Model 3}
\label{ref:resultModel3}
\end{figure}
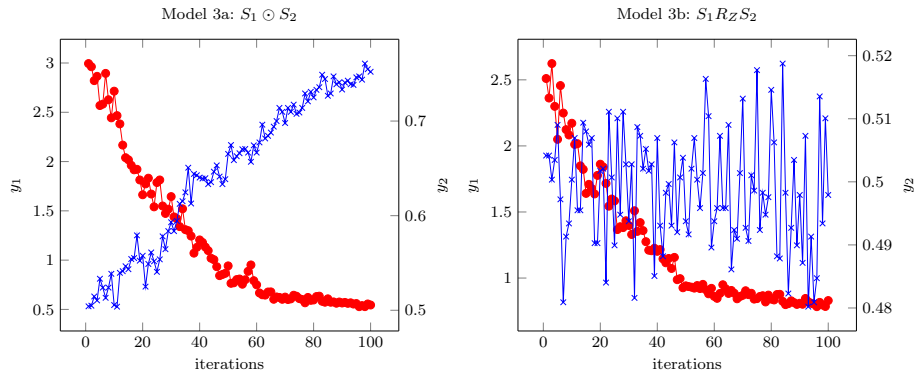 


\section{Summary, Conclusions, Future Work}

This paper had two main parts. The first part was on the theory of a specific modal Lambek calculus and its applications to modelling discourse relations such as anaphora and ellipsis. In this part, we focused on the calculus of Kanovich et. al \cite{Kanovich2020}, which we denote by $\sllm$. The modalities of this calculus are in the style of  the  \emph{soft} modalities of Linear Logic \cite{Lafont2004}. They  have the advantage that the logic containing them remains decidable, despite allowing for an implicit form of copying -- via the notion of storage and projections from it. A vector space semantics for this calculus  was developed in  previous work \cite{mcpheat2021LACL}, where the modal formulae were interpreted as \emph{truncated Fock spaces}, i.e. direct sums of tensor powers, but only up to a fixed given order $k_0$. We end the first part by introducing a novel string diagrammatic semantics for this truncated Fock space semantics.

In the  second part, we followed the line of work initiated in \cite{meichanetzidis2020quantum,QnlpInPractice} and translated our string diagrams to quantum circuits. Here, vector spaces are translated to quantum states, and the operations on them to quantum gates.  We translate our truncated Fock space semantics into a quantum circuits by extending  the existing translation.  We then applied our setting to a definite pronoun resolution task, and (1) develop a smaller version of the original dataset  of \cite{levesque2012}, elaborated on in \cite{rahman-ng-2012-resolving}, (2) model the pronominal anaphora relations with  modal formulae, (3) build the corresponding   Fock  semantics of them and after depicting the reasoning in diagrams, (4)  translate these to quantum circuits.   We  learn the parameters of the resulting circuits by simulating them in the AerSimulator \cite{IBMQ-sim}. Apart from the  $\sllm$ model,  we implement and experiment with three other models which either had a notion of grammar, or a notion of discourse, or neither of the two. The highest accuracies were recorded for models with a notion of discourse (i.e. the anaphora was resolved), irregardless of presence of a grammatical structure. Furthermore, we experimented with two different ways of combining the sentences of each discourse. Our  exploration of the combination operation between the sentences of each discourse revealed that the CNOT combination (Frobenius multiplication) achieves a higher accuracy  than a controlled- $R_Z$ rotation. 


It is tempting to immediately conclude that resolving discourse is more important than modelling grammatical structure, at least for definite pronoun resolution. There are, however, reasons to abstain from a quick conclusion. Most important of these is that our dataset is small;  experimentation on a larger dataset is work in progress. We would also like to model other types of coreference relations such as  ellipsis.  Moving away from simulations and training our circuit parameters on the real quantum computers is another avenue for future work.  Finally, efforts to develop Fock space quantum computers and developing NLP packages for them is work in progress, see for example \cite{shane2022QPL,Felice2022QPL}, and it would be more natural to use these computers for our task.

\bibliographystyle{plain}
\bibliography{references}

\begin{thebibliography}{10}

\bibitem{IBMQ}
{IBM} {Q}uantum.
\newblock \url{https://quantum-computing.ibm.com/}, 2021.

\bibitem{IBMQ-sim}
Qiskit: An open-source framework for quantum computing, 2021.

\bibitem{Coeckeetal2013}
Mehrnoosh~Sadrzadeh Bob~Coecke, Edward~Grefenstette.
\newblock Lambek vs. lambek: Functorial vector space semantics and string
  diagrams for lambek calculus.
\newblock {\em Ann. Pure and Applied Logic}, 164:1079 -- 1100, 2013.

\bibitem{shane2022QPL}
Alexandre Clément, Nicolas Heurtel, Shane Mansfield, Simon Perdrix, and
  Benoît Valiron.
\newblock Lov-calculus: A graphical language for linear optical quantum
  circuits, 2022.

\bibitem{Felice2022QPL}
Giovanni de~Felice and Bob Coecke.
\newblock Quantum linear optics via string diagrams, 2022.

\bibitem{de_Felice_2021}
Giovanni de~Felice, Alexis Toumi, and Bob Coecke.
\newblock {DisCoPy}: Monoidal categories in python.
\newblock {\em Electronic Proceedings in Theoretical Computer Science},
  333:183--197, feb 2021.

\bibitem{Havl_ek_2019}
Vojt{\v{e}}ch Havl{\'{\i}}{\v{c}}ek, Antonio~D. C{\'{o}}rcoles, Kristan Temme,
  Aram~W. Harrow, Abhinav Kandala, Jerry~M. Chow, and Jay~M. Gambetta.
\newblock Supervised learning with quantum-enhanced feature spaces.
\newblock {\em Nature}, 567(7747):209--212, mar 2019.

\bibitem{Humphreys}
J.~E. Humphreys.
\newblock {\em introduction to lie algebras and representation theory}.
\newblock Springer-Verlag, 1972.

\bibitem{jager1998multi}
G.~J\"{a}ger.
\newblock A multi-modal analysis of anaphora and ellipsis.
\newblock {\em University of Pennsylvania Working Papers in Linguistics},
  5(2):2, 1998.

\bibitem{jager2006anaphora}
G.~J\"{a}ger.
\newblock {\em Anaphora and type logical grammar}, volume~24.
\newblock Springer Science \&amp; Business Media, 2006.

\bibitem{Kanovich2020}
M.~Kanovich, S.~Kuznetsov, V.~Nigam, and A.~Scedrov.
\newblock {Soft Subexponentials and Multiplexing}.
\newblock In {\em Lecture Notes in Computer Science (including subseries
  Lecture Notes in Artificial Intelligence and Lecture Notes in
  Bioinformatics)}, 2020.

\bibitem{Kanovich2016}
M.~Kanovich, S.~Kuznetsov, and A.~Scedrov.
\newblock {Undecidability of the Lambek calculus with a relevant modality}.
\newblock {\em Lecture Notes in Computer Science (including subseries Lecture
  Notes in Artificial Intelligence and Lecture Notes in Bioinformatics)}, 9804
  LNCS:240--256, 2016.

\bibitem{lambeq_paper}
Dimitri Kartsaklis, Ian Fan, Richie Yeung, Anna Pearson, Robin Lorenz, Alexis
  Toumi, Giovanni de~Felice, Konstantinos Meichanetzidis, Stephen Clark, and
  Bob Coecke.
\newblock lambeq: An efficient high-level python library for quantum nlp, 2021.

\bibitem{Lafont2004}
Y.~Lafont.
\newblock {Soft linear logic and polynomial time}.
\newblock {\em Theoretical Computer Science}, 2004.

\bibitem{Lambek58}
Joachim Lambek.
\newblock {The Mathematics of Sentence Structure}.
\newblock {\em The American Mathematical Monthly}, 65(3):154, mar 1958.

\bibitem{levesque2012}
Hector~J. Levesque, Ernest Davis, and Leora Morgenstern.
\newblock {The winograd schema challenge}.
\newblock In {\em Proceedings of the International Workshop on Temporal
  Representation and Reasoning}, 2012.

\bibitem{QnlpInPractice}
Robin Lorenz, Anna Pearson, Konstantinos Meichanetzidis, Dimitri Kartsaklis,
  and Bob Coecke.
\newblock Qnlp in practice: Running compositional models of meaning on a
  quantum computer, 2021.

\bibitem{mcpheat2021LACL}
L.~McPheat, H.~Wazni, and M.~Sadrzadeh.
\newblock Vector space semantics for lambek calculus with soft subexponentials.
\newblock In {\em Proceedings of the tenth international conference on Logical
  Aspect of Computational Linguistics}, 2020.

\bibitem{meichanetzidis2020quantum}
Konstantinos Meichanetzidis, Stefano Gogioso, Giovanni~De Felice, Nicolò
  Chiappori, Alexis Toumi, and Bob Coecke.
\newblock Quantum natural language processing on near-term quantum computers,
  2020.

\bibitem{moortgat1996multimodal}
M.~Moortgat.
\newblock Multimodal linguistic inference.
\newblock {\em Journal of Logic, Language and Information}, 5(3-4):349--385,
  1996.

\bibitem{morrill2015computational}
Glyn Morrill and Oriol Valent{\'\i}n.
\newblock Computational coverage of tlg: Nonlinearity.
\newblock In {\em Proceedings of NLCS'15. Third Workshop on Natural Language
  and Computer Science}, volume~32, pages 51--63. EasyChair Publications, 2015.

\bibitem{morrill2016logic}
Glyn Morrill and Oriol Valent{\'\i}n.
\newblock On the logic of expansion in natural language.
\newblock In {\em Logical Aspects of Computational Linguistics. Celebrating 20
  Years of LACL (1996--2016) 9th International Conference, LACL 2016, Nancy,
  France, December 5-7, 2016, Proceedings 9}, pages 228--246. Springer, 2016.

\bibitem{rahman-ng-2012-resolving}
Altaf Rahman and Vincent Ng.
\newblock Resolving complex cases of definite pronouns: The winograd schema
  challenge.
\newblock In {\em Proceedings of the 2012 Joint Conference on Empirical Methods
  in Natural Language Processing and Computational Natural Language Learning},
  pages 777--789, Jeju Island, Korea, jul 2012. Association for Computational
  Linguistics.

\bibitem{Moortgatetal2019}
M.~Sadrzadeh, M.~Moortgat, and G.~Wijnholds.
\newblock A frobenius algebraic analysis for parasitic gaps.
\newblock In {\em Workshop on Semantic Spaces at the Intersection of NLP,
  Physics, and Cognitive Science}, Riga, Latvia, 2019.

\bibitem{Sadretal2013Frob}
Mehrnoosh Sadrzadeh, Stephen Clark, and Bob Coecke.
\newblock The frobenius anatomy of word meanings i: subject and object relative
  pronouns.
\newblock {\em Journal of Logic and Computation}, 23(6):1293--1317, 2013.

\bibitem{Sadrzadeh2016}
Mehrnoosh Sadrzadeh, Stephen Clark, and Bob Coecke.
\newblock {The Frobenius anatomy of word meanings II: Possessive relative
  pronouns}.
\newblock {\em Journal of Logic and Computation}, 26(2):785--815, 2016.

\bibitem{Shepherd_2009}
Dan Shepherd and Michael~J. Bremner.
\newblock Temporally unstructured quantum computation.
\newblock {\em Proceedings of the Royal Society A: Mathematical, Physical and
  Engineering Sciences}, (2105):1413--1439, feb 2009.

\bibitem{705889}
J.C. Spall.
\newblock Implementation of the simultaneous perturbation algorithm for
  stochastic optimization.
\newblock {\em IEEE Transactions on Aerospace and Electronic Systems},
  34(3):817--823, 1998.

\bibitem{Wijnholds2018}
Gijs Wijnholds and Mehrnoosh Sadrzadeh.
\newblock {Classical copying versus quantum entanglement in natural language:
  The case of VP-ellipsis}.
\newblock {\em Electronic Proceedings in Theoretical Computer Science, EPTCS},
  283:103--119, 2018.

\end{thebibliography}
\end{document}